\documentclass[numbers,webpdf,imaiai]{ima-preprint}

\theoremstyle{thmstyletwo}

\newtheorem{remark}{Remark}
\newtheorem*{prob*}{Problem}
\newtheorem*{obj*}{Objective}
\newtheorem{proposition}{Proposition}
\newtheorem{lemma}{Lemma}
\newtheorem{corollary}{Corollary}
\newtheorem*{assumption*}{\assumptionnumber}
\providecommand{\assumptionnumber}{}
\makeatletter
\newenvironment{assumption}[1]
 {  \renewcommand{\assumptionnumber}{Assumption A-#1}  \begin{assumption*}  \protected@edef\@currentlabel{#1} }
 {  \end{assumption*}
 }
\makeatother
\newcommand{\asref}[1]{\textbf{A-\ref{#1}}}
\newtheorem{definition}{Definition}

\numberwithin{equation}{section}

\usepackage{thm-restate}

\newcommand{\vertiii}[1]{{\left\vert\kern-0.25ex\left\vert\kern-0.25ex\left\vert #1
    \right\vert\kern-0.25ex\right\vert\kern-0.25ex\right\vert}}

\def\a{{\bf a}}
\def\Abf{{\mathbf{A}}}

\def\Bbf{{\bf B}}
\def\Cbf{{\bf C}}
\def\Dbf{{\bf D}}
\def\ebf{\mathbf{e}}
\def\Mbf{{\bf M}}
\def\sbf{\mathbf{s}}
\def\Tbf{{\mathbf{T}}}
\def\ubf{{\bf u}}
\def\U{{\bf U}}
\def\Ubf{{\mathbf{U}}}
\def\vbf{{\bf v}}
\def\V{{\bf V}}
\def\Vbf{\mathbf{V}}
\def\xbf{\mathbf{x}}
\def\X{{\bf X}}
\def\ybf{\mathbf{y}}
\def\Y{{\bf Y}}
\def\zbf{\mathbf{z}}
\def\Z{{\bf Z}}
\def\Sigmab{\boldsymbol\Sigma}
\def\Thetab{\boldsymbol\Theta}

\def\Acal{\mathcal{A}}

\def\Hcal{{\mathcal{H}}}
\def\Ncal{\mathcal{N}}
\def\Ocal{\mathcal{O}}
\def\Tcal{\mathcal{T}}
\def\Ucal{\mathcal{U}}

\def\E{\mathbb{E}}
\def\R{{\mathbb{R}}}
\def\Pbb{{\mathbb{P}}}
\def\Sbb{{\mathbb{S}}}

\newcommand{\Sfrak}{\mathfrak{S}}
\newcommand{\Xfrak}{\mathfrak{X}}

\newcommand{\dr}{\mathrm{D}}

\newcommand{\ssexpo}{K_\Lambda}
\newcommand{\D}{\Delta}
\newcommand{\integ}[1]{{[\![#1]\!]}}

\newcommand{\Fro}{{\operatorname{Fro}}}

\def\op{{\operatorname{op}}}

\def\tr{{\operatorname{tr}}}

\newcommand{\spec}{\operatorname{spec}}
\newcommand{\GLASSO}{\operatorname{GL}}
\newcommand{\dom}{\operatorname{dom}}
\newcommand{\interior}{\operatorname{int}}
\newcommand{\RIP}{\operatorname{RIP}}
\newcommand{\inv}{\operatorname{inv}}

\newcommand{\ie}{\textit{i.e.}}

\newcommand{\proba}[2]{\mathbb{P}_{#1}\left(#2 \right)}
\newcommand{\expect}[2]{\mathbb{E}_{#1}\left[ #2 \right]}

\begin{document}

\DOI{DOI HERE}
\copyrightyear{2021}
\vol{00}
\pubyear{2021}
\access{Advance Access Publication Date: Day Month Year}
\appnotes{Paper}
\copyrightstatement{Published by Oxford University Press on behalf of the Institute of Mathematics and its Applications. All rights reserved.}
\firstpage{1}

\title[Compressive Recovery of Sparse Precision Matrices]{Compressive Recovery of Sparse Precision Matrices}

\author{Titouan~Vayer*, Etienne~Lasalle, Rémi~Gribonval and Paulo~Gonçalves
\address{\orgname{Univ Lyon, Inria, CNRS, ENS de Lyon, UCB Lyon 1,\\
       LIP UMR 5668}, \orgaddress{\postcode{F-69342 Lyon}, \country{France}}}}

\authormark{T.~Vayer et al.}

\corresp[*]{Corresponding author: \href{email:titouan.vayer@ens-lyon.fr}{titouan.vayer@ens-lyon.fr}}

\received{Date}{0}{Year}
\revised{Date}{0}{Year}
\accepted{Date}{0}{Year}

\abstract{We consider the problem of learning a graph modeling the statistical relations of the $d$ variables from a dataset with $n$ samples $\mathbf{X} \in \mathbb{R}^{n \times d}$. Standard approaches amount to searching for a precision matrix $\boldsymbol\Theta$ representative of a Gaussian graphical model that adequately explains the data. However, most maximum likelihood-based estimators usually require storing the $d^{2}$ values of the empirical covariance matrix, which can become prohibitive in a high-dimensional setting. In this work, we adopt a ‘‘compressive'' viewpoint and aim to estimate a sparse $\boldsymbol\Theta$ from a \emph{sketch} of the data, \textit{i.e.} a low-dimensional vector of size $m \ll d^{2}$ carefully designed from $\mathbf{X}$ using non-linear random features. Under certain assumptions on the spectrum of $\boldsymbol\Theta$ (or its condition number), we show that it is possible to estimate it from a sketch of size $m=\Omega\left((d+2k)\log(d)\right)$ where $k$ is the maximal number of edges of the underlying graph. These information-theoretic guarantees are inspired by compressed sensing theory and involve restricted isometry properties and instance optimal decoders. We investigate the possibility of achieving practical recovery with an iterative algorithm based on the graphical lasso, viewed as a specific denoiser. We compare our approach and graphical lasso on synthetic datasets, demonstrating its favorable performance even when the dataset is compressed. }

\maketitle

\section{Introduction \label{sec:introduction}}
Inferring a complex network from data is used in many applications, such as in neuroscience for the treatment of epilepsy \cite{frusquephd}, for biological networks \cite{Jieping} or in genomics to identify gene interactions \cite{lingjaerde2021tailored,zuo2017incorporating}. We consider in this paper the problem of estimating a certain graph representing the statistical relations between $d$ variables from $n$ observed signals $\xbf_1, \cdots, \xbf_n$ where $\xbf_i \in \R^{d}$ follow a certain distribution $\mu$, assumed to be centered for simplicity and with covariance matrix $\Sigmab$. This graph is often associated with the so-called \emph{precision matrix} $\Thetab = \Sigmab^{-1}$ which is sparse in many practical situations due to limited conditional dependencies \cite{friedman_sparse_2008}. However, simply inverting the empirical covariance matrix $\widehat{\Sigmab}$ usually does not yield a sparse estimation of the precision matrix estimation. The challenge is thus to infer a sparse precision matrix from $\widehat{\Sigmab}$. One of the most popular methods to do this graph estimation is probably the graphical lasso \cite{friedman_sparse_2008,Banerjee}. Given the empirical covariance matrix
\begin{equation}
\widehat{\Sigmab} \stackrel{\D}{=} \frac{1}{n} \sum_{i=1}^{n} \xbf_i \xbf_i^{\top},
\label{eq:emp_cov}
\end{equation}
the graphical lasso estimator searches for a certain matrix $\widehat{\Thetab}_{\GLASSO} \in \R^{d \times d}$, representative of the graph edges, that solves the optimization problem
\begin{equation}
\label{eq:glasso}
\widehat{\Thetab}_{\GLASSO} \stackrel{\D}{=}\underset{\Thetab \succ 0}{\arg\min} -\log\det(\Thetab)+\langle \widehat{\Sigmab}, \Thetab \rangle+ \lambda \|\Thetab\|_{1,\operatorname{off}}\,,
\end{equation}
where the $\arg\min$ is taken over the set of symmetric positive definite matrices and $\|\Thetab\|_{1,\operatorname{off}}\stackrel{\D}= \sum_{i<j} |\Theta_{ij}|$ promotes sparsity. An interpretation is as follows: for a Gaussian model $\mu = \Ncal(0, \Sigmab = \Thetab^{-1})$, equation \eqref{eq:glasso} corresponds to an $\ell_1$-penalized maximum likehood estimator \cite{yuan2007model}. In other words, the graphical lasso estimates a sparse graph that best fits the data. 
More precisely, in the Gaussian model, the pattern of zeros of the precision matrix $\Thetab$ corresponds to conditional independencies among the variables, hence $\Thetab$ encodes the statistical relations between them
\cite{dempster1972covariance,lauritzen1996graphical}. Despite its many good properties, the graphical lasso suffers from a scaling problem with respect to the dimension. Indeed computing \eqref{eq:glasso} requires to store in memory the entire empirical covariance matrix $\widehat{\Sigmab}$ and thus has a space complexity of $\Ocal(d^{2})$; this is problematic for applications where $d$ is very large, such as high-resolution fMRI datasets or gene-microarray data where the number of genes $d$ is typically around tens of thousands.

Graphical models estimation is a very active field of research and many alternatives to graphical lasso have been proposed in the literature. Numerous strategies aim to address the computational scalability of the optimization problem \eqref{eq:glasso} through the use of approximation techniques, \textit{e.g.} QUIC \cite{ChoQUIC}, BIG \& QUIC \cite{bigQUIC}, SQUIC \cite{squic}. Others are looking for estimators that have either better statistical guarantees or better algorithmic properties. We can mention some estimators that rely on non-convex penalties \cite{ell0GLASSO,nonconvexglasso,barnajeenonconvex,scadglasso}, $\ell_2$ penalties \cite{rope} or other estimators like CLIME (and Dantzig-like estimators) \cite{clime,lin_prog}, RobustCLIME \cite{robustCLIME}, D-trace \cite{dtrace} or Elem-GGM \cite{elementary_esti}. Finally, some approaches make assumptions on the underlying graph (\textit{e.g.} that it is chordal) in order to find fast algorithms or try to constrain the structure of the graph sought in the estimation \cite{Ying2020,Sandeep,Ortega}.

\begin{figure}
\centering
\includegraphics[width=0.9\linewidth]{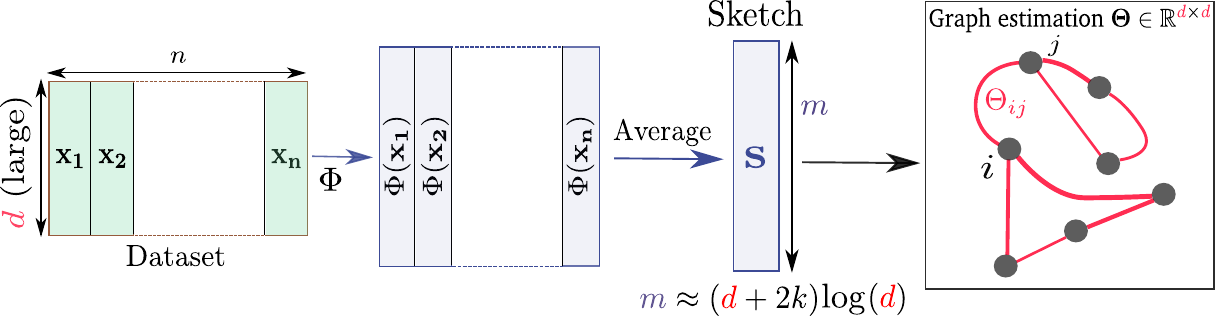}
\caption{Summary of our approach. The objective is to estimate a graph $\Thetab \in \R^{d \times d}$ between the $d^{2}$ variables $(i,j)$ of the data set from a sketch of the data, $\sbf \stackrel{\D}{=} \frac{1}{n} \sum_{i=1}^{n} \Phi(\xbf_i) \in \R^{m}$ . We will show in the following sections that, under certain graph sparsity assumptions, it is theoretically possible to estimate $\Thetab$ using a sketch of size $m \approx (d+2k)\log(d) \ll d^{2}$ where $k$ is the maximal number of edges of the underlying graph.}\label{fig:summary_fig}
\end{figure}

\subsection{Compressive learning of graphical models}

In this work we take another path: that of compression. Inspired by the theories of compressed sensing \cite{foucart13} and sketching \cite{gribonval:hal-03350599}, we try to estimate a sparse precision matrix $\Thetab$ associated to a covariance matrix $\Sigmab = \Thetab^{-1}$, not from the whole dataset, nor from the empirical covariance matrix $\widehat{\Sigmab}$, but from a \emph{compressed version} of the data called \emph{sketch}. More precisely, and given a well-chosen function $\Phi: \R^{d} \rightarrow \R^{m}$, we seek to estimate $\Thetab$ from the vector $\sbf \in \R^{m}$ defined by (see Figure \ref{fig:summary_fig})
\begin{equation}
\label{eq:sketch}
\sbf \stackrel{\D}{=} \frac{1}{n} \sum_{i=1}^{n} \Phi(\xbf_i) \in \R^{m}. 
\end{equation}
In sketching theory the $\Phi$ function is typically an adequately chosen \emph{random function} and the reconstruction guarantees are usually based on ideas from compressed sensing \cite{gribonval2020compressive}. The advantages of this type of approach are multiple: 1) when $m \ll n d$ the dataset is massively compressed, with benefits for storage and transfer 2) as averages, the sketches can be computed online, in one pass on the data, or in a distributed manner 3) as the data get aggregated, sketches have also good properties regarding differential privacy \cite{chatalic:tel-03023287}. The main objective of this paper is to use the sketching approach and to answer the following question:
\begin{obj*}
Can we find $\Phi$ and a sketch dimension $m \ll d^{2}$ such that the sparse precision matrix $\Thetab$ can be well estimated from a sketch of the data $\sbf = \frac{1}{n} \sum_{i=1}^{n} \Phi(\xbf_i) \in \R^{m}$ ?
\end{obj*}
In other words, can we obtain an estimate of the graph structure using a small-sized vector compared to the size of the empirical covariance matrix? We will shortly furnish a precise definition of what we consider to be ‘‘well estimated''. However, the underlying intuition guiding this inquiry is that, similar to the graphical lasso model, graphs estimated from data typically demonstrate sparsity. Therefore, there is hope that good approximations of these graphs can be obtained by accessing only a limited number of measurements compared to $\Ocal(d^2)$, the total number of elements in the covariance matrix.

\subsection{Contributions}
The article presents the following key contributions:
\begin{enumerate}
\item We investigate, from an information-theoretic perspective, a sketching mechanism where we project against \textit{i.i.d.} rank-one matrices. We show that when $m \gtrsim \|\Thetab\|_0 \log(d)$, where $\|\Thetab\|_0$ denotes the number of non-zero coefficients of $\Thetab$, the relevant information is preserved during the sketching phase. More precisely, we prove that the corresponding sketching operator satisfies a \textit{Restricted Isometry Property} (RIP), ensuring that the precision matrix can be reconstructed accurately using a so-called robust decoder. 
\item This robust decoder being \textit{a priori} intractable, we propose a practical decoding scheme to estimate the sparse precision matrix from the sketch of the data. The algorithm relies on an iterative procedure that alternates between a gradient descent step associated to a sketch fidelity term and a ‘‘denoising step'' performed by the graphical lasso. 
\item For practical applications, we propose to deviate from the theoretical framework of 1) and leverage structured matrices (\textit{e.g.,} Walsh-Hadamard matrices) to define a similar sketching mechanism but with boosted algorithmic efficiency. We demonstrate in the experiments that the combination of this sketching mechanism and our practical decoding scheme is able to significantly compress the data (\textit{i.e.,} $m \ll d^2$) while still retaining the ability to recover the covariance and precision matrices. 
\end{enumerate}

\subsection{Organization}
The paper is organized as follows. 
Section~\ref{sec:guarantees} outlines the main information-theoretic results. The main theorem (Theorem~\ref{theo:theRIPRO}) proves that the sketching operator satisfies a certain RIP with high probability. For an easier read, the technical part required to prove this theorem is deferred to the end of the paper, Section~\ref{sec:cov_and_concentration}, where we obtain the control of covering numbers and a concentration inequality for the sketching operator.
In Section~\ref{sec:optimization}, we present a random \emph{structured} matrix approach for the sketching phase (\textit{i.e.,} encoding phase) with better memory and computational efficiency. We also provide a practical algorithm to retrieve the precision matrix (\textit{i.e.,} decoding phase) from the sketch. Section~\ref{sec:experiments} presents the experiments on the encoding and decoding phases.

\subsection{Notations and definitions.} For any integer $m$, $\integ{m} \stackrel{\D}{=} \{1, \dots, m \}$. $S_d(\R)$ is the linear space of symmetric matrices of size $d \times d$, while $S_d^{++}(\R)$ (\textit{resp.} $S_d^{+}(\R)$) is the set of symmetric positive definite matrices (\textit{resp.} positive semi-definite). We often write $\Abf \in S_d^{++}(\R)$ as $\Abf \succ 0$ (\textit{resp} $\Abf \succeq 0$ for $S_d^{+}(\R)$). 
We denote by $\|\cdot\|_{2 \to 2}$ the $2$-operator norm or spectral norm for matrices \ie\ $\|\Abf\|_{2 \to 2} \stackrel{\D}{=} \sup_{\|\xbf\|_2 \leq 1} \|\Abf\xbf\|_2 = \sigma_{\max}(\Abf)$ where $\sigma_{\max}(\Abf)$ denotes the largest singular value of $\Abf$. The spectrum of $\Abf \in S_d(\R)$ is denoted by $\spec(\Abf)$.
The standard euclidean scalar product on matrices is denoted by $\langle \cdot, \cdot \rangle$ and its associated norm (the Frobenius norm) is denoted by $\|\cdot\|_{\Fro}$. We denote by $\|\Abf\|_{0}$ the number of non-zeros components of $\Abf$, that is to say the $\ell_0$ ‘‘norm'' of $\Abf$. We also introduce the function $\inv(\Abf)\stackrel{\D}{=}\Abf^{-1}$.

\section{Recovery guarantees \label{sec:guarantees}}

The sketching procedure proposed in this article is based on \emph{rank-one projections} and (well-chosen) \emph{randomness}. Recall that we are given a sample of $d$-dimensional vectors $(\xbf_1, \dots, \xbf_n)$, and we want to define a function $\Phi : \R^d \to \R^m$ such that the sketch is the average of the $m$-dimensional vectors $\Phi(\xbf_i)$ as in \eqref{eq:sketch}. Given a collection of $d$-dimensional random vectors $(\a_1, \dots \a_m)$ we consider in this work\footnote{Readers familiar with sketching methods might be puzzled by the factor $1/m$ rather than the usual $1/\sqrt{m}$. In our case, as we will consider the $\ell_1$-norm on $\R^m$, it is in fact the right scaling to obtain averages.}   
\begin{equation}
\Phi(\xbf) \stackrel{\D}{=} \frac{1}{m} \left( |\a_1^\top \xbf|^2, |\a_2^\top \xbf|^2, \cdots, |\a_m^\top \xbf|^2\right)^{\top}. \label{eq:ro}
\end{equation}
Before proceeding, it is important to note that $\Phi$ can also be computed as $\Phi(\xbf) = \frac{1}{m} \{ \a_j^\top \xbf \  \xbf^\top \a_j \}_{j \in \integ{m}} = \frac{1}{m} \{\langle \a_j \a_j^\top , \xbf \xbf^\top \rangle\}_{j \in \integ{m}}$. Therefore, by linearity of the inner product and equations \eqref{eq:emp_cov} and $\eqref{eq:sketch}$, the induced sketch $\sbf$ is the vector whose coordinates are the projections of the empirical covariance matrix $\widehat{\Sigmab}$ against the rank-one matrices $\a_j \a_j^\top$: $\sbf = \frac{1}{m} \{\langle \a_j \a_j^\top , \widehat{\Sigmab} \rangle\}_{j \in \integ{m}}$.

Consequently, the entry point of our theoretical analysis is to notice that the random function $\Phi$ defined in \eqref{eq:ro} involves a \emph{linear operator} on symmetric matrices \cite{delogne2022rop}. Indeed, our objective can be reformulated as finding $\Thetab = \Sigmab^{-1}$ from $\sbf= \Acal(\widehat{\Sigmab})$ where $\Acal: S_d(\R) \rightarrow \R^{m}$ is defined by
\begin{equation}
\Acal(\Mbf) \stackrel{\D}{=}  \frac{1}{m}\left\{ \langle \a_j \a_j^{\top} , \Mbf \rangle \right\}_{j \in \integ{m}}\,.
\label{eq:ROoperator}
\end{equation}
We can therefore reformulate our problem as a compressed sensing problem: given a noisy linear measurement of the true covariance matrix $\sbf = \Acal(\Sigmab)+\ebf \in \R^{m}$, where $\ebf = \Acal(\widehat{\Sigmab}-\Sigmab)$ is the noise, we want to recover the precision matrix $\Thetab = \Sigmab^{-1}$ that underlies certain sparsity assumptions. There are two difficulties here. First $m < d^{2}$, thus the problem is \textit{a priori} ill-posed. Second, the sparsity assumption does not involve the ‘‘signal'' $\Sigmab$ itself but a non-linear transformation of it, given by its inverse $\Thetab = \Sigmab^{-1}$.

\begin{remark}[Related works]
This sketching mechanism appears in various works on compressed sensing, in particular for low-rank matrix completion \cite{candeslowrank,ROParticle,Chen_yuxin,low_rank_Kabanava,KUENG201788}, estimation of low-rank \& positive semi-definite (PSD) matrices \cite{trace_reg,7605512}, low-rank \& sparse matrices \cite{7383810} or in statistical learning for PCA \cite{gribonval2020compressive} and in phase retrieval \cite{jaganathan2016phase}. Here we propose to study it for the \emph{recovery of sparse precision matrices}.
\end{remark}

\begin{remark}[Dense Gaussian projections]
We want to mention that another compressive scheme, maybe more natural for theoretical analysis, could have been considered. Instead of projecting against rank-one matrices, one could project against symmetric matrices with independent Gaussian entries \textit{via} $\Phi(\xbf) = \{\xbf^\top \Abf_j \xbf\}_{j \in \integ{m}}$ where $\Abf_j \in \R^{d \times d}$ are dense Gaussian matrices. The theoretical analysis of this approach would be quite straightforward following classical compressed sensing works \cite{foucart13, recipies_rip}. However, this sketching mechanism would be far from efficient as the computation of $\Phi(\xbf)$ would have an $\Ocal(md^2)$ time and space complexity, which is even greater than directly storing $\widehat{\Sigmab}$. Nevertheless, it is worth noting that \emph{optical computing}, which relies on optical processing units, could be leveraged to compute these random features in constant time in any dimension \cite{saade2016random}, albeit constrained by the current hardware capabilities.
\end{remark} 

\subsection{Recipe for recovery guarantees \label{sec:recipies}}

The general idea to obtain guarantees is the following: we will assume that $\Thetab$ lies in a ‘‘low dimensional'' set so that it is possible to estimate it from a very small number of measurements compared to the ambient dimension $d^{2}$ \cite{Bourrier}. This hypothesis is formalized here by assuming that the true covariance matrix $\Sigmab$ belongs to a certain subset of $S_d^{++}(\R)$, called \emph{model set}. More precisely, and given $k \in \mathbb{N}, 0< a \leq b$, we will study here the model set
\begin{equation}
\label{eq:model_set}
\Sfrak_{k,a,b} \stackrel{\D}{=}\left\{\Sigmab \in S_{d}^{++}(\R): \Thetab=\Sigmab^{-1} \succ 0, \|\Thetab\|_{0} \leq d+2k \text{ and } \spec(\Thetab) \subseteq [a,b]\right\}\,.
\end{equation}
This set corresponds to covariance matrices associated to the $d+2k$-sparse precision matrices whose spectra are well localized in $[a,b]$. This constraint on the spectrum of the precision matrix can be relaxed to a constraint on its condition number
as we will see later. Similar spectral conditions are usually assumed in the case of graphical lasso to derive sample complexities of the estimators \cite{clime,Wang_fast_and_sparse,LIU2015153}.
The number $k$ represents the maximal number of non-zero components of $\Thetab$ on its upper triangular part, 
\textit{i.e.,} the maximal number of edges of the graph corresponding to $\Thetab$, without counting self-loops.
The sparsity of the graph is also quite natural in many applications where one wants to have a simple graph thus with few connections. 

We will see later that $\Sfrak_{k,a,b}$ is the image by $\inv$ of a ‘‘low dimensional'' set and, by using a certain notion of stability of the inverse, we will be able to recover $\Thetab$ from $\sbf$. We emphasize that while we have presented this specific model set, the majority of the results in this paper are general and can be extended to accommodate other model sets. Consequently, these results can apply to alternative assumptions concerning the underlying graph structure.  

\bigskip

The central property for our guarantees is the \emph{Restricted Isometric Property} (RIP) \cite{recipies_rip,candes2005decoding}, adapted to our context. In the following, we consider two general norms $\|\cdot\|_{S_d}$ and $\|\cdot\|_{\R^m}$ on $S_d$ and $\R^m$, respectively. 
\begin{definition}[RIP]
\label{def:def_rip}
A linear operator $\Acal: (S_d(\R), \|\cdot\|_{S_d}) \rightarrow (\R^{m}, \|\cdot\|_{\R^{m}})$ satisfies the $\RIP_{\delta}$ for some $\delta \in [0,1[$ on a set $\Sfrak \subset S_d^{++}(\R)$ if for every $(\Sigmab_1, \Sigmab_2) \in \Sfrak^{2}$, it satisfies
\begin{equation}
(1-\delta) \|\Sigmab_1-\Sigmab_2\|_{S_d} \leq \|\Acal(\Sigmab_1-\Sigmab_2)\|_{\R^{m}} \leq (1+\delta)\|\Sigmab_1-\Sigmab_2\|_{S_d}\,.
\end{equation}
\end{definition}

We can show that, when the linear operator $\Acal$ satisfies the $\RIP_{\delta}$, we can find a \emph{decoder} that is robust to noise and that will allow us to recover $\Sigmab$ from $\sbf$. More precisely, following standard results in compressed sensing \cite{foucart13,keriven2018instance}, we have the following result (the proof is given in Appendix~\ref{proof:invriptodecoder} for completeness):
\begin{restatable}{theorem}{invripgotodecoder}
\label{theo:invriptodecoder}
Let $\Acal: (S_d(\R), \|\cdot\|_{S_d}) \rightarrow (\R^{m}, \|\cdot\|_{\R^{m}})$ be a linear operator. Suppose that $\Acal$ satisfies $\RIP_{\delta}$ on a model set $\Sfrak \subset S_d(\R)$ and consider the decoder $\D: \R^{m} \rightarrow \Sfrak$ defined by
\begin{equation}
\label{eq:decoder_def}
\D[\sbf] \in \underset{\Sigmab \in \Sfrak}{\arg\min} \ \|\Acal(\Sigmab)-\sbf\|_{\R^m} \,.
\end{equation}
Suppose that $\xbf_1, \cdots , \xbf_n \stackrel{i.i.d}{\sim} \mu$ where $\mu$ is a centered probability distribution with covariance $\Sigmab = \expect{\xbf \sim \mu}{\xbf \xbf^\top} \succ 0$.
Consider $\widehat{\Sigmab} = \frac{1}{n} \sum_{i=1}^{n} \xbf_i \xbf_i^{\top}$ the empirical covariance matrix and $\sbf = \Acal(\widehat{\Sigmab})$. If $\Sigmab \in \Sfrak$, then $\Sigmab^\star \stackrel{\D}{=} \D[\sbf]$ satisfies 
\begin{equation}
\|\Sigmab^\star-\Sigmab\|_{S_d} \leq \frac{2}{1-\delta}\|\Acal(\widehat{\Sigmab})-\Acal(\Sigmab)\|_{\R^m}\,.
\label{eq:decoder_prop}
\end{equation}
When $\Sigmab \not\in \Sfrak$, the bound \eqref{eq:decoder_prop} still holds up to an additional ``modeling error'' term $d^{\circ}(\Sigmab, \Sfrak)$ which quantifies the distance from $\Sigmab$ to $\Sfrak$ (see Appendix \ref{proof:invriptodecoder}).
\end{restatable}
We always assume that the minimization problem \eqref{eq:decoder_def} has at least one solution. The decoder can be adjusted as in \cite{Bourrier}, to handle the case where the $\operatorname{argmin}$ is only approximated to a certain accuracy.
The estimator given by \eqref{eq:decoder_def} has many desirable properties: 1) it is robust to noise as shown in \cite{keriven2018instance,gribonval2020compressive} 2) it allows a recovery of $\Sigmab$ from $\sbf$ as the number of samples $n$ grows. Indeed $\|\Acal(\widehat{\Sigmab})-\Acal(\Sigmab)\|_{\R^m} \leq (\sup_{\|\U\|_{S_d}=1} \|\Acal(\U)\|_{\R^m}) \|\widehat{\Sigmab}-\Sigmab\|_{S_d}$ and $\|\widehat{\Sigmab}-\Sigmab\|_{S_d}$ typically behaves as $\Ocal(n^{-1/2})$ under standard sub-Gaussian assumptions \cite[Section 6]{wainwright_2019}. 
Thus, solutions of the optimization problem \eqref{eq:decoder_def} theoretically yield good approximations of $\Sigmab$. As an additional outcome, by leveraging the regularity of the inverse mapping for matrices with bounded spectra, this estimation of $\Sigmab$ also results in a reliable estimate of $\Thetab$. We refer to the discussion below Theorem \ref{theo:theRIPRO} for a more precise statement.

With these remarks in mind, the strategy is now to obtain this RIP assumption, which enables the information-theoretic decoder and recovery guarantees. By the linearity of $\Acal$, we can observe that obtaining $\RIP_{\delta}$ is equivalent to showing that, for $0 \leq \delta <1$,
\begin{equation}
\label{eq:eq_to_control}
\underset{\U \in S[\Sfrak]}{\sup} \ \big| \|\Acal(\U)\|_{\R^m}-1 \big| < \delta\,,
\end{equation}
where we set
\begin{equation}
S[\Sfrak]\stackrel{\D}{=} \left\{\frac{\Sigmab_1-\Sigmab_2}{\|\Sigmab_1-\Sigmab_2\|_{S_d}} \ : (\Sigmab_1,\Sigmab_2) \in \Sfrak^{2}, \|\Sigmab_1-\Sigmab_2\|_{S_d} >0 \right\}\,.
\label{eq:normalized_sec}
\end{equation}
This set is called the \emph{normalized secant set} of $\Sfrak$ \cite{recipies_rip}. We will provide a more detailed description of the space $S[\Sfrak]$ later on. For now, we present the classical framework that will enable us to prove the $\RIP_{\delta}$ property of $\Acal$.

The operator $\Acal$ being random we will show that, given a sufficient (but reasonable) dimension $m$, we can have a control of type \eqref{eq:eq_to_control} with high probability. In the following we denote by $\Ncal(S[\Sfrak],\|\cdot\|_{S_d},\varepsilon)$ the \emph{covering number} of $S[\Sfrak]$ which, informally, quantifies the effective ``dimension'' of this set (see Section \ref{sec:covering} for more details). The following theorem (whose proof is deferred to Appendix~\ref{proof:rip_cond2}) describes the main ingredients for establishing $\RIP_{\delta}$.

\begin{restatable}{theorem}{ripcondbis}
\label{theo:rip_cond2}
Consider a random sketching operator $\Acal: S_d(\R) \rightarrow \R^{m}$ and denote its operator norm by $\vertiii{\Acal} \stackrel{\D}{=} \underset{\|\U\|_{S_d}=1 }{\sup} \|\Acal (\U)\|_{\R^m}$ . Suppose that  we are given two functions $C_1, C_2 : \R_+ \to \R_+$ such that 
\begin{equation}
\label{concentration_12}
\forall t > 0, \ \Pbb\left(\vertiii{\Acal}> t\right) \leq C_1(t)\,,
\end{equation}
\begin{equation}
\label{concentration_22}
\forall \U \in S[\Sfrak],\  \forall t >0, \ \Pbb\big(\big|\|\Acal (\U) \|_{\R^m} -1\big|> t\big) \leq C_2(t)\,.
\end{equation}
Then, for any $\varepsilon>0$ and $\delta \in [0,1[$,
\begin{equation}
\underset{\U \in S[\Sfrak]}{\sup} \ \big| \|\Acal(\U)\|_{\R^m}-1 \big| < \delta\,,
\label{eq:RIP_on_secant}
\end{equation}
with probability at least $1-\Ncal(S[\Sfrak],\|\cdot\|_{S_d},\varepsilon)C_2(\frac{\delta}{2}) - C_1(\frac{\delta}{2 \varepsilon})$. Consequently with the same probability the operator $\Acal$ satisfies $\RIP_{\delta}$ on $\Sfrak$.
\end{restatable}

\begin{restatable}{remark}{concentrationLink}
\label{rem:C1fromC2}
The above theorem is presented in its most usual form, with both controls \eqref{concentration_12} and \eqref{concentration_22}. However, \eqref{concentration_22} is sometimes easier to obtain (by leveraging classical concentration inequalities). 
Fortunately, we can obtain \eqref{concentration_12} from \eqref{concentration_22}, as for $\varepsilon'>0$, defining $C_1$ by
\begin{equation}
\label{eq:C1_from_C2}
C_1(t) = \Ncal( B_{S_d}, \|\cdot\|_{S_d}, \varepsilon' ) \ C_2( (1-\varepsilon') t - 1) \quad \forall t>1/(1-\varepsilon') \,,
\end{equation}
where $\Ncal( B_{S_d}, \|\cdot\|_{S_d}, \varepsilon' )$ is the covering number of the unit sphere  $B_{S_d} = \{ \U \in S_d : \ \| \U \|_{S_d}=1 \}$, yields a valid upper-bound of $\Pbb\left(\vertiii{\Acal}> t\right)$. See Appendix~\ref{pr:concentrationLink} for the proof.
\end{restatable}

\subsection{Main theoretical result \label{sec:main_results}}

Guided by Theorem \ref{theo:rip_cond2} we need to control the covering number of $S[\Sfrak_{k,a,b}]$ as well as the concentration of the random operator $\Acal$ to obtain the desired $\RIP_{\delta}$ with high probability. To do so, the randomness related to the vectors $\a_1, \cdots, \a_m$ needs to be specified. We consider in this section two probability distributions for the vectors $\a_j \stackrel{\textit{i.i.d.}}{\sim} \Lambda$, one being $\Lambda = \Lambda_{\operatorname{G}} \stackrel{\D}{=} \Ncal(0,\frac{1}{d} \mathbf{I}_{d})$ the multivariate Gaussian distribution with covariance matrix $d^{-1}\mathbf{I}_{d}$  and the other $\Lambda = \Lambda_{\operatorname{U}} \stackrel{\D}{=} \Ucal(\Sbb^{d-1})$ the uniform distribution on the hypersphere. The scaling of the random vectors $\a_j$ is such that they have unit expected square Euclidean norm under both distributions. These distributions are classic choices when it comes to random projections. We envision that our results remain valid for any choice of sub-Gaussian rotational-invariant distribution. Both norms $\|\cdot\|_{S_d}$ and $\|\cdot\|_{\R^m}$ in Theorem \ref{theo:rip_cond2} must also be defined. For the former, we choose a norm that is \emph{specific} to how our random operator $\Acal$ is drawn. Indeed, the probability distribution $\Lambda \in \{\Lambda_{\operatorname{G}}, \Lambda_{\operatorname{U}}\}$ induces a norm on $S_{d}(\R)$ defined by
\begin{equation}
\label{eq:norm_lambda}
\forall \Mbf \in S_{d}(\R), \ \|\Mbf\|_{\Lambda}\stackrel{\D}{=} \E_{\a \sim \Lambda}\left[\left|\langle \a \a^\top, \Mbf \rangle\right| \right] = \E_{\a \sim \Lambda}\left[\left|\a^\top \Mbf \a\right| \right].
\end{equation} 
This choice is adapted to our problem when $\|\cdot\|_{\R^m} = \|\cdot\|_{1}$ since $\expect{\a_j \sim \Lambda}{\|\Acal (\Mbf)\|_1} = \| \Mbf \|_\Lambda$. Therefore, if for every $\U \in S[\Sfrak]$, $\| \Acal(\U) \|_1$ well concentrates around its expectation, then it has a high probability of being close to its expectation $\| \U \|_\Lambda = 1$, and thus \eqref{eq:RIP_on_secant} will hold.

From these choices, we show that with a reasonable value of $m$ the rank-one operators $\Acal$ satisfy the RIP on $\Sfrak_{k,a,b}$. It is formalized in the theorem below which is the main theoretical result of the paper.
We advise readers that are interested in the proof to refer to Section~\ref{sec:cov_and_concentration} where the results on the covering numbers (Section~\ref{sec:covering}) and the concentration of $\Acal$ (Section~\ref{sec:rank_one}) are derived.

\begin{restatable}{theorem}{theRIPRO}
\label{theo:theRIPRO}
Let $\Acal: (S_d(\R), \|\cdot\|_{\Lambda}) \rightarrow (\R^{m}, \|\cdot\|_{1})$ be a rank-one projection operator as defined in \eqref{eq:ROoperator}, with $(\a_j)_j$ either Gaussian or uniform ($\Lambda \in \{\Lambda_{\operatorname{G}}, \Lambda_{\operatorname{U}}\}$). For all $\delta, \rho \in ]0,1[$, there exists $C = C(\delta, \rho, b/a)$, independent of $m, k$ and $d$, such that, whenever 
\begin{equation}
 m \geq C (d+2k) \log d \, ,
\label{eq:m_for_RIP_ROP_quali}
\end{equation}
the operator $\Acal$ satisfies $\RIP_{\delta}$ on $\Sfrak_{k,a,b}$ with probability at least $1-\rho$. 
In particular the following holds uniformly on $\Sigmab \in \Sfrak_{k,a,b}$ with probability at least $1-\rho$:
Let $\xbf_1, \cdots , \xbf_n \sim \mu$ with $\mu$ a centered probability distribution with covariance $\Sigmab$, $\widehat{\Sigmab}$ the empirical covariance matrix and $\sbf = \Acal(\widehat{\Sigmab})$ a sketch of the data. The estimator $\Sigmab^\star=\D[\sbf]$ defined in \eqref{eq:decoder_def} satisfies
\begin{equation}
\label{eq:eq_est_RO}
\|\Sigmab^\star-\Sigmab\|_\Lambda \leq \frac{2}{1-\delta} \|\Acal(\widehat{\Sigmab})-\Acal(\Sigmab)\|_1\,.
\end{equation}
\end{restatable}
A more precise condition on $m$ can be found in the proof of this theorem in Appendix~\ref{proof:theRIPRO} (in particular see \eqref{eq:m_for_RIP_ROP_quanti}). This theorem shows that our sketching operators satisfies the $\RIP$ with high probability with a sketching dimension $m \gtrsim (d+2k) \log d$, which is much smaller than the $d^2$ required to recover a matrix of size $d \times d$. Let us mention that although in \eqref{eq:eq_est_RO} the error between the estimator $\Sigmab^\star$ and the true covariance matrix $\Sigmab$ is measured with the unusual $\Lambda$-norm, we can have the same type of control for $\|\Sigmab^\star-\Sigmab\|_\Fro$, at the expense of a multiplicative constant $1/c_\Fro = \frac{9 \sqrt{15}}{2} d \leq 18 d$ (see Proposition~\ref{prop:lambda_norm_controls}). 
In addition, \eqref{eq:eq_est_RO} also provides guarantees for the recovery of the precision matrix~$\Thetab$. By making use of the bounded spectra of $(\Sigmab^\star)^{-1}$ and $\Thetab$, we can exploit the regularity of the inverse map to obtain $\| (\Sigmab^\star)^{-1} - \Thetab \|_\Fro \leq b^2 \|\Sigmab^\star-\Sigmab\|_\Fro\,$. Hence, whenever \eqref{eq:eq_est_RO} is verified, we also have 
\begin{equation}
\label{eq:sigma2theta}
\| (\Sigmab^\star)^{-1} - \Thetab \|_\Fro \leq \frac{9 \sqrt{15} \ b^2 d}{(1-\delta)} \|\Acal(\widehat{\Sigmab})-\Acal(\Sigmab)\|_1\,. 
\end{equation}

\begin{remark}[A similar result with an unbounded model set.]
We want to point out that in \eqref{eq:m_for_RIP_ROP_quanti}, the condition on $m$  depends on $a$ and $b$ only through the ratio $b/a$. This might suggest that the fundamental quantity to consider is the ratio between the largest and smallest eigenvalues, \textit{i.e.,} the condition number of the matrix, instead of the eigenvalues themselves. In Appendix~\ref{sec:covering_cond_nb}, we introduce a model set $\Sfrak_{k, \kappa_0}$, where precision matrices have a condition number bounded by some $\kappa_0\geq1$. Even though $\Sfrak_{k, \kappa_0}$ is much ‘‘bigger'' than $\Sfrak_{k, a, b}$, as the latter is bounded and the former is not, we obtain an upper-bound on the covering number of its normalized secant $S[\Sfrak_{k, \kappa_0}]$ (see Theorem~\ref{theo:big_theo_covering_cond_nb} in Appendix~\ref{sec:covering_cond_nb}). 
This is sufficient to derive a result similar to Theorem~\ref{theo:theRIPRO} with the same $m \gtrsim (d+2k) \log d$ condition, yielding guarantees on the recovery of $\Sigmab$ with only bounded condition numbers rather than bounded spectra. 
\end{remark}

\subsection{Connection to prior works}

Theorem~\ref{theo:theRIPRO} indicates that it is theoretically possible to recover $\Thetab$, with a $\Ocal(n^{-1/2})$ error, by keeping in memory only a single sketch of size $m \gtrsim (d+2k)\log(d)$. To the best of our knowledge this is the first result for compressive recovery of sparse precision matrices from empirical covariance measurements. It can however be put in perspective with \cite{gabor} which tries to find the support of $\Thetab$ by observing, in an adaptive manner, only a small fraction of the entries of the \emph{true} covariance $\Sigmab$. The authors show that it is possible to find $\Thetab$ from $\Ocal(d \operatorname{polylog}(d))$ elements of $\Sigmab$ with some assumptions on the graph (small treewidth). Interestingly enough, we are able to obtain the same type of guarantees but with the big difference that, in our case, we observe a non-adaptive compressed version of the empirical covariance that takes the whole matrix into account, which is more in line with concrete applications.

\subsection{Practical limitations of Theorem~\ref{theo:theRIPRO}}

Theorem~\ref{theo:theRIPRO} indicates that the sketch contains the necessary information to recover the true covariance matrix using the decoder 
\begin{equation}
\D[\sbf] \in \underset{\Sigmab \in \Sfrak_{k,a,b}}{\arg\min} \ \|\Acal(\Sigmab)-\sbf\|_{\R^m}\,.
\label{probleme_optim}
\end{equation}
However, even though this theorem holds theoretical significance, its practical application encounters two significant challenges when employing the decoder \eqref{probleme_optim} and the sketching operator with \emph{independent} rank-one projections.
\begin{itemize}
\item[$\bullet$] To retrieve $\Thetab$ from the sketch, one needs to use $\Acal$ and thus to store the $m$ $d$-dimensional vectors $(\mathbf{a}_j)_{j \in \integ{m}}$. Hence, the overall memory cost comprises a reasonable $\mathcal{O}(m)$ expense to store the sketch $\sbf$ but also an additional $\mathcal{O}(md)$ cost to store the $(\mathbf{a}_j)_{j \in \integ{m}}$. Unfortunately, according to Theorem~\ref{theo:theRIPRO}, the sketch size must be $m \gtrsim d \log(d)$, yielding a total memory cost larger than $\Ocal(d^2)$, similar to the cost of storing the empirical covariance matrix.

\item[$\bullet$] Directly solving the optimization problem \eqref{probleme_optim} is probably intractable as the constraint $\Sigmab \in \Sfrak_{k,a,b}$ implies to search among the matrices $\Sigmab$ such that $\|\Sigmab^{-1}\|_0 \leq (d+2k)$ which is a highly non-convex constraint.
\end{itemize}

\section{Towards practical recovery \label{sec:optimization}}

As a remedy to the above-mentioned issues, we propose in this section to use structured matrices for efficient sketching and a tractable approximate decoder as a more practical alternative to \eqref{probleme_optim}.

\subsection{Structured matrices for efficient sketching \label{sec:structured}}

Using structured random matrices is a recurrent idea in compressive learning \cite{le2013fastfood, orthogonal_RF, chatalichal01701121}. It reduces the degrees of freedom while mimicking the behavior of random matrices with \textit{i.i.d.} columns. In the following, we assume that $d = 2^K$ is a power of 2 and that $m = B \times d$ is a multiple of $d$. Padding strategies can be implemented when these requirements are not met, but we leave this technicality out of the scope of this paper for the sake of simplicity and refer the interested reader to \cite{chatalic:tel-03023287}.
Here, we adopt the same approach as \cite{orthogonal_RF,chatalichal01701121}. Instead of independent random vectors, we define $(\a_1, \dots \a_m)$ as the columns of the random matrix $\Abf = (\Bbf_1 | \dots | \Bbf_B) \in \R^{d \times m}$ made of $B$ independent structured blocks $\Bbf_l \in \R^{d \times d}$ defined as triple-Rademacher matrices:
\begin{equation}
\Bbf_l \stackrel{\D}{=} \frac{1}{d^{3/2}} \mathbf{H} \Dbf_l^{(1)} \mathbf{H} \Dbf_l^{(2)} \mathbf{H} \Dbf_l^{(3)}.
\end{equation}
The matrix $\mathbf{H}$ denotes the Walsh-Hadamard matrix with entries in $\{\pm 1\}$ and $\Dbf_{l}^{(k)}$ are independent random diagonal ‘‘sign-flipping'' matrices, \textit{i.e.,} random diagonal matrices with independent Rademacher entries. The scaling factors $d^{3/2}$ yields $\| \a_j \|_2 = 1$. We emphasize that this strategy results in the use of \emph{dependent} random vectors $\a_1, \cdots, \a_m$.

This approach has a dual advantage: it improves both memory and computational efficiency. Indeed, notice that $\Phi(\xbf)$ can be expressed as $\Phi(\xbf) = \{(\Abf^\top \xbf)_j^2\}_{j \in \integ{m}} = (\Abf^\top \xbf) \odot (\Abf^\top \xbf)$ where $\odot$ denotes the Hadamard product. Thus the memory and computational bottleneck \textit{a priori} lies in the storage of $\Bbf_l$ and the computations of $\Bbf_l^\top \xbf$ for $l \in \integ{B}$. 
Fortunately, we benefit from the properties of the fast Walsh-Hadamard transform: 1) it computes $\mathbf{H}^\top \ybf$ for any vector $\ybf$ without storing $\mathbf{H}$ as it is hard-coded into the algorithm, 2) the matrix-vector multiplication $\mathbf{H}^\top \ybf$ requires only $\Ocal(d \log d)$ operations. 
As a consequence for our sketching mechanism, only the diagonal matrices $\Dbf_{l}^{(k)}$ must be stored. This reduces the space complexity of storing the dense matrix $\Abf \in \R^{d \times m}$ from $\Ocal(md)$ to $\Ocal(m)$ \cite{orthogonal_RF,Fino}. 
Moreover, point 2) implies that $\Phi(\xbf)$ can be computed with a time complexity of $\Ocal(m \log d)$ as each of the $B$ matrix-vector multiplications $\Bbf_l^\top \xbf$ requires $\Ocal(d \log d)$ operations with the fast Walsh-Hadamard transform. 

Theoretical examination of recovery guarantees for these structured sketching operators is deferred to future research. To achieve the same reconstruction guarantees as in the independent case, the key lies in verifying the concentration of $\Acal$ in the structured case, which amounts to proving a result similar to Proposition~\ref{prop:C2forROP}. We expect that guarantees comparable to those in the independent case can be obtained.

\begin{algorithm}[t]
    \caption{\label{alg:decoder_based_on_glasso} Algorithm for solving the decoding problem}
    \begin{algorithmic}[1]
    \State Input: Sketching operator $\Acal$ and sketch of the data $\sbf = \frac{1}{n} \sum_{i=1}^{n} \Phi(\xbf_i) = \Acal(\widehat \Sigmab) \in \R^{m}$.  
    \State Initial guess $\Sigmab_{0} \succ 0$, regularization parameter $\lambda > 0$, step size $\gamma >0$, and maximum number of iterations $t_{\max}$.
     \For{$t \in 0, 1 \cdots, t_{\max}-1$}
     \State $\Sigmab_{t+\frac{1}{2}} \leftarrow \Sigmab_t- \gamma \Acal^{\star}(\Acal(\Sigmab_t)-\sbf)$
     \State $\Sigmab_{t+1} \leftarrow \operatorname{GLASSO}_{\lambda\gamma}[\Sigmab_{t+\frac{1}{2}}]$ \text{ // with standard graphical lasso solver}
    \EndFor \\
    \Return estimated covariance $\Sigmab_{t_{\max}}$ and precision $\Sigmab_{t_{\max}}^{-1}$.
     \end{algorithmic}
\end{algorithm}

\subsection{Algorithmic solution to decoding \label{sec:algorithmic_sol}}

Finally, we present a heuristic algorithmic approach to obtain the precision matrix from a sketch of the data. We will demonstrate that a computationally simpler alternative decoder to \eqref{probleme_optim} works effectively in practice.

\paragraph{The graphical lasso as a denoiser} Our algorithmic solution is inspired by the connections between proximal operators and denoisers in the context of inverse problems. Numerous studies have indeed demonstrated that proximal operators can be regarded as efficient denoisers \cite{hurault2022proximal, hurault2023convergent, cohen2021has,cohen2021regularization}. The core concept behind our approach thus revolves around utilizing a decoder that relies on the graphical lasso \eqref{eq:glasso}, which not only benefits from efficient algorithms but also carries the interpretation of a proximal operator \cite{Beck}. Subsequently, we introduce
\begin{equation}
\label{eq:glasso_operator}
\operatorname{GLASSO}_{\lambda}[\Z] \stackrel{\D}{=} \underset{\Sigmab \succ 0}{\operatorname{\arg\min}} \ -\log\det \Sigmab^{-1} + \langle \Z, \Sigmab^{-1}\rangle + \lambda \|\Sigmab^{-1}\|_{1, \operatorname{off}}\,,
\end{equation}
where we recall that $\|\Mbf\|_{1, \operatorname{off}}=\sum_{i<j} |M_{ij}|$. This operator is equivalent to the one introduced in \eqref{eq:glasso}, which computes the precision matrix instead of the covariance matrix. Indeed, although \eqref{eq:glasso_operator} is non-convex, a solution can be computed by solving the convex graphical lasso problem \eqref{eq:glasso} with the change of variable $\Thetab = \Sigmab^{-1}$. We also emphasize that most graphical lasso solvers such as \cite{friedman_sparse_2008,Banerjee, Rahul_new_perspective} compute both the covariance $\Sigmab$ and the precision $\Thetab = \Sigmab^{-1}$, without calculating the inverse, by relying instead on duality theory.  We argue that the operator \eqref{eq:glasso_operator}, when applied to the empirical covariance of the data $\widehat \Sigmab$,  can be interpreted as a ‘‘denoiser'' of $\widehat \Sigmab$ in the sense that it returns a covariance matrix whose inverse is sparse. This interpretation is motivated by the fact that the graphical lasso can be seen as a specific proximal operator in the framework of \emph{Bregman divergences} \cite{bregman1967relaxation,censor1992proximal} which we now briefly describe (we refer to \cite{teboulle2018simplified} for a more precise discussion). Let $\Hcal$ be a Hilbert space and $h: \Hcal \rightarrow \R \cup \{+\infty\}$ be proper, convex and differentiable on its open domain $\operatorname{dom}(h)$. The Bregman divergence associated to $h$ is given by
\begin{equation}
\forall \xbf,\ybf \in \Hcal \times \Hcal, D_{h}(\xbf | \ybf) = \begin{cases} &h(\xbf)-h(\ybf)-\langle \nabla h(\ybf), \xbf-\ybf\rangle \text{ if } \ybf \in \operatorname{dom}(h)\,, \\ & +\infty \text{ otherwise.}\end{cases}
\end{equation}
The function $D_h$ measures a similarity or ‘‘distance'' between the points in $\Hcal$. For instance if $\Hcal = \R^{d}$ and $h = \frac{1}{2} \|\cdot\|_2^2$ then $D_{h}$ is simply $D_{h}(\xbf | \ybf) = \frac{1}{2} \|\xbf-\ybf\|_2^2$. For a function $\varphi: \Hcal \rightarrow \R$, the so-called \emph{(left) Bregman proximal operator} of $\varphi$ is defined as $\operatorname{prox}_{\varphi}^{h}(\zbf) \stackrel{\D}{=} \arg\min_{\xbf} \ \varphi(\xbf) + D_{h}(\xbf|\zbf)$ \cite[Definition 2.3]{teboulle2018simplified}. It generalizes the standard Euclidean proximal operator that can be computed with $h = \frac{1}{2} \|\cdot\|_2^2$. To relate with the context of the graphical lasso, we introduce the function $h(\X) = -\log\det\X$ if $\X \succ 0$ and $h(\X) = +\infty$ otherwise. It is strictly convex, continuously differentiable over its domain, and $\nabla h(\X) = -\X^{-1}$  \cite[Appendix A.4.1]{cboyd}. The associated Bregman divergence writes (see \textit{e.g.} \cite{Ravikumar, benfenati2020proximal})
\begin{equation}
\label{eq:log_det_breg}
\forall \X, \Y \in S_{d}^{++}(\R), \ D_{h}(\X|\Y) \stackrel{\D}{=} -\log\det(\X)+\log\det \Y+\langle \Y^{-1}, \X\rangle -d\,.
\end{equation}
Based on the definition of $D_h$ we can rewrite the operator \eqref{eq:glasso_operator} as $\operatorname{GLASSO}_{\lambda}[\Z] = \underset{\Sigmab \succ 0}{\arg\min} \  \lambda \|\Sigmab^{-1}\|_{1, \operatorname{off}} + D_{h}(\Z|\Sigmab)$. The graphical lasso can thus be interpreted as a proximal Bregman operator of the function $\varphi: \Sigmab \mapsto \lambda \|\Sigmab^{-1}\|_{1, \operatorname{off}}$ but by operating on the \emph{right variable} of the Bregman divergence. Although not previously studied with the graphical lasso, this type of operators, known as \emph{right proximity operator} \cite{bauschke2006joint, bauschke2018regularizing, lau2022bregman}, has nevertheless been considered in the context of Poisson inverse problems \cite{bauschke2017descent}, for image restoration problems \cite[Section 4]{woo2017characterization} or in \cite{gribonval2020characterization} where it was shown to admit a characterization in terms of gradient of a convex function. This discussion highlights that $\operatorname{GLASSO}_{\lambda}[\widehat{\Sigmab}]$ computes a covariance matrix candidate which is both close to $\widehat{\Sigmab}$ in the sense of $D_h$ and whose inverse tends to be sparse.
\paragraph{Algorithm} Inspired by this interpretation of the graphical lasso as a denoiser, we present an iterative algorithm for estimating the true covariance from the sketch. In the following, we consider the data fidelity term $f(\Sigmab) \stackrel{\D}{=} \frac{1}{2} \|\Acal(\Sigmab) - \sbf\|_2^2$ whose gradient is $\nabla f(\Sigmab) = \Acal^{\star}(\Acal(\Sigmab)-\sbf)$ where 
\begin{equation}
\label{eq:adjoint}
\Acal^{\star}: \ybf \in \R^{m} \mapsto \frac{1}{m} \sum_{j=1}^{m} y_i \a_j \a_j^{\top} \in S_d 
\end{equation} is the adjoint operator of $\Acal$. Given a initial estimate $\Sigmab_0 \succ 0$ and a step-size  $\gamma > 0$, our proposed algorithmic solution computes
\begin{equation}
\label{eq:_algorithm_iter}
\forall t \in \{0, \cdots, t_{\max}\}, \ \Sigmab_{t+1} = \operatorname{GLASSO}_{\gamma \lambda}[\Sigmab_t- \gamma \nabla f(\Sigmab_t)] \,.
\end{equation}
In other words, we alternate between a gradient step $\Sigmab_{t+\frac{1}{2}} = \Sigmab_t- \gamma \nabla f(\Sigmab_t)$ in the direction of minimizing $f$ and a denoising/proximal Bregman step $\Sigmab_{t+1} = \operatorname{GLASSO}_{\gamma \lambda}[\Sigmab_{t+\frac{1}{2}}]$ (with parameter $\gamma \lambda$) in the vein of standard forward-backward algorithms. The overall procedure is summarized in Algorithm \ref{alg:decoder_based_on_glasso}. The computational bottleneck of this algorithm is the graphical lasso step that has a computational $\Ocal(d^{3})$ complexity with standard graphical lasso solvers such as \cite{friedman_sparse_2008,Banerjee, Rahul_new_perspective}, which rely on the dual formulation of the graphical lasso, or \cite{rolfs2012iterative} which rely on an iterative thresholding procedure. Although more efficient solvers exist \cite{ChoQUIC, bigQUIC, squic} we choose to solve graphical lasso with the \texttt{scikit-learn} implementation \cite{scikitlearn}. It relies on the block coordinate procedure described in \cite{Rahul_new_perspective} and finds the solution of the graphical lasso with cubic complexity. It is noteworthy that the iterations of our algorithm, as described in \eqref{eq:_algorithm_iter}, can be linked to Bregman Proximal Gradient (BPG) descent \cite{bauschke2017descent} with the Bregman divergence $D_h$. In Appendix \ref{sec:connections}, we demonstrate that the iterations in \eqref{eq:_algorithm_iter} correspond to those of a BPG algorithm, albeit with a Riemannian gradient instead of the usual Euclidean gradient in \eqref{eq:_algorithm_iter}. In practice, we observe that this algorithm always converges when $\gamma > 0$ is sufficiently small to ensure that the iterates $(\Sigmab_{t+\frac{1}{2}})_t$ remain positive definite (we give a safe step-size strategy ensuring this condition in Appendix \ref{sec:safe_step}). The intriguing questions regarding the convergence rate of this algorithm and the minimizers associated with it are left for future research. We envision the use of guarantees inspired by Plug-and-Play literature, as described \textit{e.g.} in \cite{ryu2019plug}.  However, we experimentally demonstrate in the next section that the associated estimator effectively recovers a precision matrix from a sketch of the data.

\section{Experiments \label{sec:experiments}}

In this section we provide experiments for assessing the efficiency of Algorithm \ref{alg:decoder_based_on_glasso}. The following experiments are conducted using \emph{structured sketching} as described in Section \ref{sec:structured}. The objectives of these experiments is to answer the following questions: 
\begin{enumerate}
\item Does the decoder described in Algorithm \ref{alg:decoder_based_on_glasso} give qualitatively coherent results?
\item What is the impact of the sketch size $m$ on the final estimation? More precisely, can we obtain a good estimate of the true precision matrix with a number of measurements $m$ close to the theoretical $m_0 = C (d+2k) \log(d)$ obtained in Theorem~\ref{theo:theRIPRO}?
\item How does the sketching approach combined with the decoder described in Algorithm \ref{alg:decoder_based_on_glasso} compare to classical methods such as the graphical lasso in terms of performance?
\end{enumerate}
In all experiments, the setting is as follows: we generate a true sparse precision matrix $\Thetab \in S_{d}^{++}(\R)$ and we consider $\xbf_1, \cdots, \xbf_n \sim \Ncal(0, \Thetab^{-1})$ \ie\ $n$ i.i.d. samples of a multivariate Gaussian with covariance $\Sigmab = \Thetab^{-1}$. In the experiments, we explore two methods for generating sparse precision matrices $\Thetab$. For each method, the matrix $\Thetab$ first consists of $L$ blocks, each having a size of $M \times M$ with $L \times M = d$. The generation of each block follows the distribution of a random graph.  In the first method, referred to as \texttt{Erdos}, each block follows the distribution of an Erd\"{o}s-R\'{e}nyi graph \cite{erdos59a} where the probability of connection is set to $p = 0.2$. In the second method, referred to as \texttt{PowerLaw}, the random graph is a tree with a power-law degree distribution. Both methods utilize the \texttt{Networkx} library \cite{hagberg2008exploring} for generating these distributions. After fixing the support with the above strategy, the value of each coefficient is arbitrarily set to $\varepsilon u$ where $u \sim \operatorname{Unif}[1, 4]$ and $\varepsilon = 1$ with probability $0.5$ and $-1$ with probability $0.5$. The matrix $\Thetab$ is then symmetrized and made positive definite by adding a sufficiently large diagonal $(0.1+\lambda_{\min})\mathbf{I}$. Finally a random permutation permutes the rows and columns of the matrix.
Two examples of matrices $\Thetab$ generated according to these procedures are shown on the left side of Figure \ref{one_example} (for $d=64$).

In the experiments, we consider two performance measures. The first one is the relative error computed as 
\begin{equation}
\label{eq:relative_error}
\operatorname{RE} \stackrel{\D}{=} \frac{\|\Thetab_{\operatorname{true}}-\Thetab_{\operatorname{esti}}\|_{\Fro}}{\|\Thetab_{\operatorname{true}}\|_{\Fro}}
\end{equation}
and the second one is the $F_1 \in [0,1]$ score between the true matrix and the estimated one. It is calculated as $F_1 = \frac{2 \operatorname{tp}}{2 \operatorname{tp}+ \operatorname{fp}+ \operatorname{fn}}$ where true positive $\operatorname{tp}$ stands for the case when there is an actual edge and the algorithm detects it; false positive $\operatorname{fp}$ stands for the case when there is no actual edge but the algorithm detects one, and false negative $\operatorname{fn}$ stands for the case when the algorithm failed to detect an actual edge.

\subsection{First illustration}

\begin{figure}[t]
\begin{center}
  \includegraphics[width=0.9\linewidth]{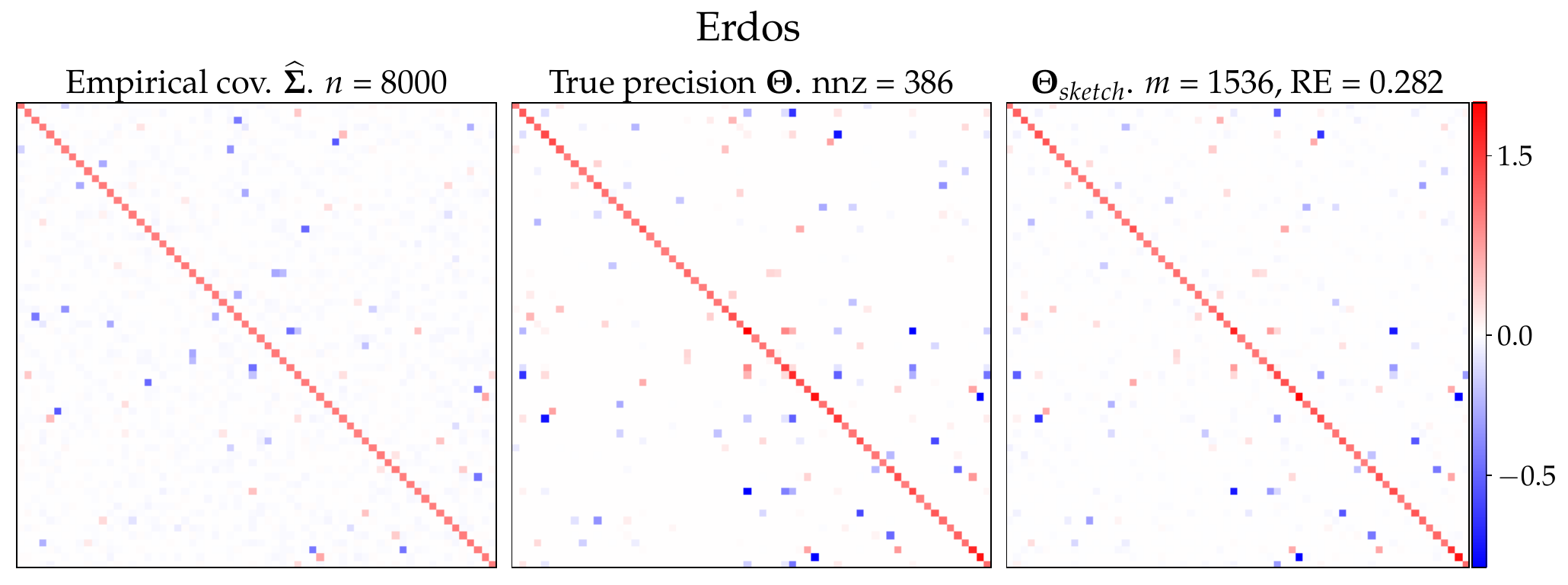}
  \includegraphics[width=0.9\linewidth]{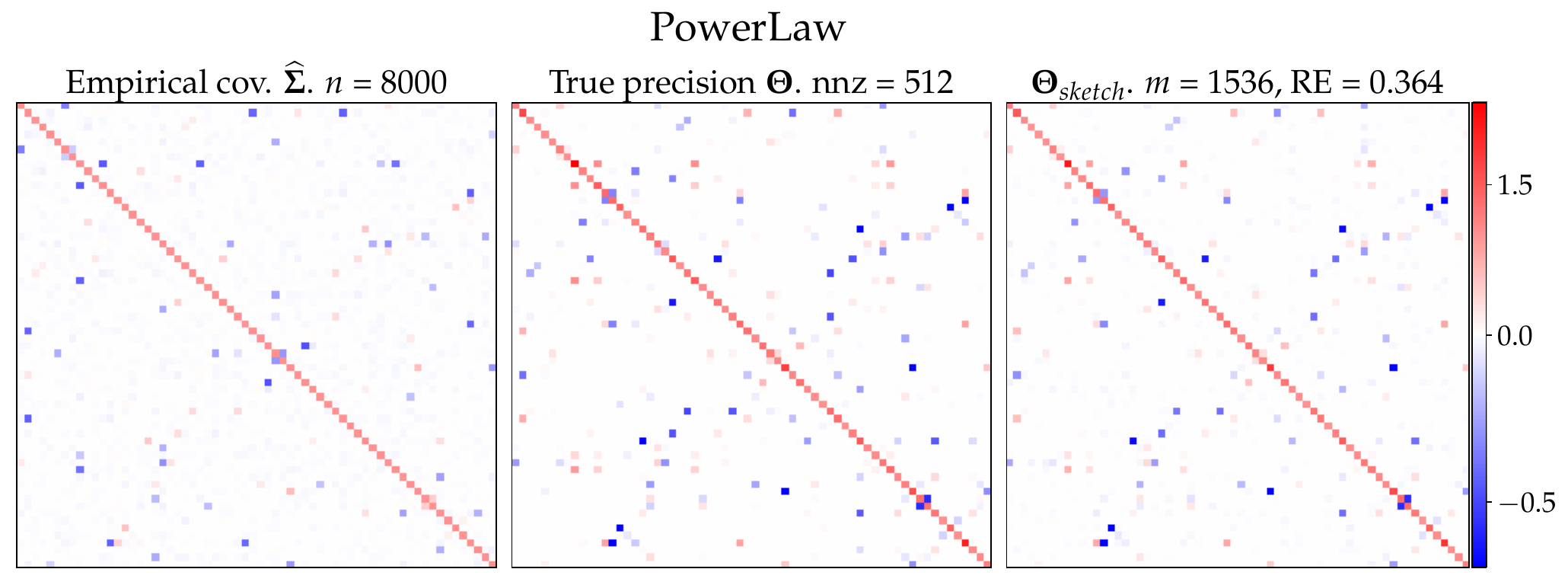}
\end{center}
\caption{Illustrative example of estimation using the decoding procedure described in Algorithm \ref{alg:decoder_based_on_glasso}. The dimension is $d = 64$ and the sketch size $m=1536$. We set $\lambda = 0.008$ for the $\ell_1$ penalty parameter for all solvers. RE stands for the relative error between the true precision matrix and the estimated one \eqref{eq:relative_error}. \textbf{(First row)} With a precision matrix $\Thetab$ generated from the \texttt{Erdos} process. \textbf{(Second row)} With a matrix $\Thetab$ generated from the \texttt{Powerlaw} process. The columns are (from left to right): the empirical covariance $\widehat{\Sigmab}$ for $n = 8000$, the true precision matrix $\Thetab$ and the decoder Algorithm \ref{alg:decoder_based_on_glasso} based on a sketch of the data with structured rank-one projections. \label{one_example}}
 \end{figure}

\begin{figure}[t]
\begin{center}
  \includegraphics[width=\linewidth]{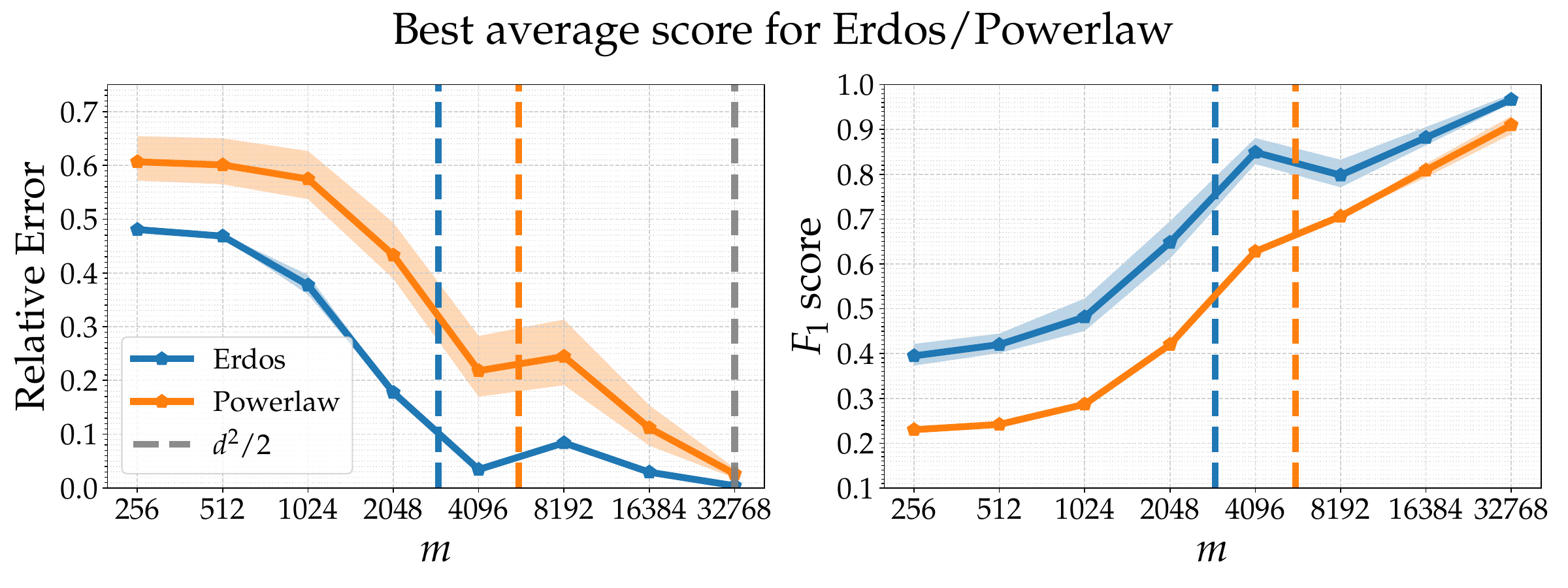}
\end{center}
\caption{\label{fig:impact_m} Impact the of number of measurements on the final estimation for the datasets \texttt{Erdos} and \texttt{PowerLaw}. \textbf{(Left)} the best average relative relative error on the three draws of the true precision matrix for the \texttt{PowerLaw} and \texttt{Erdos} settings. \textbf{(Right)} the best average $F_1$ score. The $10$-th percentile for these scores are reported in shaded line. Vertical colored dashed lines are located at $\hat{k}\log(d)$ where $\hat{k}$ is the average number of non-zero elements $\|\Thetab\|_0$ of the precision matrices for each setting. }
\end{figure}

\begin{figure}[t]
\begin{center}
   \includegraphics[width=\linewidth]{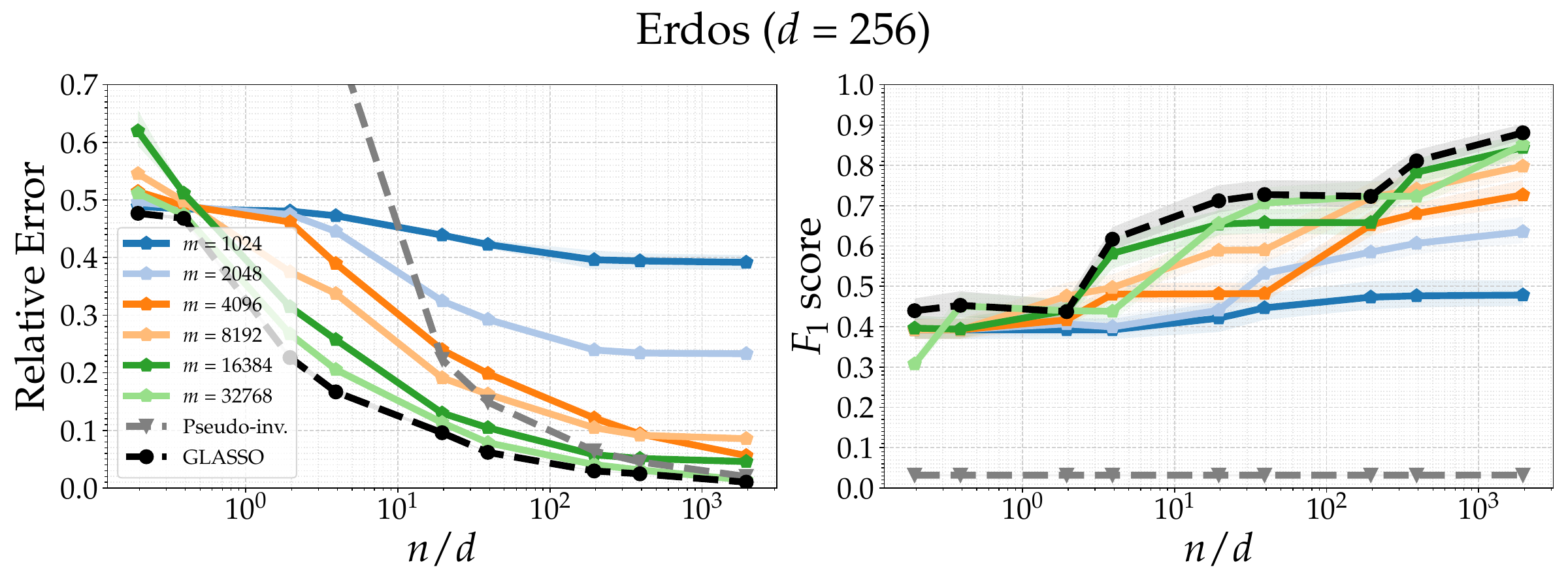}
  \includegraphics[width=\linewidth]{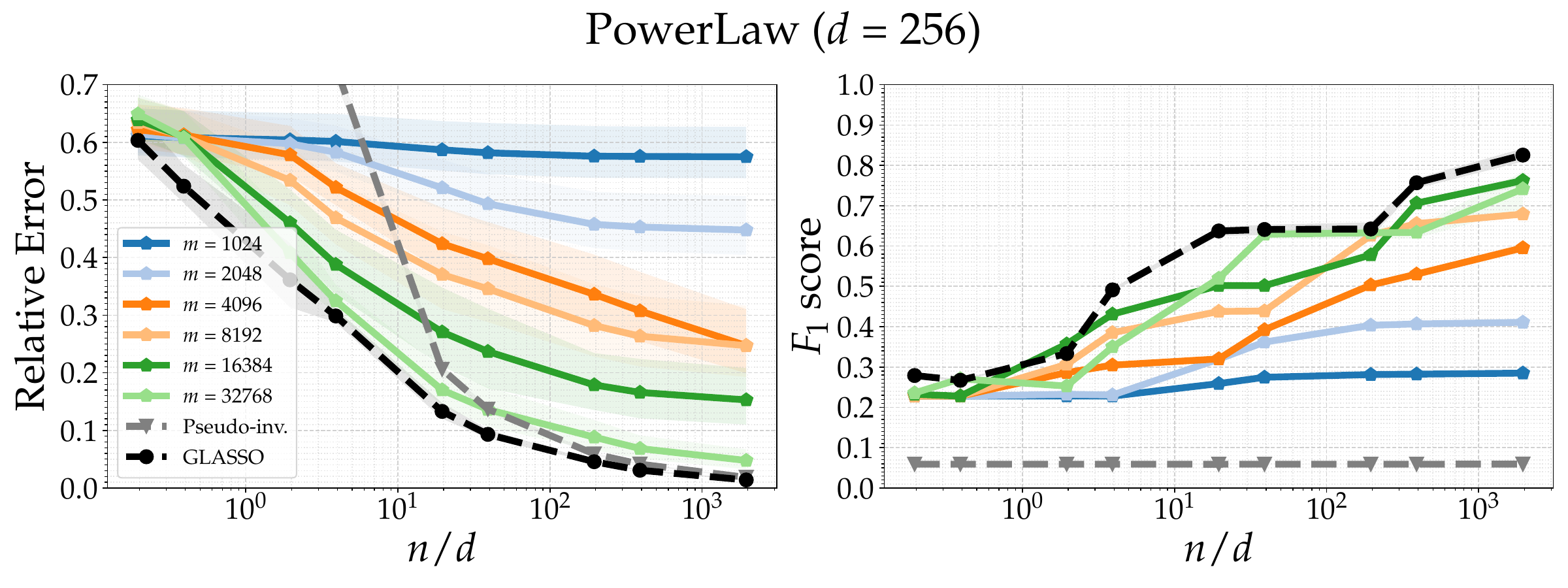}
\end{center}
\caption{\label{error_plot} Comparison between our estimator and graphical lasso type estimators with the \texttt{Erdos} \textbf{(top row)} and \texttt{PowerLaw} settings \textbf{(bottom row)}. The $10$-th percentile for these scores are reported in shaded line. }
\end{figure}

First, we qualitatively illustrate the behavior of our Algorithm \ref{alg:decoder_based_on_glasso}. In this experiment, we set $d = 64$ and generate $n = 8000$ samples from $\Thetab$ using the \texttt{Erdos} and \texttt{Powerlaw} procedures. The number of non-zero elements are respectively $512$ for \texttt{PowerLaw} and $386$ for \texttt{Erdos}. We consider one draw of structured sketching operator as described in Section~\ref{sec:structured}. We compute a sketch of the dataset $\sbf = \frac{1}{n} \sum_{i=1}^{n} \Phi(\xbf_i) = \Acal(\widehat{\Sigmab}) \in \R^{m}$ and set the number of measurements for the sketch to $m = 1536$. In other words, the final sketch has a number of measurements that is approximately $75\%$ the degrees of freedom of $d \times d$ symmetric matrices, which is $d(d+1)/2 = 2080$. We fix the number of iterations to $t_{\max} = 3500$,  the step size to $\gamma = 0.005$ and $\lambda = 0.008$. The results are depicted in Figure \ref{one_example}. We can observe that in both cases, our method provides a visually consistent estimation of the precision matrix.

\subsection{Impact of the number of measurements on the estimation - asymptotic regime }

In order to quantitatively evaluate the impact of the number of measurements $m$ on the final estimation we consider the asymptotic regime where $n = +\infty$ or equivalently we sketch the true covariance matrix as $\sbf = \Acal(\Sigmab)$ instead of the empirical covariance matrix. We take $d= 256$ and we draw three precision matrices from the settings \texttt{PowerLaw} and \texttt{Erdos}. For each setting \texttt{PowerLaw} and \texttt{Erdos}, each $m \in \{256, 512, 1024, 2048, 4096, 8192, 16384, 32768 = d^{2}/2\}$, and each score function ($\operatorname{RE}$ and $F_1$-score) we choose the parameter $\lambda \in \{1\mathrm{e}-4, 5\mathrm{e}-4, 1\mathrm{e}-3, \cdots, 5\mathrm{e}-1\}$ that gives the best average score. We report the results in Figure \ref{fig:impact_m}. 

We observe that for \texttt{Erdos} the estimator based on the sketching procedure and Algorithm \ref{alg:decoder_based_on_glasso} gives a recovery with a relative error that is below $10\%$ starting from $m \approx 4096$ that is $\frac{m}{d^2/2} \approx 12\%$. This corresponds to a compression rate of approximately $88\%$. This result is also consistent with the theoretical bound $m \approx \|\Thetab\|_0 \log(d)$ (represented by the vertical dashed lines) found in Theorem~\ref{theo:theRIPRO}: we can see that the best average relative error is below $10\%$ starting from this limit. From $\frac{m}{d^2/2} = 50\%$, the recovery is also nearly perfect. The $F_1$ score is in agreement with the relative error: starting from $m \approx 4096$, the $F_1$ score is above $0.8$, indicating that our estimator captures the correct statistical relationships between the variables.

For the \texttt{PowerLaw} setting, the results are slightly less favorable, and a relative error below $10\%$ is only achieved for $\frac{m}{d^2/2} \approx 50\%$, resulting in a compression rate of approximately $50\%$. This can be explained by the fact that the precision matrices in this case are less sparse (vertical dashed orange line), and the theory also involves constants (\textit{e.g.,} the constant $C(\delta, \rho, b/a)$ in Theorem~\ref{theo:theRIPRO}) that can have a significant impact on the actual number of measurements required for reconstruction.

In all cases, this experiment indicates that the proposed sketching method preserves information, and the decoding method allows for a good estimation in an optimistic scenario with a large number of samples and a optimally-chosen regularization parameter.

\subsection{Comparison with graphical lasso type estimators - finite sample regime}

In a last experiment we compare quantitatively our approach with the graphical lasso estimator (computed with the \texttt{Scikit-learn} implementation \cite{scikitlearn}) and we investigate the influence of the sample size on the final estimation. We consider $d = 256$ and $n$ samples $\xbf_1, \cdots, \xbf_n \sim \Ncal(0, \Thetab^{-1})$ with three draws of precision matrices for each \texttt{Erdos} and \texttt{Powerlaw} settings. We use a sketch of the data for a number of samples $n$ varying from $50$ (in order to have the small sample regime) to $500000$. We compute the sketch of the data for sketch size $m \in \{1024, 2048, 4096, 8192, 16384, 32768 = d^{2}/2\}$. As in the previous experiment, for each method, each $n$ and each score function we pick the regularization parameter $\lambda \in \{1\mathrm{e}-4, 5\mathrm{e}-4, 1\mathrm{e}-3, \cdots, 5\mathrm{e}-1)\}$ that leads to the best average score.

We also consider a simple baseline, namely the pseudo-inverse of the empirical covariance matrix $(\widehat{\Sigmab})^{\dagger}$ as an estimate for $\Thetab_{\operatorname{true}}$. Note that when $n > d$ the matrix $\widehat{\Sigmab}$ is almost surely invertible to that this baseline corresponds to the maximum likelihood estimator of $\Thetab_{\operatorname{true}}$ without $\ell_1$ regularization (same as graphical lasso with $\lambda =0$ in this case). We report these results in Figure \ref{error_plot}.

In both settings, the results show that our estimator improves as $m$ and $n$ increase, as expected. Moreover, for $m = 32768 = d^{2}/2$, the results of the graphical lasso and our estimator are nearly identical, demonstrating that our estimator is consistent with the graphical lasso when the number of measurements reaches the number of degrees of freedom of the empirical covariance matrix. Furthermore, we notice that for the \texttt{Erdos} setting, the performances are very similar to those of the graphical lasso even for $m \approx 8192$. This result indicates that even in the case of a finite number of samples, we are able to accurately estimate the precision matrices with a limited number of measurements ($8192$ corresponds to a compression rate of $\approx 75\%$). For the \texttt{PowerLaw} setting, the results in terms of relative error are more mixed. However, the $F_1$ score remains comparable to that of the graphical lasso, indicating that our approach captures the correct statistical dependencies, but may struggle to accurately estimate the intensity of these dependencies. In a rather reassuring way, in the small sample regime, our estimator outperforms the MLE estimator $(\widehat{\Sigmab})^{\dagger}$ in terms of relative error and consistently outperforms it in terms of the $F_1$ score (this is reasonable because $(\widehat{\Sigmab})^{\dagger}$ unlike $\Thetab_{\operatorname{true}}$).

\section{Covering numbers bounds and concentration inequalities for Theorem~\ref{theo:theRIPRO} \label{sec:cov_and_concentration}}

In this section, we provide the necessary ingredients to prove Theorem~\ref{theo:theRIPRO}. Guided by Theorem~\ref{theo:rip_cond2}, we start by giving general results to control covering numbers before applying them to the one considered in this article. Then, we provide the concentration inequality needed on the sketching operator $\Acal$.

\subsection{Controlling covering numbers \label{sec:covering}}
 We take the general point of view of normed vector  spaces $E,F$ with norms $\|\cdot\|_E, \|\cdot\|_F$ and a function $f: \Omega \subseteq E \rightarrow F$ defined on some domain $\Omega= \dom(f)$ of $E$. For a subset $\Xfrak \subseteq E$ the covering number of $\Xfrak$ \textit{w.r.t.} $\|\cdot\|_E$ with radius $\varepsilon>0$, denoted by $\Ncal(\Xfrak, \|\cdot\|_E, \varepsilon)$, is the minimal number of closed balls of radius $\varepsilon$ (\textit{w.r.t.} $\|\cdot\|_E$) required to entirely cover $\Xfrak$ and whose centers are in $\Xfrak$ (see Appendix \ref{sec:generality_covering_numbers} for a formal definition).  
\begin{remark}[Link with the precision matrix estimation problem]
This section is presented in a general case, but its results will ultimately be applied to control the covering number of the normalized secant of our model set $\Sfrak_{k,a,b}$ defined in \eqref{eq:model_set}.  Therefore, one should keep in mind that our final application framework will consider $E = F = S_d$, $f=\inv$ with $\Omega$ the set of symmetric and invertible matrices and $\Xfrak = \Sfrak_{k,a,b}^{-1} = \{ \Thetab = \Sigmab^{-1}: \ \Sigmab \in  \Sfrak_{k,a,b}\}$.
\end{remark} 
The core of our reasoning is based on the following sets, which generalize definition \eqref{eq:normalized_sec}.
\begin{definition}[Normalized secant sets]
\label{def:thesecant_sets}
We define the normalized secant set of some $\Xfrak \subseteq E$ as
\begin{equation}
\begin{split}
S[\Xfrak]&\stackrel{\D}{=}\{\frac{x-y}{\|x-y\|_E}: \ (x,y) \in \Xfrak^{2}, \|x-y\|_E > 0\}\,.
\end{split}
\end{equation}
We introduce also the normalized secant set of $\Xfrak \subseteq \Omega= \dom(f)$ embedded by $f$ as
\begin{equation}
\begin{split}
S[f(\Xfrak)]&\stackrel{\D}{=}\{\frac{f(x)-f(y)}{\|f(x)-f(y)\|_F}: \ (x,y) \in \Xfrak^{2}, \|f(x)-f(y)\|_F > 0\}\,.
\end{split}
\end{equation}
\end{definition}
The main contribution of this section is the control of the covering number of a normalized secant $S[f(\Xfrak)] \subseteq F$ by the covering number of $\Xfrak$ and $S[\Xfrak]$, for a sufficiently smooth $f$. The intuition is the following: if we consider a signal $x \in \Xfrak$ in a ‘‘low-dimensional'' space, then its image by $f$ is also ‘‘low-dimensional'' if $f$ is sufficiently regular on $\Xfrak$.

To carry out the analysis of $S[f(\Xfrak)]$ we introduce the notions of \emph{long} and \emph{short chords}. These objects are inspired by results in compressive independent component analysis and the theory of random projections on manifolds \cite{clakrson_chords}.

\smallskip
\paragraph{Long and short chords} The set $S[f(\Xfrak)]$ can be divided in two subsets that are more analytically tractable. First, we introduce, for $\eta >0$, the following sets 
\begin{equation}
\begin{split}
\operatorname{C}^{+}_{\eta}(\Xfrak,f) &\stackrel{\D}{=}\{(x,y) \in \Xfrak^{2} : \|f(x)-f(y)\|_F > \eta\}\,, \\
\operatorname{C}^{-}_{\eta}(\Xfrak,f) &\stackrel{\D}{=}\{(x,y) \in \Xfrak^{2} : 0<\|f(x)-f(y)\|_F \leq \eta\}\,.
\end{split}
\end{equation}
With these notations, $S[f(\Xfrak)]$ can be decomposed into long and short chords $S[f(\Xfrak)] = S^{+}_{\eta}[f(\Xfrak)] \cup S^{-}_{\eta}[f(\Xfrak)]$, where the long and short chords are respectively defined by
\begin{equation}
\label{eq:short_and_long_covering}
\begin{split}
&S^{+}_{\eta}[f(\Xfrak)]\stackrel{\D}{=}\{\frac{f(x)-f(y)}{\|f(x)-f(y)\|_F} : \ (x,y) \in \operatorname{C}^{+}_{\eta}(\Xfrak,f)\}\,, \\
&S^{-}_{\eta}[f(\Xfrak)]\stackrel{\D}{=}\{\frac{f(x)-f(y)}{\|f(x)-f(y)\|_F} : \ (x,y) \in \operatorname{C}^{-}_{\eta}(\Xfrak,f)\}\,.
\end{split}
\end{equation}
In order to control the covering number of $S[f(\Xfrak)]$ we will exploit its decomposition via the spaces $S^{\pm}_{\eta}[f(\Xfrak)]$ whose covering numbers are easier to obtain, assuming some regularity of the function $f$ on $\Xfrak$. 
\begin{definition}[Bi-Lipschitz assumption]
Given positive constants $\alpha$ and $\beta$, we say that a function $f$ is $(\alpha,\beta)$-bi-Lipschitz if for all $(x,y) \in \Xfrak^2$ it verifies
\begin{align}
\alpha \|x-y\|_E \leq &\|f(x)-f(y)\|_F  \label{eq:invlipsch} \\
&\|f(x)-f(y)\|_F \leq \beta \|x-y\|_E \,. \label{eq:lipsch} 
\end{align}
\end{definition}

In order to define the notion of differential we assume that the model set is contained in the interior of $\Omega =\dom(f)$, \textit{i.e.,} $\Xfrak \subseteq \interior(\Omega)$.
In particular when $f$ is differentiable on $\Xfrak$, \eqref{eq:lipsch} implies that ${\sup}_{x \in \Xfrak} \|\dr f_{x}\|_{\op} \leq \beta <+\infty$, where $\|\cdot\|_\op$ is the operator norm defined as $\|\dr f_{x}\|_{\op}\stackrel{\D}{=} \sup_{\|h\|_E \leq 1} \|\dr f_{x}(h)\|_F$.

We further consider the following regularity assumptions:
\begin{assumption}{1}{(Second order approximation)}\label{ass:A2}
$f$ is differentiable on $\interior(\Omega)$ and there exists $L>0$ such that
\begin{equation}
\forall (x,y) \in \Xfrak^{2}, \ \|f(x)-f(y)-\dr f_{y}(x-y)\|_F \leq L \|x-y\|^2_E\,.
\end{equation}
\end{assumption}
\begin{assumption}{2}{(Bounded curvature)}\label{ass:A4}
$f$ is differentiable on $\interior(\Omega)$ and $\zeta$-smooth on $\Xfrak$, where $\zeta >0$, \ie\,
\begin{equation}
\forall (x,y) \in \Xfrak^2, \ \|\dr f_x-\dr f_y\|_\op \leq \zeta\|x-y\|_E\,.
\end{equation}
\end{assumption}
As described in Lemma \ref{lemma_descent_lemma} (see Appendix~\ref{sec:descent_lemma}) the condition \asref{ass:A4} implies \asref{ass:A2} when $\Xfrak$ is a \emph{convex} model set. 

\smallskip
\paragraph{Covering numbers of $S^{+}_{\eta}[f(\Xfrak)]$} controlling the covering with long chords is relatively easy under some assumptions on $f$ as proven in the following proposition (proof in Appendix \ref{proof:prop:true_cov_number_long_chors}).

\begin{restatable}{proposition}{longchords}
\label{prop:true_cov_number_long_chors}
Assume that $f$ is $\beta$-Lipschitz as in \eqref{eq:lipsch}. Then for any $\eta >0$ and $\varepsilon >0$,
\begin{equation}
\Ncal(S^{+}_{\eta}[f(\Xfrak)],\|\cdot\|_F,\varepsilon) \leq \Ncal(\Xfrak,\|\cdot\|_E,\frac{\eta}{16 \beta} \varepsilon)^{2}\,.
\end{equation}
\end{restatable}
In other words, the ‘‘dimension'' of the long chords can be controlled by that of the model set itself, assuming only that $f$ is  Lipschitz-continuous.

\smallskip
\paragraph{Covering numbers of $S^{-}_{\eta}[f(\Xfrak)]$} controlling the covering number of short chords is a little more delicate and generally requires a fine analysis of $\Xfrak$ \cite{gribonval2020statistical}. However, by adding the hypothesis \asref{ass:A2} and \asref{ass:A4} we are able to prove the following result (proof in Appendix \ref{proof:prop:true_cov_number_short_chors}).

\begin{restatable}{proposition}{shortchords}
\label{prop:true_cov_number_short_chors}
Assume that $f$ is $(\alpha, \beta)$-bi-Lipschitz and satisfies assumptions \asref{ass:A2} and \asref{ass:A4}. Then for any $\varepsilon > 0$ and $\eta > 0$,
\begin{equation}
\begin{split}
&\Ncal(S^{-}_{\eta}[f(\Xfrak)], \|\cdot\|_F, 2(\varepsilon+\frac{L}{\alpha^{2}} \eta)) \leq \frac{\zeta+\beta \alpha}{\alpha^2 \varepsilon} \Ncal(\Xfrak, \|\cdot\|_E, \frac{\alpha\varepsilon }{\zeta+\beta\alpha})\times \Ncal(S[\Xfrak], \|\cdot\|_E, \frac{\alpha^2\varepsilon }{2(\zeta+\beta\alpha)})\,. \\
\end{split}
\end{equation}
\end{restatable}

\begin{proof}[Intuition of the proof]
The idea is to show that each element of $S^{-}_{\eta}[f(\Xfrak)]$ can be described by a ‘‘tangent'' vector to $S[\Xfrak]$ and that the set of tangent vectors has a covering number which can be controlled by those of $\Xfrak$ and $S[\Xfrak]$.
\end{proof}
By combining the two propositions we are now ready to state the main theorem of this section:
\begin{restatable}{theorem}{bigtheocovering}
\label{theorem:big_theo_covering}
Assume that $f$ is $(\alpha, \beta)$-bi-Lipschitz and satisfies assumptions \asref{ass:A2} and \asref{ass:A4} Then for all $\eta >0, \varepsilon > 0$, we have
\begin{equation}
\begin{split}
\Ncal(S[f(\Xfrak)], \|\cdot\|_F, 2\left[\varepsilon+\frac{L}{\alpha^{2}} \eta\right]) \leq \ &  \Ncal(\Xfrak,\|\cdot\|_E,\frac{\eta}{8 \beta} \left[\varepsilon+\frac{L}{\alpha^{2}} \eta\right])^{2} \\
& + \frac{\zeta+\beta \alpha}{\alpha^2 \varepsilon}\Ncal(\Xfrak, \|\cdot\|_E, \frac{\alpha\varepsilon }{\zeta+\beta\alpha})\times \Ncal(S[\Xfrak], \|\cdot\|_E, \frac{\alpha^{2} \varepsilon}{2(\zeta+\beta\alpha)})\,.\\
\end{split}
\end{equation}
\end{restatable}
\begin{proof}
We use that for any $\eta >0$ $S[\Xfrak]=S^{+}_{\eta}[f(\Xfrak)]\cup S^{-}_{\eta}[f(\Xfrak)]$ thus the covering number of $S[f(\Xfrak)]$ is less than the sum of the coverings of $S^{+}_{\eta}[f(\Xfrak)]$ and $S^{-}_{\eta}[f(\Xfrak)]$. Then we apply Proposition \ref{prop:true_cov_number_long_chors}  and Proposition \ref{prop:true_cov_number_short_chors}.
\end{proof}

\smallskip
\paragraph{Putting everything together in the case of sparse precision matrix} we apply the previous results to our framework, that is $E = F = S_d$, $f=\inv$ and $\Xfrak = \Sfrak_{k,a,b}^{-1} = \{ \Thetab = \Sigmab^{-1}: \ \Sigmab \in  \Sfrak_{k,a,b}\}$. We set the norms as follow : $\|\cdot\|_E = \|\cdot\|_\Fro$ the Frobenius norm  and $\|\cdot\|_F = \|\cdot\|_{\Lambda}$ as defined in \eqref{eq:norm_lambda}. We have the following lemma which shows that $f=\inv$ satisfies all the necessary regularity assumptions (the proof can be found in Appendix \ref{proof:lemma:finvsatisfies}).
\begin{restatable}{lemma}{finvsatisfies}
\label{lemma:finvsatisfies}
Assume that there exist constants $c_\Fro$ and $C_\Fro$ such that $c_\Fro \|\Mbf\|_\Fro \leq \|\Mbf\|_\Lambda \leq C_\Fro \|\Mbf\|_\Fro$.
Then the function $f = \inv$ is $(\alpha, \beta)$-bi-Lipschitz and satisfies assumptions \asref{ass:A2} and \asref{ass:A4} on $\Sfrak_{k,a,b}^{-1}$ with $\alpha~=~\frac{c_\Fro}{b^2}, \quad \beta~=~\frac{C_\Fro}{a^2}, \quad L~=~\frac{C_\Fro}{a^3}, \quad \zeta~=~2L$ .
\end{restatable}
Note that expressions for the constants $c_\Fro$ and $C_\Fro$ will be provided in Proposition~\ref{prop:lambda_norm_controls}. In order to control the covering number of $S[\Sfrak_{k,a,b}] = S[\inv(\Sfrak_{k,a,b}^{-1})]$ we only have to check those of $\Sfrak_{k,a,b}^{-1}$ and $S[\Sfrak_{k,a,b}^{-1}]$. It is done in the following lemma (the proof can be found in Appendix \ref{proof:lemma:covering_secant_and_sig}).
\begin{restatable}{lemma}{coveringsec}
\label{lemma:covering_secant_and_sig}
For any $\varepsilon >0$ we have
\begin{equation}
\begin{split}
&\Ncal(\Sfrak_{k,a,b}^{-1}, \|\cdot\|_{\Fro}, \varepsilon) \leq (\frac{ed^2}{2k})^{k}(\frac{18 \sqrt{d} \times b}{\varepsilon})^{d+k} \text{ and } \Ncal(S[\Sfrak_{k,a,b}^{-1}], \|\cdot\|_{\Fro},\varepsilon) \leq (\frac{ed^2}{4k})^{2k}(\frac{18}{\varepsilon})^{d+2k}\,.
\end{split}
\end{equation}
\end{restatable}
This result show that the \emph{box counting dimensions} \cite{robinson_2010} (also called entropy dimensions) of $\Sfrak_{k,a,b}^{-1}$ and $S[\Sfrak_{k,a,b}^{-1}]$ are smaller than $(d+k)\log(d)$ and $(d+2k)\log(d)$ and thus much smaller than $d^{2}$, which will allow us to have the guarantees presented in the introduction.
\begin{restatable}{corollary}{covnbappli}
Assuming the existence of the constant $C_\Fro, c_\Fro$ as in Lemma~\ref{lemma:finvsatisfies}, there exist absolute constants $c_0, c_1 \geq 1$, such that, for any $\varepsilon>0$, the covering number of $S[\Sfrak_{k,a,b}]$ verifies 
\begin{equation}
\Ncal\left(S[\Sfrak_{k,a,b}], \|\cdot\|_\Lambda, \varepsilon \right) \leq \left(\frac{ed^2}{2k}\right)^{4k}\left[\left(c_0\frac{\sqrt{d} C_\Fro^2}{\varepsilon^2 c_\Fro^2} \frac{b^5}{a^5} \right)^{2(d+k)} +\left(c_1 \frac{\sqrt{d} C_\Fro^2}{\varepsilon^2 c_\Fro}(\frac{2}{c_\Fro}+1)\frac{b^5}{a^5}\right)^{d+2k+1}\right] \,.
\end{equation}
\label{coro:cov_nb_appli}
\end{restatable}
This is a direct consequence of Theorem~\ref{theorem:big_theo_covering}, combined with Lemma~\ref{lemma:finvsatisfies} and \ref{lemma:covering_secant_and_sig}. See Appendix~\ref{proof:cov_nb_appli} for the proof. This corollary allows for a control of the covering number that is required to prove a $\RIP$ for rank-one projections.

\subsection{Application to rank-one projections \label{sec:rank_one}}

The results of the previous section allowed us to control one of the three quantities of interest for establishing the $\RIP_{\delta}$ : the covering number. We are left with studying the concentration properties of the operator $\Acal$ (Theorem \ref{theo:rip_cond2}).

A natural choice for the norm $\|\cdot\|_{\R^m}$ would be the standard Euclidean norm $\|\cdot\|_{2}$. However, in order for the $\RIP$ to hold, one would have to choose $\|\Mbf \|_{S_d} = \left(\E_{\Abf \sim \Lambda}\left[\left|\langle \Mbf_j, \Mbf \rangle\right|^{2}\right]\right)^{1/2}$.
Unfortunately, with these choices, we were unable to find sufficiently tight concentration functions $C_1, C_2$ that lead to better guaranties than $m \gtrsim d^{2}$. This phenomenom had already been observed, as written in \cite{ROParticle} the rank-one projections lead to loose RIP constants for low-rank matrix completion problems (unless  $m \gtrsim d^{2}$), and we envision that similar results hold in our case. The remedy found in \cite{ROParticle} is to consider instead an $\ell_1$ RIP (\ie\ taking the $\ell_1$ norm for $\|\cdot\|_{\R^m}$ in Definition \ref{def:def_rip} instead of the $\ell_2$ norm), leading to the choice of $\|\cdot \|_{\Lambda}$ defined in \eqref{eq:norm_lambda} for $\|\cdot\|_{S_d}$. This next part shows that this remedy helps for the recovery of covariance matrices and thus of precision matrices.

\begin{restatable}{proposition}{concentrationForROP}
\label{prop:C2forROP}
Consider the operator $\mathcal{A}(\Sigmab) = \frac{1}{m} \left( \a_1^\top \Sigmab \a_1, \dots, \a_m^\top \Sigmab \a_m  \right)^{\top}$, where the vectors $\a_1, \cdots, \a_m$ are \textit{i.i.d.} and either follow $\Lambda = \Lambda_{\operatorname{G}} =\Ncal(0, \frac{1}{d} \mathbf{I}_d)$ or $\Lambda = \Lambda_{\operatorname{G}} = \mathcal{U}(\mathbb{S}^{d-1})$ with the associated norm $\|\Sigmab\|_\Lambda = \E_{\a \sim \Lambda}[|\a^\top \Sigmab \a|]$.
Then, for any $\U \in S_d(\R)$ satisfying $\|\U\|_\Lambda = 1$ and for any $t>0$, we have
\begin{equation}
  \mathbb{P} \left( \big| \|\Acal(\U)\|_1 - 1 \big| > t \right) \leq 2 \exp \left(- \frac{m}{8e^2} \min \left(\frac{t^2}{\ssexpo^2}, \frac{t}{\ssexpo}\right)\right),
\label{eq:C2forROP}
\end{equation}
where $\ssexpo$ is an absolute constant given by $K_{\Lambda_{\operatorname{G}}} = \frac{76 e^2 \sqrt{15}}{\log 2}$ and $K_{\Lambda_{\operatorname{U}}} = \frac{304 e^3 \sqrt{15}}{\log 2}$.
\end{restatable}

\begin{proof}[Intuition of the proof]
  The proof is based on a Bernstein-type concentration inequality for sums of sub-exponential variables. The essential ingredient of the proof is thus to show that the centered random variable $|\a^\top \U \a| - 1$ involved in $ \|\Acal(\U)\|_1 - 1$ is subexponential with a subexponential norm bounded by an absolute constant $\ssexpo$. The full proof can be found in Appendix~\ref{proof:RO}. 
\end{proof}
This result provides the function $C_2$ required to obtain a RIP with Theorem~\ref{theo:rip_cond2}. 
\begin{remark}
This is essentially the only result that would need to be adapted if one wants to provide information-theoretic guarantees to the sketching operator defined from random structured matrices. Indeed, we emphasize that all previous results on covering numbers are still valid in the structured case. 
\end{remark}
Finally to be able to control the covering number, the norm $\|\cdot\|_\Lambda$ needs to verify the hypothesis of Corollary~\ref{coro:cov_nb_appli}. This is done in the following proposition (see Appendix~\ref{proof:prop:lambda_norm_controls} for the proof).

\begin{restatable}{proposition}{lambdaNormControls}
\label{prop:lambda_norm_controls}
Let $\Lambda \in \{ \Lambda_{\operatorname{G}}, \Lambda_{\operatorname{U}} \}$ be either the Gaussian or uniform distribution on $\R^d$ and consider the associated $\Lambda$-norm defined in \eqref{eq:norm_lambda}. Then,
\begin{equation}
  \forall \Mbf \in S_d(\R), \ \frac{2}{9\sqrt{15}d} \|\Mbf\|_\Fro \leq \|\Mbf\|_\Lambda \leq \frac{1}{\sqrt{d}} \|\Mbf\|_\Fro\,.
\end{equation}
This gives $c_\Fro = 2 / (9\sqrt{15}d)$ and $C_\Fro = 1/\sqrt{d}$ in Lemma \ref{lemma:finvsatisfies}.
\end{restatable}

In this section, we have established control over the covering numbers via Corollary~\ref{coro:cov_nb_appli} and introduced the concentration inequality for the sketching operator through Proposition~\ref{prop:C2forROP}. These components provide the necessary foundation for proving Theorem~\ref{theo:theRIPRO}. The comprehensive proof of this theorem can be found in Appendix~\ref{proof:theRIPRO}.

\section{Conclusion \& perspectives}

In this work, we have presented a compressive approach based on sketching to estimate sparse precision matrices. We have shown that it is possible to estimate a $(d+2k)$-sparse precision matrix from a data sketch of the order $(d+2k) \log(d)$, which is significantly smaller than the typical memory complexity of $d^2$ associated with the graphical lasso. Our analysis is supported by information-theoretic guarantees, where we have established restricted isometries and instance optimality properties. Finally, we have proposed a practical algorithmic solution for computing an estimation of the precision matrix from the sketch.

Our work opens several new lines of research. Given the generality of the tools presented in this paper, it would be interesting to explore whether similar guarantees can be established for other model sets based on specific graph structures \cite{Sandeep}. Also, as practitioners are not always interested in the graph associated with the precision matrix \textit{per se} but rather in some of its properties (\textit{e.g.,} a group structure among the nodes), it would be interesting to see how these properties can directly be inferred using a compressive learning approach and whether it can help further reduce the sketch's dimension. 

From an algorithmic point of view, our work also raises several questions. The proposed estimator is costly as it requires to solve several graphical lasso. An interesting further work would be to design a more efficient decoder that is sufficiently close to the optimal decoder given by the theory. In this context, the choice proposed in this paper is a first step toward practical recovery, but other algorithms could be used based on different precision matrix estimators like \cite{elementary_esti}. Finally, from an application point of view, an interesting perspective would be to use the sketching approach to learn, in an online way, a dynamic graph, in the way of the time varying graphical lasso \cite{time_glasso}.

\appendix

\section{Proofs}

\subsection{Proofs of Section \ref{sec:recipies}}

\subsubsection{Proof of Theorem \ref{theo:invriptodecoder}}
\label{proof:invriptodecoder}
\begin{proof}
With the notations of the theorem $\Sigmab^* = \D[\Acal(\widehat{\Sigmab})] \in \arg\min_{\Sigmab \in \Sfrak} \|\Acal(\Sigmab)-\Acal(\widehat{\Sigmab})\|_{\R^m} $. Then for any $\Sigmab_{\Sfrak} \in \Sfrak$:
\begin{equation}
\begin{split}
\|\Sigmab^*-\Sigmab\|_{S_d} &\leq \|\Sigmab-\Sigmab_{\Sfrak}\|_{S_d}+\|\Sigmab_{\Sfrak}-\Sigmab^*\|_{S_d} \\
&\leq \|\Sigmab-\Sigmab_{\Sfrak}\|_{S_d} + \frac{1}{1-\delta}\|\Acal(\Sigmab^*)-\Acal(\Sigmab_\Sfrak)\|_{\R^m} \\
&\leq \|\Sigmab-\Sigmab_{\Sfrak}\|_{S_d} + \frac{1}{1-\delta} \left( \|\Acal(\Sigmab^*)-\Acal(\widehat{\Sigmab})\|_{\R^m}+\|\Acal(\widehat{\Sigmab})-\Acal(\Sigmab_\Sfrak)\|_{\R^d} \right) \\
&\leq \|\Sigmab-\Sigmab_{\Sfrak}\|_{S_d} + \frac{2}{1-\delta} \|\Acal(\widehat{\Sigmab})-\Acal(\Sigmab_\Sfrak)\|_{\R^m} \\
& \leq  \|\Sigmab-\Sigmab_{\Sfrak}\|_{S_d} +\frac{2}{1-\delta} \|\Acal(\widehat{\Sigmab})-\Acal(\Sigmab)\|_{\R^d} +\frac{2}{1-\delta} \|\Acal(\Sigmab)-\Acal(\Sigmab_\Sfrak)\|_{\R^m} \,.
\end{split}
\end{equation}
We introduce the following ‘‘distance'' to the model set $\Sfrak$:
\begin{equation}
d^{\circ}(\Sigmab, \Sfrak)=\inf_{\Mbf \in \Sfrak} \|\Sigmab-\Mbf\|_{S_d}+\frac{2}{1-\delta} \|\Acal(\Sigmab)-\Acal(\Mbf)\|_{\R^m}\,.
\end{equation}
Then we have $d^{\circ}(\Sigmab, \Sfrak) = 0$ if $\Sigmab \in \Sfrak$ and 
\begin{equation}
\|\Sigmab^*-\Sigmab\|_{S_d} \leq d^{\circ}(\Sigmab, \Sfrak)+\frac{2}{1-\delta} \|\Acal(\Sigmab)-\Acal(\Mbf)\|_{\R^m}\,.
\end{equation}
\end{proof}

\subsubsection{Proof of Theorem \ref{theo:rip_cond2}}
\label{proof:rip_cond2}
\begin{proof}

Let us start by considering $\overline{S}_{\varepsilon}$ , an $\varepsilon$-net of $S[\Sfrak]$ with respect to $\|\cdot\|_{S_d}$, for some $\varepsilon>0$. Then, for every $\U \in S[\Sfrak]$, there exists $\overline{\U} \in \overline{S}_{\varepsilon}$ such that $\|\U-\overline{\U}\|_{S_d} \leq \varepsilon$. From the triangular inequality we have
\begin{equation*}
\big|\|\Acal(\U)\|_{\R^m}-1\big|  \leq \big| \|\Acal(\U) \|_{\R^m}-\|\Acal(\overline{\U})\|_{\R^m}\big| + \big| \|\Acal(\overline{\U})\|_{\R^m}-1\big| .
\end{equation*}
Focusing on the first term of the right-hand side, we obtain
\begin{equation}
\begin{split}
\big| \|\Acal(\U) \|_{\R^m}-\|\Acal(\overline{\U})\|_{\R^m}\big| &\leq \|\Acal(\U-\overline{\U})\|_{\R^m} \leq \vertiii{\Acal} \cdot \|\U-\overline{\U}\|_{S_d} \leq \varepsilon  \vertiii{\Acal} \,.
\end{split}
\end{equation}
Hence for $ \U \in S[\inv(\Sfrak)]$:
\begin{equation}
\big|\|\Acal(\U)\|_{\R^m}-1\big|  \leq \max_{\overline{U} \in \overline{S}_{\varepsilon}}\big|\|\Acal(\overline{\U})\|_{\R^m} -1\big| + \varepsilon  \vertiii{\Acal} .
\end{equation}
So we have for any $0<\delta <1$:
\begin{equation}
\begin{split}
&\Pbb\left(\sup_{\U \in S[\Sfrak]} \big|\|\Acal(\U)\|_{\R^m}-1\big| \leq \delta \right) \geq 1- \Pbb(\varepsilon \vertiii{\Acal} > \frac{\delta}{2})-\Pbb(\max_{\overline{U} \in \overline{S}_{\varepsilon}}  \left|\|\Acal(\overline{\U})\|_{\R^m} -1\right| > \frac{\delta}{2})\,.
\end{split}
\end{equation}
We will control these two terms. For the first one we have the concentration with $C_1$. For the second one, using the union bound yields
\begin{equation}
\label{eq:eqintermedita}
\begin{split}
\Pbb(\max_{\overline{U} \in \overline{S}_{\varepsilon}}\big|\|\Acal(\overline{\U})\|_{\R^m} -1\big| > \frac{\delta}{2}) &\leq \sum_{\overline{\U} \in \overline{S}_{\varepsilon}} \Pbb(\big|\|\Acal(\overline{\U})\|_{\R^m} -1\big| > \frac{\delta}{2})
\end{split}
\end{equation}
Using the concentration given by $C_2$, for $\overline{\U} \in \overline{S}_{\varepsilon}$ and $t \in ]0,1[$, we have  $\Pbb(\big|\|\Acal \overline{\U}\|_{\R^m} -1\big| > t) \leq C_2(t)$. 
Applying this with $t=\delta/2$ and using \eqref{eq:eqintermedita} gives
\begin{equation}
\Pbb(\max_{\overline{U} \in \overline{S}_{\varepsilon}}\big|\|\Acal(\overline{\U})\|_2 -1\big| > \frac{\delta}{2}) \leq |\overline{S}_{\varepsilon}| C_2(\delta/2)\,. 
\end{equation}
As a result, we obtain for $\delta \in ]0,1[$ and $\varepsilon >0$
\begin{equation}
\begin{split}
\Pbb\left(\sup_{\U \in S[\Sfrak]} \big|\|\Acal(\U)\|_2-1\big| \leq \delta \right) & \geq 1 - C_1(\frac{\delta}{2 \varepsilon}) - |\overline{S}_{\varepsilon}| C_2(\delta/2)\,.
\end{split}
\end{equation}
\end{proof}

\subsubsection{Proof of Remark~\ref{rem:C1fromC2} }
\label{pr:concentrationLink}

\begin{proof}
 For some $\varepsilon'>0$, consider $T_{\varepsilon'}$ an $\varepsilon'$-net of the sphere $B_{S_d}$. Then, for all $\U \in B_{S_d}$, there exists $\overline{\U} \in T_{\varepsilon'}$ such that
 \begin{equation*}
   \| \Acal(\U)  \|_{\R^m} \leq \varepsilon' \vertiii{\Acal} + \|\Acal(\overline{\U}) \|_{\R^m}\,.
 \end{equation*}
 Thus, taking the supremum over $\U$ yields
 \begin{equation*}
   \sup_{\U \in B_{S_d}} \| \Acal(\U)  \|_{\R^m} \leq \varepsilon' \vertiii{\Acal} + \max_{\overline{\U} \in T_{\varepsilon'}} \|\Acal(\overline{\U}) \|_{\R^m}\,,
 \end{equation*}
 therefore $\vertiii{\Acal} \leq (1-\varepsilon')^{-1} \ \max_{\overline{\U} \in T_{\varepsilon'}} \|\Acal(\overline{\U}) \|_{\R^m}$. As a consequence, for $t>1/(1-\varepsilon')$,
 \begin{align*}
   \proba{}{\vertiii{\Acal} > t} &\leq \proba{}{\max_{\overline{\U} \in T_{\varepsilon'}} \|\Acal(\overline{\U}) \|_{\R^m} > (1- \varepsilon') t} \leq \sum\limits_{\overline{\U} \in T_{\varepsilon'}} \proba{}{|\|\Acal(\overline{\U}) \|_{\R^m}-1| > (1- \varepsilon') t-1} \\
   &\leq \Ncal( B_{S_d}, \|\cdot\|_{S_d}, \varepsilon' ) C_2((1- \varepsilon') t-1) \,.
 \end{align*}
\end{proof}

\subsection{Proofs of Section \ref{sec:covering}}

\subsubsection{General results on covering numbers \label{sec:generality_covering_numbers}}

Let $(E,d)$ be a semi-metric space. The covering number of $\Sfrak \subseteq E$ with radius $\varepsilon$ with respect to $d$ is defined as:
\begin{equation}
\Ncal(\Sfrak,d, \varepsilon)\stackrel{\D}{=}\min\left\{N \in \mathbb{N} : \exists x_1, \cdots, x_N \in \Sfrak, \Sfrak \subseteq \bigcup_{i=1}^{N} B_{d}(x_i,\varepsilon)\right\}\,.
\end{equation}
If $N = \Ncal(\Sfrak,d, \varepsilon)$, then for any $x \in \Sfrak$ there exists $i \in \integ{N}$ such that $d(x,x_i) \leq \varepsilon$. We recall the following Lemma regarding covering numbers that can be found in \cite[Lemma A.3.]{gribonval2020statistical}.
\begin{lemma}
\label{lemma:generalized_cover}
Let $Y,Z$ we two subset of a pseudo metric space $(X,d)$ such that the following holds:
\begin{equation}
\forall z \in Z, \exists y \in Y, d(z,y) \leq \delta\,,
\end{equation}
where $\delta \geq 0$. Then for all $\varepsilon > 0$
\begin{equation}
\Ncal(Z, d, 2 (\delta+\varepsilon)) \leq \Ncal(Y, d, \varepsilon)\,.
\end{equation}
\end{lemma}

\subsubsection{Descent Lemma}
\label{sec:descent_lemma}

\begin{lemma}[Descent Lemma]
\label{lemma_descent_lemma}
Let $E, F$ be two normed vector spaces and $\Omega$ a subset of $E$ and $\Xfrak \subseteq \interior(\Omega)$. Consider $f: \Omega \rightarrow F$ a $L$-smooth function on $\Xfrak$ \ie\
\begin{equation}
\forall (x,y) \in \Xfrak^{2}, \ \|\dr f_x - \dr f_y\|_{\op} \leq L \|x-y\|_E\,.
\end{equation}
Let $(x,y) \in \Xfrak^{2}$ such that the segment $\integ{x,y}$ lies in $\Xfrak$. Then,
\begin{equation}
\|f(x)-f(y)-\dr f_{y}(x-y)\|_F \leq L \|x-y\|^2_E\,.
\end{equation}
In particular if $\Xfrak$ is convex then \asref{ass:A4} implies \asref{ass:A2}.
\end{lemma}

\begin{proof}
From \cite[Corollary 3.3]{coleman2012calculus} $f$ verifies
\begin{equation*}
\|f(x)-f(y)-\dr f_{y}(x-y)\|_F \leq \sup_{z \in \integ{x,y}} \|\dr f_{z} - \dr f_{x} \|_{\op} \|x-y\|_E\,.
\end{equation*}
Now take $z \in \integ{x,y}$ and write it as $z = (1-t)x+t y \in \Xfrak$ for some $t \in [0,1]$. Since $f$ is $L$-smooth on $\Xfrak$ we have:
\begin{equation*}
\begin{split}
&\|\dr f_{z} - \dr f_{x} \|_{\op} = \|\dr f_{(1-t)x+t y} - \dr f_{x} \|_{\op} \leq L \|(1-t)x+t y - x\|_E = L t \| x - y \|_E \leq L \| x - y \|_E
\end{split}
\end{equation*}
Hence $\sup_{z \in \integ{x,y}} \|\dr f_{z} - \dr f_{x} \|_{\op} \leq L \| x - y \|_E$ and thus $\|f(x)-f(y)-\dr f_{y}(x-y)\|_F  \leq L \|x-y\|^2_E$
\end{proof}

\subsubsection{Proof of Proposition \ref{prop:true_cov_number_long_chors} \label{proof:prop:true_cov_number_long_chors}}

\begin{proof}
In the following we note $\|(x_1,y_1)-(x_2,y_2)\|_{\otimes2}=\|x_1-x_2\|_{E}+\|y_1-y_2\|_{E}$. We introduce,  for $\eta \geq 0$, the set 
\begin{equation}
\Xfrak^{2}_{\eta}=\{(x,y) \in \Xfrak^{2}: \  \|f(x)-f(y)\|_{F}> \eta\}\,.
\end{equation}
First note that $\Xfrak^{2}_{\eta} \subset \Xfrak^{2}$, and consequently
\begin{equation}
\Ncal(\Xfrak^{2}_{\eta},\|\cdot\|_{\otimes2},\varepsilon) \leq \Ncal(\Xfrak^{2},\|\cdot\|_{\otimes2},\frac{\varepsilon}{2})\leq \Ncal(\Xfrak,\|\cdot\|_{E},\frac{\varepsilon}{4})^{2}\,.
\end{equation}
We consider $\eta >0$ and define $g: \Xfrak_{\eta}^{2} \rightarrow S_{\eta}^+(f(\Xfrak))$ by $g(x,y)=\frac{f(x)-f(y)}{\|f(x)-f(y)\|_{F}}$ for $(x,y) \in \Xfrak_{\eta}^{2}$. 
By definition, $g$ is surjective. We will show that it is also Lipschitz. With $(x_1,y_1),(x_2,y_2) \in \Xfrak^{2}_{\eta} \times \Xfrak^{2}_{\eta}$ we obtain
\begin{equation*}
\begin{split}
&\| \frac{f(x_1)-f(y_1)}{\|f(x_1)-f(y_1)\|_F}- \frac{f(x_2)-f(y_2)}{\|f(x_2)-f(y_2)\|_F}\|_F \\
\leq & \| \frac{f(x_1)-f(y_1)}{\|f(x_1)-f(y_1)\|_F}- \frac{f(x_1)-f(y_1)}{\|f(x_2)-f(y_2)\|_F}\|_{F} + \| \frac{f(x_1)-f(y_1)}{\|f(x_2)-f(y_2)\|_F}- \frac{f(x_2)-f(y_2)}{\|f(x_2)-f(y_2)\|_F} \|_{F} \\
\leq & \frac{1}{\|f(x_2)-f(y_2)\|_F}\left(\|f(x_1)-f(x_2)\|_F+\|f(y_1)-f(y_2)\|_F\right)\\
&\qquad + \quad \left|\frac{1}{\|f(x_1)-f(y_1)\|_F} - \frac{1}{\|f(x_2)-f(y_2)\|_F}\right| \|f(x_1)-f(y_1)\|_F \\
\leq & \frac{2\beta}{\|f(x_2)-f(y_2)\|_F} \|(x_1,y_1)-(x_2,y_2)\|_{\otimes2} \\
& \qquad + \quad \left|\|f(x_2)-f(y_2)\|_{F}-\|f(x_1)-f(y_1)\|_{F}\right| \times \frac{\|f(x_1)-f(y_1)\|_F}{\|f(x_1)-f(y_1)\|_{F} \|f(x_2)-f(y_2)\|_{F}}\\
\leq &\frac{2\beta}{\eta} \|(x_1,y_1)-(x_2,y_2)\|_{\otimes2} + \|f(x_2)-f(x_1)-f(y_2)+f(y_1)\|_F \times \frac{1}{\|f(x_2)-f(y_2)\|_{F}} \\
\leq & \frac{2\beta}{\eta} \|(x_1,y_1)-(x_2,y_2)\|_{\otimes2} + 2 \beta \|(x_1,y_1)-(x_2,y_2)\|_{\otimes2} \times \frac{1}{\|f(x_2)-f(y_2)\|_{F}} \\
\leq& \frac{4 \beta}{\eta}\|(x_1,y_1)-(x_2,y_2)\|_{\otimes2}\,.
\end{split}
\end{equation*}
Consequently,
\begin{equation}
\|g(x_1,y_1)-g(x_2,y_2)\|_F \leq \frac{4 \beta}{\eta}\|(x_1,y_1)-(x_2,y_2)\|_{\otimes2}\,.
\end{equation}
So $g$ is a surjective $\frac{4 \beta}{\eta}$-Lipschitz function from $(\Xfrak^{2}_{\eta},\|\cdot\|_{\otimes2})$ to $(S_{\eta}^+(f(\Xfrak)),\|\cdot\|_{F})$. Hence using \cite[Lemma A.2.]{gribonval2020statistical} we obtain
\begin{equation}
\begin{split}
\Ncal(S_{\eta}^+(f(\Xfrak)),\|\cdot\|_F,\varepsilon) &\leq \Ncal(\Xfrak^{2}_{\eta},\|\cdot\|_{\otimes2},\frac{\eta}{4 \beta} \varepsilon) \leq \Ncal(\Xfrak,\|\cdot\|_E,\frac{\eta}{16 \beta} \varepsilon)^{2}\,.
\end{split}
\end{equation}

\end{proof}

\subsubsection{Proof of Proposition \ref{prop:true_cov_number_short_chors} \label{proof:prop:true_cov_number_short_chors}}

In order to prove the result, the reasoning will be the following: 1) we will show that any element of $S_{\eta}^{-}[f(\Xfrak)]$ is close to an element of a certain ‘‘tangent space'' $\dr \frac{x-y}{\|f(x)-f(y)\|_F}$ 2) we will control the covering number of this space. We begin with the following result.

\begin{lemma}
\label{lemma:exists_chords}
Assume that $f$ is $(\alpha,\beta)$-bi-Lipschitz and satisfies assumptions \asref{ass:A2}. For $\varepsilon_0 > 0$, consider the following subset of $\Xfrak^{2}$:
\begin{equation}
I_{\varepsilon_0}\stackrel{\D}{=}\{(x,y) \in \Xfrak^{2}: \ 0< \|x-y\|_E \leq \varepsilon_0\}\,.
\end{equation}
Then, for any $\varepsilon_0 > 0$ and $(x,y) \in I_{\varepsilon_0}$,
\begin{equation*}
\|\frac{f(x)-f(y)}{\|f(x)-f(y)\|_F}-\dr f_{y} \frac{x-y}{\|f(x)-f(y)\|_F}\|_F \leq \frac{L}{\alpha}\varepsilon_0\,.
\end{equation*}
In particular, for any $\varepsilon_0 > 0$,
\begin{equation*}
\begin{split}
&\forall (x,y) \in I_{\varepsilon_0} , \ \exists h \in \Xfrak, \ \|\frac{f(x)-f(y)}{\|f(x)-f(y)\|_F}-\dr f_{h} \frac{x-y}{\|f(x)-f(y)\|_F}\|_F \leq \frac{L}{\alpha}\varepsilon_0\,.
\end{split}
\end{equation*}
\end{lemma}
\begin{proof}
First, assumption \asref{ass:A2} gives
\begin{equation}
\label{eq:hypotheseA1used}
\begin{split}
\|f(x)-f(y)-\dr f_{y} (x-y)\|_F &\stackrel{\asref{ass:A2}}{\leq} L\|x-y\|_E^{2}\leq L\varepsilon_0 \|x-y\|_E\,.
\end{split}
\end{equation}
Now since $\|x-y\|_E >0$, we have $\|f(x)-f(y)\|_F >0$ using \eqref{eq:invlipsch} (from the inverse Lipschitz property). Thus by dividing by $\|f(x)-f(y)\|_F$ in \eqref{eq:hypotheseA1used} we obtain
\begin{equation*}
\begin{split}
&\|\frac{f(x)-f(y)}{\|f(x)-f(y)\|_F}-\dr f_{y} \frac{x-y}{\|f(x)-f(y)\|_F}\|_F \leq L\varepsilon_0 \frac{\|x-y\|_E}{\|f(x)-f(y)\|_F} \stackrel{\eqref{eq:invlipsch}}{\leq} \frac{L}{\alpha}\varepsilon_0\,.
\end{split}
\end{equation*}
\end{proof}
The above result induces the following corollary.
\begin{corollary}
\label{corr:exists_tanget}
Assume that $f$ is $(\alpha, \beta)$-bi-Lipschitz and satisfies assumptions \asref{ass:A2}. For $\eta >0$ consider $S^{-}_{\eta}[f(\Xfrak)]$ as defined in \eqref{eq:short_and_long_covering} and the normalized secant set $S[\Xfrak]$ (see Definition \ref{def:thesecant_sets}). Let
\begin{equation}
\label{eq:definitionC}
C\stackrel{\D}{=} \{\lambda v: \ \lambda \in ]0,\frac{1}{\alpha}], v \in S[\Xfrak] \}\,,
\end{equation}
and
\begin{equation}
\label{eq:definitionT}
T_C \stackrel{\D}{=} \left\{\dr f_{h}(c): (h,c) \in \Xfrak \times C\right\}\,.
\end{equation}
Then
\begin{equation}
\forall u \in S^{-}_{\eta}[f(\Xfrak)], \exists t \in T_C, \ \|u-t\|_F \leq \frac{L}{\alpha^{2}} \eta\,.
\end{equation}

\end{corollary}
\begin{proof}
Take $u = \frac{f(x)-f(y)}{\|f(x)-f(y)\|_F} \in S^{-}_{\eta}[f(\Xfrak)]$ (thus with $0< \|f(x)-f(y)\|_F \leq \eta$). By using that $f$ is $(\alpha, \beta)$-bi-Lipschitz, we have $0< \|x-y\|_E \leq \eta/\alpha$. So we can apply the Lemma \ref{lemma:exists_chords} with $\varepsilon_0 = \eta/\alpha$ to prove that there exists $h \in \Xfrak$ such that
\begin{equation*}
\|\frac{f(x)-f(y)}{\|f(x)-f(y)\|_F}-\dr f_{h} \frac{x-y}{\|f(x)-f(y)\|_F} \|_F \leq \frac{L}{\alpha^{2}} \eta\,.
\end{equation*}
Rewrite $\frac{x-y}{\|f(x)-f(y)\|_F} = \frac{x-y}{\|x-y\|_F} \frac{\|x-y\|_E}{\|f(x)-f(y)\|_F}$ and  define $\lambda = \frac{\|x-y\|_E}{\|f(x)-f(y)\|_F}$. Therefore, we have $\lambda >0$ and $|\lambda| \leq \frac{1}{\alpha}$. It proves that there exists $\lambda \in ]0, 1/\alpha ]$ and $h \in \Xfrak$, such that
\begin{equation*}
\|\frac{f(x)-f(y)}{\|f(x)-f(y)\|_F}-\dr f_{h} \lambda \frac{x-y}{\|x-y\|_E} \|_F \leq \frac{L}{\alpha^{2}} \eta\,.
\end{equation*}
Considering $v = \frac{x-y}{\|x-y\|_E}\in S[\Xfrak]$ concludes the proof.

\end{proof}

The last thing to do is to control the covering number of $T_C$ in the previous result. This will be done using the following lemma.
\begin{lemma}
\label{lemma:subset}
Let $C \subseteq E$ be any set such that  $\forall c \in C, \|c\|_E \leq \delta$ for some $\delta >0$. Assume that $f$ is $(\alpha, \beta)$-bi-Lipschitz and satisfies \asref{ass:A4}. Consider $T_C \stackrel{\D}{=} \left\{\dr f_{h}(c): (h,c) \in \Xfrak \times C\right\}$.
Then for any $\varepsilon >0$,
\begin{equation*}
\begin{split}
&\Ncal(T_C, \|\cdot\|_F,   \varepsilon) \leq \Ncal(\Xfrak, \|\cdot\|_E, \frac{\varepsilon}{\zeta\delta+\beta})\times \Ncal(C, \|\cdot\|_E, \frac{\varepsilon}{\zeta\delta+\beta}) \,.
\end{split}
\end{equation*}

\end{lemma}
\begin{proof}

Take $\overline{\Xfrak}$ a $\varepsilon$-net of $\Xfrak$ and $\overline{C}$ a $\varepsilon$-net of $C$. Then take $t= \dr f_h(u) \in T_C$ and consider $\overline{u}, \overline{h} \in \overline{\Xfrak} \times \overline{C}$ such that $\|\overline{c}-c\|_E \leq \varepsilon$ and $\|\overline{h}-h\|_E \leq \varepsilon$. Then with $\overline{t}= \dr f_{\overline{h}}(\overline{c}) \in T_C$
\begin{equation}
\begin{split}
\|t-\overline{t}\|_F &= \|\dr f_h(c)-\dr f_{\overline{h}}(\overline{c})\|_F \leq \|\dr f_h(c)-\dr f_{\overline{h}}(c)\|_F+\|\dr f_{\overline{h}}(c)-\dr f_{\overline{h}}(\overline{c})\|_F \\
&\leq \|\dr f_h-\dr f_{\overline{h}}\|_{\op}\|c\|_E + \|\dr f_{\overline{h}}\|_{\op} \|c-\overline{c}\|_E \\
&\stackrel{\asref{ass:A4}}{\leq} \zeta\|h-\overline{h}\|_E\delta+ \sup_{h \in \Xfrak} \|\dr f_h\|_{\op} \varepsilon \\
&\leq \varepsilon(\zeta\delta+\sup_{h \in \Xfrak} \|\dr f_h\|_{\op}) \stackrel{\eqref{eq:lipsch}}{\leq} \varepsilon(\zeta\delta+\beta)\,.
\end{split}
\end{equation}
This gives $\Ncal(T_C, \|\cdot\|_F, \varepsilon(\zeta\delta+\beta)) \leq |\overline{\Xfrak}|\times |\overline{C}| \leq \Ncal(\Xfrak, \|\cdot\|_E, \varepsilon) \times \Ncal(C, \|\cdot\|_E, \varepsilon)$.

\end{proof}

We can now prove Proposition \ref{prop:true_cov_number_short_chors} which we recall first.
\shortchords*
\begin{proof}
Consider $C$ and $T_C$ as defined in Corollary \ref{corr:exists_tanget}. We have for any $c \in C, \|c\|_E \leq \frac{1}{\alpha}$ since $\forall u \in S[\Xfrak], \|u\|_E = 1$. So by applying Lemma \ref{lemma:subset} with $\delta = \frac{1}{\alpha}$ we obtain
\begin{equation*}
\begin{split}
&\Ncal(T_C, \|\cdot\|_F,   \varepsilon) \leq \Ncal(\Xfrak, \|\cdot\|_E, \frac{\varepsilon}{\frac{\zeta}{\alpha}+\beta})\times \Ncal(C, \|\cdot\|_E, \frac{\varepsilon}{\frac{\zeta}{\alpha}+\beta})\,.
\end{split}
\end{equation*}
Also, by Corollary \ref{corr:exists_tanget},
\begin{equation}
\forall u \in S^{-}_{\eta}[f(\Xfrak)], \exists t \in T_C, \ \|u-t\|_F \leq \frac{L}{\alpha^{2}} \eta\,.
\end{equation}
We then apply Lemma \ref{lemma:generalized_cover} (with $\delta = \frac{L}{\alpha^{2}} \eta$) to prove that, for any $\varepsilon >0$,
\begin{equation}
\Ncal(S^{-}_{\eta}[f(\Xfrak)], \|\cdot\|_F, 2(\varepsilon+\frac{L}{\alpha^{2}} \eta)) \leq \Ncal(T_C, \|\cdot\|_F,   \varepsilon)\,.
\end{equation}
Consequently,
\begin{equation}
\Ncal(S^{-}_{\eta}[f(\Xfrak)], \|\cdot\|_F, 2(\varepsilon+\frac{L}{\alpha^{2}} \eta)) \leq \Ncal(\Xfrak, \|\cdot\|_E, \frac{\varepsilon}{\frac{\zeta}{\alpha}+\beta})\times \Ncal(C, \|\cdot\|_E, \frac{\varepsilon}{\frac{\zeta}{\alpha}+\beta})\,.
\end{equation}
All we need now is to control $\Ncal(C, \|\cdot\|_F, \frac{\varepsilon}{\frac{\zeta}{\alpha}+\beta})$. Take $\overline{S[\Xfrak]}$ a $\varepsilon$-net of $S[\Xfrak]$ with respect to $\|\cdot\|_E$ and $\overline{(0, \frac{1}{\alpha}]}$ a $(\varepsilon/\alpha)$-net of $(0, \frac{1}{\alpha}]$ with respect to $|\cdot|$. Consider $c = \lambda u \in C$ with $\lambda \in (0,\frac{1}{\alpha}]$ and $u \in S[\Xfrak]$. Then there exists $\overline{u} \in \overline{S[\Xfrak]}$ such that $\|u-\overline{u}\|_E \leq \varepsilon$ and there exists $\overline{\lambda} \in \overline{(0, \frac{1}{\alpha}]}$ such that $|\lambda -\overline{\lambda}| \leq \varepsilon/\alpha$. We consider $\overline{c} = \overline{\lambda} \overline{u}$ which belongs to $C$. Then $\|c-\overline{c}\|_E = \|\lambda u - \overline{\lambda} \overline{u} \|_E \leq |\lambda -\overline{\lambda}| \|\overline{u}\|_E + |\lambda| \|\overline{u}-u\|_E \leq 2 \varepsilon/\alpha $. Thus for any $\varepsilon>0$ we have 
\begin{equation}
\Ncal(C, \|\cdot\|_E, 2 \varepsilon/\alpha) \leq \Ncal(S[\Xfrak], \|\cdot\|_E, \varepsilon) \times \Ncal((0,\frac{1}{\alpha}], |\cdot|, \varepsilon/\alpha) = \Ncal(S[\Xfrak], \|\cdot\|_E, \varepsilon) \times \Ncal((0,1], |\cdot|, \varepsilon)\,. 
\end{equation}
Equivalently with a change of variable $\varepsilon \leftarrow 2 \varepsilon/\alpha$ we have $\forall \varepsilon>0, \Ncal(C, \|\cdot\|_E, \varepsilon) \leq \Ncal(S[\Xfrak], \|\cdot\|_E, \frac{\varepsilon \alpha}{2}) \times \Ncal((0,1], |\cdot|, \frac{\varepsilon \alpha}{2})$. Overall, as for all $\varepsilon > 0, \Ncal((0,1], |\cdot|, \varepsilon) \leq \frac{1}{2 \varepsilon}$,
\begin{equation}
\begin{split}
& \Ncal(S^{-}_{\eta}[f(\Xfrak)], \|\cdot\|_F, 2(\varepsilon+\frac{L}{\alpha^{2}} \eta)) \\
\leq & \left(\frac{\zeta+\beta \alpha}{\alpha^2 \varepsilon}\right) \times \Ncal(\Xfrak, \|\cdot\|_E, \frac{\varepsilon}{\frac{\zeta}{\alpha}+\beta})\times \Ncal(S[\Xfrak], \|\cdot\|_E, \frac{\alpha \varepsilon}{2(\frac{\zeta}{\alpha}+\beta)})\,,
\end{split}
\end{equation}
which concludes the proof.
\end{proof}

\subsubsection{Proof of Lemma \ref{lemma:finvsatisfies} \label{proof:lemma:finvsatisfies}}
\begin{proof}
The proof is based on various computations and the identity 
\begin{equation}
\label{eq:eq_sum_inverse}
\Thetab_1^{-1}+\Thetab_2^{-1}=\Thetab_1^{-1}(\Thetab_1+\Thetab_2)\Thetab_2^{-1}, \qquad \forall (\Thetab_1, \Thetab_2) \in (\Sfrak_{k,a,b}^{-1})^2.
\end{equation}
We will also use
\begin{equation*}
\|\Abf\Bbf\|_\Fro \leq \|\Abf\|_{2 \to 2} \|\Bbf\|_\Fro \quad \text{and} \quad \|\Abf \Bbf\|_\Fro \leq \|\Abf\|_{\Fro} \|\Bbf\|_{2 \to 2} \ , \quad \forall \Abf,\Bbf \in S_d(\R) \,.
\label{eq:fro_op_submult}
\end{equation*}
For the rest of the proof, we take $\Thetab_1, \Thetab_2 \in \Sfrak_{k,a,b}^{-1}$. Hence we have $\|\Thetab_1\|_{2 \to 2} \leq b$ and $\|\Thetab_1^{-1}\|_{2 \to 2} \leq \frac{1}{a}$ (the same inequalities hold for $\Thetab_2$). 

Now we can prove that
\begin{equation*}
\begin{split}
\|\Thetab_1^{-1}-\Thetab_2^{-1}\|_\Lambda & \leq C_\Fro \|\Thetab_1^{-1}(\Thetab_1-\Thetab_2)(-\Thetab_2)^{-1}\|_\Fro \leq C_\Fro \|\Thetab_1^{-1}\|_{2 \to 2} \|\Thetab_2^{-1}\|_{2 \to 2} \|\Thetab_1-\Thetab_2\|_\Fro \\
&\leq  C_\Fro \frac{1}{a^2}\|\Thetab_1-\Thetab_2\|_\Fro \,.
\end{split}
\end{equation*}
This proves \eqref{eq:lipsch} with $\beta = \frac{C_\Fro}{a^2}$. The reverse inequality for \eqref{eq:invlipsch} can be proven by using this time
\begin{equation*}
\begin{split}
\| \Thetab_1-\Thetab_2\|_\Fro &=  \| \Thetab_1 ( \Thetab_1^{-1}-\Thetab_2^{-1} ) \Thetab_2 \|_\Fro  \leq \|\Thetab_1\|_{2 \to 2} \|\Thetab_2\|_{2 \to 2} \|\Thetab_1^{-1}-\Thetab_2^{-1}\|_\Fro \\
&\leq b^2 \frac{1}{c_\Fro}  \|\Thetab_1^{-1}-\Thetab_2^{-1}\|_\Lambda\,.
\end{split}
\end{equation*}
Thus we have \eqref{eq:invlipsch} with $\alpha = \frac{c_\Fro}{ b^2}$.

For assumption \asref{ass:A2}, we use the formula of the differential of the inverse of a matrix $\dr \inv_{\Thetab}[\mathbf{H}]=-\Thetab^{-1}\mathbf{H}\Thetab^{-1}$. We have:
\begin{equation*}
\begin{split}
\|\Thetab_1^{-1}-\Thetab_2^{-1}-\dr \inv_{\Thetab_2}(\Thetab_1-\Thetab_2)\|_\Lambda &=\|\Thetab_1^{-1}-\Thetab_2^{-1}-[-\Thetab_2^{-1}(\Thetab_1-\Thetab_2)\Thetab_2^{-1}]\|_\Lambda \\
&= \|\Thetab_1^{-1}(\Thetab_2-\Thetab_1)\Thetab_2^{-1}-\Thetab_2^{-1}(\Thetab_2-\Thetab_1)\Thetab_2^{-1}\|_\Lambda \\
&=\|\left(\Thetab_1^{-1}(\Thetab_2-\Thetab_1)-\Thetab_2^{-1}(\Thetab_2-\Thetab_1)\right)\Thetab_2^{-1}\|_\Lambda \\
&=\|\left(\Thetab_1^{-1}-\Thetab_2^{-1}\right)(\Thetab_2-\Thetab_1)\Thetab_2^{-1}\|_\Lambda \\
&\leq C_\Fro \|\Thetab_2^{-1}\|_{2 \to 2} \|\Thetab_1^{-1}-\Thetab_2^{-1}\|_\Fro \|\Thetab_1-\Thetab_2\|_\Fro \\
&\leq C_\Fro \|\Thetab_2^{-1}\|_{2 \to 2} \|\Thetab_1^{-1}\|_{2 \to 2} \|\Thetab_2^{-1}\|_{2 \to 2} \|\Thetab_1-\Thetab_2\|_\Fro^{2} \\
&\leq C_\Fro \frac{1}{a^3} \|\Thetab_1-\Thetab_2\|_\Fro^{2}.
\end{split}
\end{equation*}
This gives \asref{ass:A2} with $L = \frac{C_\Fro}{a^3}$.

Now take $\Mbf$, such that $\|\Mbf\|_\Fro \leq 1$. We have
\begin{equation*}
\begin{split}
\|\dr \inv_{\Thetab_1}(\Mbf)-\dr \inv_{\Thetab_2}(\Mbf)\|_\Lambda &= \|\Thetab^{-1}_1 \Mbf \Thetab^{-1}_1 - \Thetab_2^{-1} \Mbf \Thetab_2^{-1} \|_\Lambda \\
&\leq \|\Thetab^{-1}_1 \Mbf \Thetab^{-1}_1 - \Thetab_1^{-1} \Mbf \Thetab_2^{-1} \|_\Lambda + \|\Thetab^{-1}_1 \Mbf \Thetab^{-1}_2 - \Thetab_2^{-1} \Mbf \Thetab_2^{-1} \|_\Lambda \\
&=\|\Thetab^{-1}_1 \Mbf(\Thetab^{-1}_1-\Thetab^{-1}_2)\|_\Lambda + \|(\Thetab^{-1}_1 - \Thetab_2^{-1} )\Mbf \Thetab_2^{-1} \|_\Lambda \\
&\leq C_\Fro \|\Thetab^{-1}_1 \Mbf\|_\Fro \|\Thetab^{-1}_1-\Thetab^{-1}_2\|_\Fro + C_\Fro \|\Mbf\Thetab^{-1}_2 \|_\Fro \|\Thetab^{-1}_1-\Thetab^{-1}_2\|_\Fro \\
&\leq C_\Fro \left( \|\Thetab^{-1}_1\|_{2 \to 2} + \|\Thetab^{-1}_2\|_{2 \to 2} \right)\|\Thetab^{-1}_1-\Thetab^{-1}_2\|_\Fro \\
&\leq 2 C_\Fro \frac{1}{a}\|\Thetab^{-1}_1-\Thetab^{-1}_2\|_\Fro \\
&\leq 2 C_\Fro \frac{1}{a^3} \|\Thetab_1-\Thetab_2\|_\Fro \,.
\end{split}
\end{equation*}
This gives \asref{ass:A4} with $\zeta =  2 \frac{C_\Fro}{a^3} = 2L$.
\end{proof}

\subsubsection{Proof of Lemma \ref{lemma:covering_secant_and_sig} \label{proof:lemma:covering_secant_and_sig}}

In order to prove the result we will use the following lemma.
\begin{restatable}{lemma}{coveringsimple}
\label{lemma:covering_simple}
Consider
\begin{equation}
\mathfrak{W}_k= \left\{\Thetab \in S_{d}(\R): \  \|\Thetab\|_{0} \leq d+2k, \|\Thetab\|_{\Fro} \leq 1\right\}\,.
\end{equation}
Then
\begin{equation}
\Ncal(\mathfrak{W}_k,\|\cdot\|_\Fro,\varepsilon) \leq (\frac{ed^2}{2k})^{k}(\frac{9}{\varepsilon})^{d+k}\,.
\end{equation}
\end{restatable}
\begin{proof}
Take $\Thetab \in \mathfrak{W}_k$, it can be written as $\Thetab= \mathbf{D}+\mathbf{T}+\mathbf{T}^{\top}$ where $\mathbf{D}$ is diagonal with $d$ positive elements, $\mathbf{T}$ is a striclty upper triangular matrix with at most $k$ non zero elements. We have also that $\|\mathbf{D}\|_\Fro \leq \|\Thetab\|_{\Fro} \leq 1$ and same for $\mathbf{T},\mathbf{T}^{\top}$. Consider $\overline{D}$ a $\varepsilon/3$-net for the diagonal and $\overline{T}$ a $\varepsilon/3$-net for the upper triangle, both with respect to the $\|\cdot\|_{\Fro}$ norm. Then with standard covering arguments $|\overline{D}| \leq (9/\varepsilon)^{d}$ and $|\overline{T}| \leq \binom{\frac{d(d-1)}{2}}{k}(\frac{9}{\varepsilon})^{k}$ because it is included in the unit ball of $k$-sparse vector in dimension $\frac{d(d-1)}{2}$ (see \textit{e.g.} \cite{foucart13}).
Consider $\overline{\mathfrak{W}_k}=\{\mathbf{D}_*+\mathbf{T}_*+\mathbf{T}_*^{\top}, (\mathbf{D}_*,\mathbf{T}_*) \in \overline{D}\times \overline{T}\}$. Then $|\overline{\mathfrak{W}_k}| \leq \binom{\frac{d(d-1)}{2}}{k}(\frac{9}{\varepsilon})^{k}(\frac{9}{\varepsilon})^{d}=\binom{\frac{d(d-1)}{2}}{k}(\frac{9}{\varepsilon})^{d+k}$. Also, for any $\Thetab= \mathbf{D}+\mathbf{T}+\mathbf{T}^{\top}$ there exists $(\mathbf{D}_*,\mathbf{T}_*) \in \overline{D}\times \overline{T}$ such that $\|\mathbf{D}-\mathbf{D}_*\|_\Fro\leq \varepsilon/3, \|\mathbf{T}-\mathbf{T}_*\|_\Fro \leq \varepsilon/3$. Hence:
\begin{equation}
\|\mathbf{D}+\mathbf{T}+\mathbf{T}^{\top}-(\mathbf{D}_*+\mathbf{T}_*+\mathbf{T}_*^{\top})\|_\Fro\leq \|\mathbf{D}-\mathbf{D}_*\|_\Fro+2\|\mathbf{T}-\mathbf{T}_*\|_\Fro\leq \varepsilon\,.
\end{equation}
Hence,
\begin{equation}
\Ncal(\mathfrak{W}_k,\|\cdot\|_\Fro,\varepsilon) \leq \binom{\frac{d(d-1)}{2}}{k}(\frac{9}{\varepsilon})^{d+k}\leq (\frac{ed(d-1)}{2k})^{k}(\frac{9}{\varepsilon})^{d+k} \leq (\frac{ed^2}{2k})^{k}(\frac{9}{\varepsilon})^{d+k}\,,
\end{equation}
where in the last inequality we used the bound \cite[Lemma C.5]{foucart13}. Note that we only considered the fact that $\mathfrak{W}$ is the space of symmetric an $d+2k$ sparse matrices. Restricting to positive definite matrices could be an avenue for further improvements.
\end{proof}
As a consequence we can prove Lemma \ref{lemma:covering_secant_and_sig} as follows.
\begin{proof}[Proof of Lemma \ref{lemma:covering_secant_and_sig}]
Recall that
\begin{equation*}
\Sfrak_{k,a,b}^{-1}= \{\Thetab \in S_{d}^{++}(\R) \ : \|\Thetab\|_{0} \leq d+2k, \spec(\Thetab) \subseteq [a,b]\}\,.
\end{equation*}
Consider $\Thetab \in \Sfrak_{k,a,b}^{-1}$. We have $\|\Thetab\|_{2\to 2} \leq b$ which implies $\|\Thetab\|_{\Fro} \leq \sqrt{d} \|\Thetab\|_{2 \to 2} \leq \sqrt{d} b$. Thus $\Sfrak_{k,a,b}^{-1} \subset \{\Thetab \in S_{d}^{++}(\R): \ \ \|\Thetab\|_0 \leq d+2k, \|\Thetab\|_{\Fro} \leq  \sqrt{d} b\}=\sqrt{d} b \mathfrak{W}$. Consequently,
\begin{equation}
\begin{split}
\Ncal(\Sfrak_{k,a,b}^{-1}, \|\cdot\|_{\Fro}, \varepsilon) \leq \Ncal(\mathfrak{W}, \|\cdot\|_{\Fro}, \varepsilon/2\sqrt{d} b) \stackrel{\text{ Lemma \ref{lemma:covering_simple}}}{\leq} (\frac{ed^2}{2k})^{k}(\frac{18 \sqrt{d} b}{\varepsilon})^{d+k}\,,
\end{split}
\end{equation}
which concludes the first part. For the second part we recall the definition of the normalized secant set
\begin{equation*}
S[\Sfrak_{k,a,b}^{-1}]= \{\frac{\Thetab_1-\Thetab_2}{\|\Thetab_1-\Thetab_2\|_\Fro}\ : \ (\Thetab_1,\Thetab_2) \in (\Sfrak_{k,a,b}^{-1})^{2}, \|\Thetab_1-\Thetab_2\|_\Fro >0 \}\,.
\end{equation*}
Moreover, if $\U= \Thetab_1-\Thetab_2 \in \Sfrak_{k,a,b}^{-1}-\Sfrak_{k,a,b}^{-1}$ then $\|\U\|_{0} \leq d + 4k$. Indeed, $\Thetab_1$ and $\Thetab_2$ have $d$ nonzeros elements on the diagonal (since both are postitive definite) and since these matrices are symmetric then they have at most $k$ nonzeros elements in the upper-triangular (\textit{resp.} lower-triangular) part. Thus $\Thetab_1-\Thetab_2$ has at most $2k$ nonzeros elements in the upper-triangular (\textit{resp.} lower-triangular) part. Consequently,
\begin{equation*}
S[\Sfrak_{k,a,b}^{-1}] \subset \{\Mbf \in S_{d}(\R) : \|\Mbf\|_{\Fro} \leq 1, \|\Mbf\|_{0} \leq d + 4k \} = \mathfrak{W}_{2k}\,.
\end{equation*}
This gives
\begin{equation}
\begin{split}
\Ncal(S[\Sfrak^{-1}_{k,a,b}], \|\cdot\|_{\Fro},\varepsilon) &\leq \Ncal(\mathfrak{W}_{2k}, \|\cdot\|_{\Fro},\varepsilon/2) \leq (\frac{ed^2}{4k})^{2k}(\frac{18}{\varepsilon})^{d+2k}\,.
\end{split}
\end{equation}
\end{proof}

\subsubsection{Proof of Corollary~\ref{coro:cov_nb_appli} \label{proof:cov_nb_appli}}

\begin{proof}
From Lemma~\ref{lemma:finvsatisfies}, the assumptions of Theorem~\ref{theorem:big_theo_covering} hold for $f = \inv$, $\Sfrak = \Sfrak_{k,a,b}^{-1}$, $\|\cdot\|_E = \|\cdot\|_\Fro$ and $\|\cdot\|_F = \|\cdot\|_\Lambda$. Thus, for any $\eta, \varepsilon' >0$ we have the inequality 
\begin{equation}
\label{eq:tempoineq_covering}
\begin{split}
  &\Ncal(S[\Sfrak_{k,a,b}], \|\cdot\|_\Lambda, 2\left[\varepsilon'+\frac{L}{\alpha^{2}} \eta\right]) \\
   \leq & \Ncal(\Sfrak_{k,a,b}^{-1},\|\cdot\|_\Fro,\frac{\eta}{8 \beta} \left[\varepsilon'+\frac{L}{\alpha^{2}} \eta\right])^{2} \\
  &+ \frac{\zeta+\beta\alpha}{\alpha^{2} \varepsilon'}\Ncal(\Sfrak_{k,a,b}^{-1}, \|\cdot\|_\Fro, \frac{\alpha\varepsilon' }{\zeta+\beta\alpha}) \times \Ncal(S[\Sfrak_{k,a,b}^{-1}], \|\cdot\|_\Fro, \frac{\alpha^{2} \varepsilon'}{2(\zeta+\beta\alpha)})\,.
  \end{split}
\end{equation}
Let $\varepsilon > 0$ be fixed. We define $\eta \stackrel{\D}{=} \frac{\alpha^2}{4L}\varepsilon$ and $\varepsilon' \stackrel{\D}{=} \varepsilon/4$. Then we have $\varepsilon = 2\left[\varepsilon'+\frac{L}{\alpha^{2}} \eta\right]$. With these $\varepsilon', \eta$ and Lemma~\ref{lemma:covering_secant_and_sig} we obtain 
\begin{equation}
\label{eq:covering_with_good_constant}
\begin{split}
&\Ncal(\Sfrak_{k,a,b}^{-1},\|\cdot\|_\Fro,\frac{\eta}{8 \beta} \left[\varepsilon'+\frac{L}{\alpha^{2}} \eta\right])^{2} 
\leq \left(\frac{ed^2}{2k}\right)^{2k} \left(\frac{288 \beta \sqrt{d} b}{\eta \varepsilon} \right)^{2(d+k)} \\
&\Ncal(\Sfrak_{k,a,b}^{-1}, \|\cdot\|_\Fro, \frac{\alpha\varepsilon }{4(\zeta+\beta\alpha)}) \leq \left(\frac{ed^2}{2k}\right)^{k} (\frac{72 \sqrt{d} b (\zeta +\beta \alpha)}{ \alpha \varepsilon})^{d+k}  \\
& \Ncal(S[\Sfrak_{k,a,b}^{-1}], \|\cdot\|_\Fro, \frac{\alpha^{2} \varepsilon}{8(\zeta+\beta\alpha)}) \leq (\frac{ed^2}{4k})^{2k}(\frac{144 (\zeta + \beta \alpha)}{\alpha^2\varepsilon})^{d+2k}\,.
\end{split}
\end{equation}
Thus, inequality \eqref{eq:tempoineq_covering} yields
\begin{equation*}
\begin{split}
  &  \Ncal\left(S[\Sfrak_{k,a,b}], \|\cdot\|_\Lambda, \varepsilon \right) \\
\stackrel{\eqref{eq:covering_with_good_constant}}{\leq} &  \left(\frac{ed^2}{2k}\right)^{2k} \left(\frac{288 \beta \sqrt{d} b}{\eta \varepsilon} \right)^{2(d+k)} \\
&+ \frac{4(\zeta+\beta\alpha)}{\alpha^{2} \varepsilon}\left(\frac{ed^2}{2k}\right)^{k} \left(\frac{72 b\sqrt{d} (\zeta+\beta\alpha) }{\alpha \varepsilon}\right)^{d+k} \times \left(\frac{ed^2}{4k}\right)^{2k} \left(\frac{144 (\zeta+\beta\alpha) }{\alpha^2 \varepsilon}\right)^{d + 2k}\\
\leq & \left(\frac{ed^2}{2k}\right)^{4k}\left[\left(\frac{288 \beta \sqrt{d} b}{\eta \varepsilon} \right)^{2(d+k)}+\frac{4(\zeta+\beta\alpha)}{\alpha^{2} \varepsilon}\left(\frac{72 b\sqrt{d} (\zeta+\beta\alpha) }{\alpha \varepsilon}\right)^{d+k}\left(\frac{144 (\zeta+\beta\alpha) }{\alpha^2 \varepsilon}\right)^{d + 2k}\right] \\
\end{split}
\end{equation*}
Using the definition $\eta = \frac{\alpha^2}{4L}\varepsilon$ we obtain
\begin{equation*}
\begin{split}
& \Ncal\left(S[\Sfrak_{k,a,b}], \|\cdot\|_\Lambda, \varepsilon \right) \\
\leq & \left(\frac{ed^2}{2k}\right)^{4k}\left[\left(\frac{1152 \sqrt{d} b \beta L}{\alpha^2 \varepsilon^2} \right)^{2(d+k)}+\frac{4(\zeta+\beta\alpha)}{\alpha^{2} \varepsilon}\left(\frac{72 b\sqrt{d} (\zeta+\beta\alpha) }{\alpha \varepsilon}\right)^{d+k}\left(\frac{144 (\zeta+\beta\alpha) }{\alpha^2 \varepsilon}\right)^{d + 2k}\right]\,.
\end{split}
\end{equation*}
Now, from Lemma~\ref{lemma:finvsatisfies}, we have $\beta = \frac{C_\Fro}{a^2},  \quad L = \frac{C_\Fro}{a^3}, \quad \alpha = \frac{c_\Fro}{b^2}, \quad \zeta = 2L = 2\frac{C_\Fro}{a^3}$. So,
\begin{equation}
\label{ineq_covering}
\begin{split}
&\Ncal\left(S[\Sfrak_{k,a,b}], \|\cdot\|_\Lambda, \varepsilon \right) \leq\left(\frac{ed^2}{2k}\right)^{4k} \left[\left(\frac{1152 b \sqrt{d} \frac{C_\Fro}{a^2} \frac{C_\Fro}{a^3}}{\frac{c_\Fro^2}{b^4} \varepsilon^2} \right)^{2(d+k)} \right.\\
&\quad \left. +\frac{4(2\frac{C_\Fro}{a^3}+\frac{C_\Fro}{a^2}\frac{c_\Fro}{b^2})}{\frac{c_\Fro^2}{b^4}\varepsilon}\left(\frac{72 b\sqrt{d} (2\frac{C_\Fro}{a^3}+\frac{C_\Fro}{a^2}\frac{c_\Fro}{b^2}) }{\frac{c_\Fro}{b^2} \varepsilon}\right)^{d+k}\left(\frac{144 (2\frac{C_\Fro}{a^3}+\frac{C_\Fro}{a^2}\frac{c_\Fro}{b^2}) }{\frac{c_\Fro^2}{b^4} \varepsilon}\right)^{d+2k}\right]\\
&=\left(\frac{ed^2}{2k}\right)^{4k}\left[\left(\frac{1152 \sqrt{d}}{\varepsilon^2} \frac{C_\Fro^2}{c_\Fro^2} \frac{b^5}{a^5} \right)^{2(d+k)}\right.\\
& \quad \left.+\frac{4}{\varepsilon}\frac{C_\Fro}{c_\Fro}(\frac{2}{c_\Fro}\frac{b^4}{a^{3}}+\frac{b^2}{a^2})\left(\frac{72}{\varepsilon}\frac{C_\Fro}{c_\Fro}(2\frac{b^3}{a^3}+\frac{b}{a^2}c_\Fro)\sqrt{d}\right)^{d+k}\left(\frac{144}{\varepsilon}\frac{C_\Fro}{c_\Fro}(\frac{2}{c_\Fro}\frac{b^4}{a^{3}}+\frac{b^2}{a^2})\right)^{d+2k}\right]\,. \\
\end{split}
\end{equation}
We will simplify this expression using the homogeneity of the normalized secant set. More precisley if $\frac{\Sigmab_1-\Sigmab_2}{\|\Sigmab_1-\Sigmab_2\|_\Lambda} \in S[\Sfrak_{k,a,b}]$ then for any $t > 0, \frac{t\Sigmab_1-t\Sigmab_2}{\|t\Sigmab_1-t\Sigmab_2\|_\Lambda} \in S[\Sfrak_{k,a,b}]$. This implies that $\forall t >0, \ S[\Sfrak_{k,a,b}] = S[\Sfrak_{k, t \cdot a,t \cdot b}]$. In particular for $t \stackrel{\D}{=} \frac{a}{b^2}$ the previous expression \eqref{ineq_covering} gives
\begin{equation*}
\begin{split}
& \Ncal\left(S[\Sfrak_{k,a,b}], \|\cdot\|_\Lambda, \varepsilon \right) \\
=& \Ncal\left(S[\Sfrak_{k,t \cdot a,t \cdot b}], \|\cdot\|_\Lambda, \varepsilon \right) \\
\leq & \left(\frac{ed^2}{2k}\right)^{4k}\left[\left(\frac{1152 \sqrt{d}}{\varepsilon^2} \frac{C_\Fro^2}{c_\Fro^2} \frac{b^5}{a^5} \right)^{2(d+k)}\right.\\
& \left. +\frac{4}{\varepsilon}\frac{C_\Fro}{c_\Fro}(\frac{2}{c_\Fro}t\frac{b^4}{a^{3}}+\frac{b^2}{a^2})\left(\frac{72}{\varepsilon}\frac{C_\Fro}{c_\Fro}(2\frac{b^3}{a^3}+\frac{b}{t a^2}c_\Fro)\sqrt{d}\right)^{d+k}\left(\frac{144}{\varepsilon}\frac{C_\Fro}{c_\Fro}(\frac{2}{c_\Fro}t\frac{b^4}{a^{3}}+\frac{b^2}{a^2})\right)^{d+2k} \right] \\
= & \left(\frac{ed^2}{2k}\right)^{4k}\left[\left(\frac{1152 \sqrt{d}}{\varepsilon^2} \frac{C_\Fro^2}{c_\Fro^2} \frac{b^5}{a^5} \right)^{2(d+k)}\right.\\
& \left. +\frac{4}{\varepsilon}\frac{C_\Fro}{c_\Fro}(\frac{2}{c_\Fro}+1)\frac{b^2}{a^2}\left(\frac{72}{\varepsilon}\frac{C_\Fro}{c_\Fro}(2+c_\Fro)\frac{b^3}{a^3}\sqrt{d}\right)^{d+k}\left(\frac{144}{\varepsilon}\frac{C_\Fro}{c_\Fro}(\frac{2}{c_\Fro}+1)\frac{b^2}{a^2}\right)^{d+2k} \right] \\ 
\leq & \left(\frac{ed^2}{2k}\right)^{4k}\left[\left(\frac{1152 \sqrt{d}}{\varepsilon^2} \frac{C_\Fro^2}{c_\Fro^2} \frac{b^5}{a^5} \right)^{2(d+k)} \right.\\
& \left. +\left(\frac{72}{\varepsilon}\frac{C_\Fro}{c_\Fro}(2+c_\Fro)\frac{b^3}{a^3}\sqrt{d}\right)^{d+k}\left(\frac{144}{\varepsilon}\frac{C_\Fro}{c_\Fro}(\frac{2}{c_\Fro}+1)\frac{b^2}{a^2}\right)^{d+2k+1}\right] \,.
\end{split}
\end{equation*}
Now, to simplify the expression, remark that $\frac{72}{\varepsilon}\frac{C_\Fro}{c_\Fro}(2+c_\Fro)\frac{b^3}{a^3}\sqrt{d} \geq 1$ as $C_\Fro/c_\Fro \geq 1$. Therefore we can increase its power from $d+k$ to $d+2k+1$ to match with the other multiplicative term. This yields 
\begin{equation*}
\begin{split}
& \Ncal\left(S[\Sfrak_{k,a,b}], \|\cdot\|_\Lambda, \varepsilon \right) \\
 \leq &\left(\frac{ed^2}{2k}\right)^{4k}\left[\left(\frac{1152 C_\Fro^2 \sqrt{d}}{\varepsilon^2 c_\Fro^2} \frac{b^5}{a^5} \right)^{2(d+k)} +\left(\frac{72 \times 144 C_\Fro^2 \sqrt{d}}{\varepsilon^2 c_\Fro}(\frac{2}{c_\Fro}+1)\frac{b^5}{a^5}\right)^{d+2k+1}\right] \\
= & \left(\frac{ed^2}{2k}\right)^{4k}\left[\left(c_0\frac{\sqrt{d}C_\Fro^2}{\varepsilon^2c_\Fro^2}  \frac{b^5}{a^5} \right)^{2(d+k)} +\left(c_1 \frac{\sqrt{d} C_\Fro^2 }{\varepsilon^2 c_\Fro}(\frac{2}{c_\Fro}+1)\frac{b^5}{a^5}\right)^{d+2k+1}\right] \,,
\end{split}
\end{equation*}
where $c_0, c_1$ are absolute constants greater than 1. 
This concludes the proof.
\end{proof}

\subsection{Proofs of Section \ref{sec:rank_one}}
\label{proof:rank_one}

\subsubsection{Proofs of the rank-one projection operator properties \label{proof:RO}}

The goal of this section is to prove Proposition~\ref{prop:C2forROP} and \ref{prop:lambda_norm_controls}. Before that, we prove several results that will become handy afterwards. Firstly, we state a result that will be usefull to leverage results from the Gaussian case to the uniform case.
\begin{lemma}
Let $\ubf$ and $\rho$ be independent variables with the following distributions : $\ubf \sim \mathcal{U}(\mathbb{S}^{d-1})$ is a uniform vector on the hyper-sphere and $\rho^2 \sim \chi^2(d)$ is a chi-square variable with $d$ degrees of freedom.
Then, $\rho \frac{\ubf}{\sqrt{d}} \sim \mathcal{N}(0, \frac{1}{d}\mathbf{I}_d)$ is a standard normal vector.
\label{lem:unif_gauss_link}
\end{lemma}

Now, a lower bound is derived for the $\Lambda$-norm.
\begin{proposition}
\label{prop:Lambda_bound}
For any $\Mbf \in S_d(\R)$ and $\Lambda \in \{\Lambda_{\operatorname{G}}, \Lambda_{\operatorname{U}}\}$, we have
\begin{equation}
\| \Mbf \|_\Lambda \geq \frac{2}{9 \sqrt{15}d} \left( \| \Mbf \|_\Fro + |\tr(\Mbf)| \right) \,.
\label{eq:Lambda_bound}
\end{equation}
\end{proposition}

\begin{proof}
We use the fact that for any real random variable $X$, whenever its fourth moment exists, we have\footnote{It comes from applying the Hölder inequality to $\expect{}{|X|^{2/3} |X|^{4/3}}$ with $1/p = 2/3$ and $1/q = 1/3$.}
\begin{equation*}
\expect{}{|X|} \geq \sqrt{\frac{\expect{}{X^2}^3}{\expect{}{X^4}}} \,.
\end{equation*}
First, we focus on the Gaussian case. Using\footnote{The last equality is obtained from the rotation invariance of the multivariate normal distribution.} $X = \a^\top \Mbf \a \overset{(d)}{=} \sum \lambda_k b_k^2$, where $\a \sim \mathcal{N}(0,\frac{1}{d}I_d)$, the $(\lambda_k)$ are the eigenvalues of $\Mbf$ and the $(b_k)$ are \emph{i.i.d} Gaussian random variables of variance $1/d$, we can obtain the following bounds:
\begin{equation*}
\expect{}{X^2} = \frac{2}{d^2}\|\Mbf\|_\Fro^2 + \frac{1}{d^2} \tr(\Mbf)^2 \geq \frac{2}{3d} \left(  \| \Mbf \|_\Fro + |\tr(\Mbf)|  \right)^2\,,
\end{equation*}
\begin{align*}
\expect{}{X^4} = & \ \sum\limits_{i} \lambda_i^4 \expect{}{b_i^8} + \sum\limits_{i \neq j} \lambda_i \lambda_j^3 \expect{}{b_i^2} \expect{}{b_j^6} + \sum\limits_{i \neq j} \lambda_i^2 \lambda_j^2 \expect{}{b_i^4} \expect{}{b_j^4} \\
& \quad + \sum\limits_{i \neq j \neq k} \lambda_i \lambda_j \lambda_k^2 \expect{}{b_i^2} \expect{}{b_j^2} \expect{}{b_k^4} + \sum\limits_{i \neq j \neq k \neq l} \lambda_i \lambda_j \lambda_k \lambda_l \expect{}{b_i^2} \expect{}{b_j^2} \expect{}{b_k^2} \expect{}{b_l^2} \\
= & \ \frac{105}{d^4} \sum\limits_{i} \lambda_i^4 + \frac{15}{d^4} \sum\limits_{i \neq j} \lambda_i \lambda_j^3 + \frac{9}{d^4} \sum\limits_{i \neq j} \lambda_i^2 \lambda_j^2 + \frac{3}{d^4} \sum\limits_{i \neq j \neq k} \lambda_i \lambda_j \lambda_k^2 + \frac{1}{d^4} \sum\limits_{i \neq j \neq k \neq l} \lambda_i \lambda_j \lambda_k \lambda_l \\
= & \ \frac{1}{d^4} \left[ 90 \|\lambda\|_4^4 + 12 \tr(\Mbf) \sum\limits_i \lambda_i^3 + 6 \|\Mbf\|_\Fro^4 + 2 \tr(\Mbf)^2 \|\Mbf\|_\Fro^2 + \tr(\Mbf)^4 \right] \\
\leq & \ \frac{90}{d^4} \left( \| \Mbf \|_\Fro + |\tr(\Mbf)| \right)^4 \,.
\end{align*}
This yields equation \eqref{eq:Lambda_bound} for the Gaussian case.

For the uniform case, considering the independent random variables $\ubf \sim \mathcal{U}(\mathbb{S}^{d-1})$, $\rho^2 \sim \chi^2(d)$, from Lemma~\ref{lem:unif_gauss_link} we have that
\begin{equation*}
\expect{}{ \left| \a^\top \Mbf \a \right| } = \expect{}{\left| (\rho \frac{1}{\sqrt{d}} \ubf)^\top \Mbf (\rho \frac{1}{\sqrt{d}} \ubf) \right| }
= \frac{1}{d} \expect{}{\rho^2} \expect{}{ \left|\ubf^\top \Mbf \ubf \right| }
= \expect{}{ \left|\ubf^\top \Mbf \ubf \right| } \,.
\end{equation*}
So \eqref{eq:Lambda_bound} also holds in the uniform case.
\end{proof}

The proof of Proposition~\ref{prop:C2forROP} is based on a concentration inequality for subexponential variables.  Here, we prove that the variables at play are indeed subexponential by providing an upper-bound on their subexponenital norm. Recall that for a random variable $X$, its subexponential norm is defined by $\|X\|_{\psi_1} = \inf \{ s>0, \ \expect{}{e^{|X|/s}} \leq 2\}$.

\begin{proposition}
For any  $\Mbf \in S_d(\R)$ and for $\a \sim \mathcal{N}(0,\frac{1}{d} I_d)$ and $\ubf \sim \mathcal{U}( \mathbb{S}^{d-1})$, the following controls hold :
  \begin{equation}
\| |\a^\top \Mbf \a|-\expect{}{|\a^\top \Mbf \a|} \|_{\psi_1} \leq \frac{2}{d \log 2} ( \frac{76}{9} e^2 \cdot \|\Mbf\|_\Fro + | \tr(\Mbf)| )\,,
\label{eq:subexp_bound_w_cst_GRO}
\end{equation}
\begin{equation}
\| |\ubf^\top \Mbf \ubf|-\expect{}{|\ubf^\top \Mbf \ubf|} \|_{\psi_1} \leq\frac{8e}{d \log 2} ( \frac{76}{9} e^2 \cdot \|\Mbf\|_\Fro + | \tr(\Mbf)| )\,.
\label{eq:subexp_bound_w_cst_URO}
\end{equation}
\end{proposition}

\begin{proof}[Proof of Equation~\eqref{eq:subexp_bound_w_cst_GRO}]
This proof revolves around the different characterizations of subexponentiality (indexed from (\emph{a}) to (\emph{e})) presented in Proposition~2.7.1 of \cite{vershynin2018high}.
First from the centering Lemma (see Exercise~2.7.10 in \cite{vershynin2018high}), we have the existence of a constant $C_1>0$ such that $\| |\a^\top \Mbf \a|-\expect{}{|\a^\top \Mbf \a|} \|_{\psi_1} \leq C_1 \| \ |\a^\top \Mbf \a| \ \|_{\psi_1} = C_1 \|\a^\top \Mbf \a \|_{\psi_1}$.
Let us denote by $X$ our random variable of interest $X = \a^\top \Mbf \a$, and $Y$ its centered version $Y = X - \expect{}{X}$. Working with $Y$, we now characterize the constant $K_5(Y)$ appearing in statement (\emph{e}) of Proposition~2.7.1 in \cite{vershynin2018high}.
First, we can write $Y = \sum_k \lambda_k d^{-1} (z_k^2 - 1)$ where $\lambda_k$ are the eigenvalues of $\Mbf$ and $z_k$ is a standard normal variable. 
Remark that the centered $\chi^2(1)$ variables $z_k^2-1$ verify (\emph{e}) with a certain constant $K_{\chi^2}$. For $|t| \leq d / ( K_{\chi^2} \cdot \|\Mbf\|_\Fro)$ we have that
\begin{equation*}
\expect{}{e^{tY}} = \prod\limits_{k=1}^d \expect{}{e^{t \frac{\lambda_k}{d} (z_k^2-1)}} \leq \prod\limits_{k=1}^d \expect{}{e^{\frac{1}{d^2} K_{\chi^2}^2 \lambda_k^2 t^2 }} = e^{\frac{1}{d^2}K_{\chi^2}^2 \ \|\Mbf\|_\Fro^2 t^2 }\,.
\end{equation*}
This yield $K_5(Y) \leq \frac{1}{d} K_{\chi^2} \ \|\Mbf\|_\Fro$. Now, by considering statement (\emph{c}), we have that
$K_3(X) \leq K_3(Y) + |\expect{}{X}| \leq C_{3,5} K_{\chi^2} \cdot \frac{1}{d} \|\Mbf\|_\Fro + \frac{1}{d} | \tr(\Mbf)|$ (where $C_{3,5}$ is the universal constant allowing to pass from (\emph{c}) to (\emph{e})). Finally, gathering up the pieces we have
\begin{equation*}
\| |\a^\top \Mbf \a|-\expect{}{|\a^\top \Mbf \a|} \|_{\psi_1} \leq \frac{C_1 C_{4,3}}{d} ( C_{3,5} K_{\chi^2} \cdot \|\Mbf\|_\Fro + | \tr(\Mbf)| )\,,
\end{equation*}
where $C_{4,3}$ is the constant allowing to pass from (\emph{d}) (statement defining the $\psi_1$-norm) to (\emph{c}). To conclude the proof, it suffices to find the values of the various constants. This is completely general and does not depend on the rank-one projection considered here. The various constants can be set as follows :
\begin{equation*}
\begin{aligned}[c]
C_1 &= 2 \,, \\
C_{4,3}  &= \frac{1}{\log 2} \,,
\end{aligned}
\qquad \qquad
\begin{aligned}[c]
C_{3,5} &= 4e^2 \,, \\
K_{\chi^2} &= \frac{19}{9} \,.
\end{aligned}
\end{equation*}

Let us start by computing $C_1$. Let $X$ be a subexponential random variable. We have, for any $s>0$,
\begin{equation*}
\expect{}{e^{\frac{||X| - \expect{}{|X|}|}{s}}} \stackrel{(\text{$\triangle$ ineq.})}{\leq} \expect{}{e^{\frac{|X|}{s}}}  e^{\frac{\expect{}{|X|}}{s}} \stackrel{(\text{Jensen})}{\leq} \expect{}{e^{\frac{|X|}{s}}}^2 \stackrel{(\text{Jensen})}{\leq} \expect{}{e^{\frac{2|X|}{s}}} \,.
\end{equation*}
The last term is smaller than 2 for $s/2 \geq \|X\|_{\psi_1}$, so we have that $\| |X| - \expect{}{|X|} \|_{\psi_1} \leq 2 \|X\|_{\psi_1} $, yielding $C_1 =2$. 

For the constant $C_{3,5}$ we use that $C_{3.5} \leq C_{3,2} C_{2,5}$. In \cite{vershynin2018high}, the value of $C_{2,5}$ is given and equals $2e$. Let us focus on $C_{3,2}$ and assume that $K_2(X) = 1$. For any $\lambda$ such that $0 \leq \lambda \leq 1/(2e)$, we have the following inequalities
\begin{equation*}
\expect{}{e^{\lambda |X|}} = 1 + \sum\limits_{p\geq1} \frac{\lambda^p \expect{}{|X|^p}}{p!} \stackrel{K_2(X)=1}{\leq} 1 + \sum\limits_{p\geq1} \frac{\lambda^p p^p}{p!} \leq 1 + \sum\limits_{p\geq1} \frac{\lambda^p p^p}{(p/e)^p} = \frac{1}{1-\lambda e} \leq e^{2e \lambda} \,.
\end{equation*}
Thus, $K_3(X) \leq 2e$. So we can take $C_{3,5} = 4e^2$. 

For $C_{4,3}$ assume that $K_3(X) = 1$. For $\lambda \leq \log 2$ we have
\begin{equation*}
\expect{}{e^{\lambda |X|}} \stackrel{K_3(X)=1}{\leq} e^{\lambda} \leq 2 \,.
\end{equation*}
So we can take $C_{4,3} = 1 / \log 2$. 

The value of $K_{\chi^2}$ can be obtain from the following computation. Let $b \sim \mathcal{N}(0,1)$ be a standard normal distribution, then for $\lambda<1/2$
\begin{equation*}
\expect{}{e^{\lambda(b^2-1)}} = \frac{e^{-\lambda}}{\sqrt{1-2\lambda}} \,.
\end{equation*}
We can show that for $K = \frac{2}{1-x_0}$ where $x_0$ is the smallest solution of $e^{x} = e^{3}x$, we have for all $\lambda$ such that $|\lambda| \leq 1/K$, $\expect{}{e^{\lambda(b^2-1)}} \leq e^{K^2 \lambda^2}$. A numerical approximation gives $K \simeq 2.1107\dots$. So we can take $K_{\chi^2} = 19/9$. This finishes the proof.
\end{proof}

\begin{proof}[Proof of Equation~\eqref{eq:subexp_bound_w_cst_URO}.]
  As in the proof of \eqref{eq:subexp_bound_w_cst_GRO}, we start by decentering : $ \| |\ubf^\top \Mbf \ubf| - \expect{}{|\ubf^\top \Mbf \ubf|}\|_{\psi_1} \leq C_1 \| \ubf^\top \Mbf \ubf \|_{\psi_1}$. Then, recall (see \cite{vershynin2018high}) that there exists an absolute constant $C_{4,2}$ such that $\|\ubf^\top \Mbf \ubf\|_{\psi_1} \leq C_{4,2} K_2(\ubf^\top \Mbf \ubf)$ where $K_2(\ubf^\top \Mbf \ubf)$ is the smallest constant $K$ such that for all  $p \geq 1$, $\expect{}{|\ubf^\top \Mbf \ubf|^p} \leq K^p p^p$. From Lemma~\ref{lem:unif_gauss_link}, we have that

  \begin{equation*}
  \expect{}{|\a^\top \Mbf \a|^p} = \frac{1}{d^p} \expect{}{\rho^{2p}} \expect{}{|\ubf^\top \Mbf \ubf|^p} = \frac{d (d+2) \dots (d+2(p-1))}{d^p} \expect{}{|\ubf^\top \Mbf \ubf|^p} \geq \ \expect{}{|\ubf^\top \Mbf \ubf|^p}.
  \end{equation*}
  Hence, we have that $ K_2(\ubf^\top \Mbf \ubf) \leq K_2(\a^\top \Mbf \a)$.
  Then, using \eqref{eq:subexp_bound_w_cst_GRO} and the existence of a constant $C_{2,4}$ such that $K_2(\a^\top \Mbf \a) \leq C_{2,4} \|\a^\top \Mbf \a\|_{\psi_1}$. We have
  \begin{equation*}
  \| \ubf^\top \Mbf \ubf \|_{\psi_1} \leq C_{4,2} C_{2,4}  \|\a^\top \Mbf \a\|_{\psi_1} \leq  \frac{C_{4,2} C_{2,4} C_{4,3}}{d} ( C_{3,5} K_{\chi^2} \cdot \|\Mbf\|_\Fro + | \tr(\Mbf)| ) \,.
  \end{equation*}
  To finish the proof, we show that $C_{4,2}$ and  $C_{2,4}$ can be chosen as
  \begin{equation*}
    C_{4,2} = 2e \,, \qquad \qquad  C_{2,4} = 2.
  \end{equation*}

  For $C_{4,2}$, we need to prove that $\|X\|_{\psi_1} \leq C_{4,2} K_2(X)$, for any subexponential variable $X$. Without loos of generality, we can always assume that $K_2(X)=1$. Thus, for $s>e$ we have
  \begin{align*}
  \expect{}{e^{|X|/s}} &= 1 + \sum\limits_{k=1}^\infty \frac{\expect{}{|X|^k}}{k! \ s^k} \\
  & \leq 1 + \sum\limits_{k=1}^\infty \frac{k^k}{k! s^k} \stackrel{(\star)}{\leq} \sum\limits_{k=0}^\infty \left(\frac{e}{s} \right)^k = \frac{1}{1-e/s} \,,
  \end{align*}
  where the $(\star)$ inequality comes from the Stirling approximation $k! \geq (k/e)^k$. Thus, for $s \geq 2e$ we have $\expect{}{e^{|X|/s}} \leq 2$. So we can take $C_{4,2} = 2e$.

  For $C_{2,4}$, assume that $\|X\|_{\psi_1} = 1$. Then, for any $p \geq 1$,
  \begin{align*}
  \expect{}{|X|^p} &= \int\limits_0^\infty \proba{}{|X|^p>u} du = \int\limits_0^\infty \proba{}{|X|>t} pt^{p-1} dt \\
  &\leq \int\limits_0^\infty \expect{}{e^{|X|}} e^{-t} p t^{p-1} dt \leq 2 \int\limits_0^\infty e^{-t} p t^{p-1} dt = 2 p! \leq (2p)^p\,.
  \end{align*}
  Thus, we can take $C_{2,4}=2$.
\end{proof}

All this previous results allow now to prove Proposition~\ref{prop:C2forROP}, which is recalled below.
\concentrationForROP*
\begin{proof}[Proof of Proposition~\ref{prop:C2forROP}]
  From our previous results, we have :
\begin{equation}
  \| |\a^\top \U \a|-\expect{}{|\a^\top \U \a|} \|_{\psi_1} \stackrel{\eqref{eq:subexp_bound_w_cst_GRO}}{\leq} \frac{2\times76 \ e^2}{9 d \log 2} ( \|\U\|_\Fro + | \tr(\U)| ) \stackrel{\eqref{eq:Lambda_bound}}{\leq} \frac{76 e^2 \sqrt{15}}{\log 2} \|\U\|_{\Lambda} \stackrel{\|\U\|_{\Lambda}=1}{\leq} \frac{76 e^2 \sqrt{15}}{\log 2} \,.
\end{equation}
Similarly, in the uniform case, we obtain $\| |\ubf^\top \U \ubf|-\expect{}{|\ubf^\top \U \ubf|} \|_{\psi_1} \leq  \frac{304 e^3 \sqrt{15}}{\log 2}$. Therefore, in both cases, the subexponential norm is bounded by an absolute constant that will be denoted by $\ssexpo$ in the following. Now, let us recall a Bernstein-type concentration inequality for sum of subexponential variables.
\begin{lemma}[Proposition 5.16 \cite{vershynin2012}]
Let $X_1, \ldots, X_m$ be independent centered sub-exponential random variables, and $K=\max _i\left\|X_i\right\|_{\psi_1}$. Then for every $\gamma = \left(\gamma_1, \ldots, \gamma_m\right) \in \mathbb{R}^m$ and every $t \geq 0$, we have
\begin{equation*}
\proba{}{\left|\sum_{i=1}^m \gamma_i X_i\right| \geq t } \leq 2 \exp \left(-c \min \left(\frac{t^2}{K^2\|\gamma\|_2^2}, \frac{t}{K\|\gamma\|_{\infty}}\right)\right)\,,
\end{equation*}
where $c=\frac{1}{8e^2}$. (The value of $c$ can be tracked through the proofs of Lemma~5.15 and Proposition~5.16 in \cite{vershynin2012}.)
\end{lemma}

Taking $X_i = |\a_i^\top \U \a_i|-\expect{}{|\a_i^\top \U \a_i|} =  |\a_i^\top \U \a_i|-1$ (or $X_i = |\ubf_i^\top \U \ubf_i|-1$ in the uniform case), and $\gamma_i = 1/m$ for all $i$, yields
\begin{equation*}
\mathbb{P} \left( | \|\Acal (\U)\|_1 - 1 | > t \right) \leq 2 \exp \left(- \frac{m}{8e^2} \min \left(\frac{t^2}{\ssexpo^2}, \frac{t}{\ssexpo}\right)\right), \qquad \forall t \geq 0 \,.
\end{equation*}
\end{proof}

\subsubsection{Proof of Proposition~\ref{prop:lambda_norm_controls} \label{proof:prop:lambda_norm_controls}}
\begin{proof}
The lower bound is a direct consequence of Proposition~\ref{prop:Lambda_bound}. Let us prove the upper bound. In the Gaussian case, the following inequalities hold
\begin{equation*}
  \expect{}{ | \a^\top \Mbf \a | } = \expect{}{ \left| \sum\limits_{k=1}^d \lambda_k b_k^2 \right|  } \leq \sum_{k=1}^d  |\lambda_k| \expect{}{b_k^2} = \frac{1}{d} \sum_{k=1}^d |\lambda_k| \leq \frac{\sqrt{d}}{d} \sqrt{\sum_{k=1}^d \lambda_k^2} = \frac{1}{\sqrt{d}} \|\Mbf\|_\Fro \,,
\end{equation*}
where $\a,\mathbf{b} \sim \Ncal(0, \frac{1}{d} \mathbf{I}_d)$. Similarly, in the URO case,
\begin{equation*}
  \expect{}{ | \ubf^\top \Mbf \ubf | } \leq \sum_{k=1}^d |\lambda_k| \expect{}{v_k^2} =  \frac{1}{d} \sum_{k=1}^d |\lambda_k| \leq \frac{1}{\sqrt{d}} \|\Mbf\|_\Fro,
\end{equation*}
where $\ubf,\vbf \sim \Ucal( \Sbb^{d-1})$.
\end{proof}

\subsubsection{Proof of Theorem~\ref{theo:theRIPRO}}
\label{proof:theRIPRO}

\begin{proof}
From Theorem~\ref{theo:rip_cond2} and Remark~\ref{rem:C1fromC2}, we know that the probability that the operator $\Acal$ \emph{does not} satisfy the $\RIP_\delta$ is upper-bounded by

\begin{equation}
  \Ncal(S[\Sfrak_{k,a,b}],\|\cdot\|_{\Lambda},\varepsilon)C_2(\frac{\delta}{2}) + \Ncal( B_{\Lambda}, \|\cdot\|_{\Lambda}, \varepsilon' ) C_2( (1-\varepsilon') \frac{\delta}{2 \varepsilon} - 1 ), \qquad \forall \varepsilon, \varepsilon'>0 \,,
 \label{eq:proba_bound}
\end{equation}
where $B_{\Lambda} = \{ \U \in S_d(\R) : \ \|\U\|_\Lambda = 1 \}$ and $C_2(t) = 2 \exp ( -\frac{m}{8e^2} \min( t/\ssexpo , (t/\ssexpo)^2 ) )$ (see Proposition~\ref{prop:C2forROP}).
In the following, we choose $\varepsilon'=1/2$. Given $\rho$, the strategy is to find an $\varepsilon$ small enough so that the right handside term is smaller that $\rho / 2$. Then, a condition on $m$ will be derived to ensure that the left handside term is also smaller than $\rho/2$.

First of all, notice that, from standard covering argument, $\Ncal( B_{\Lambda}, \|\cdot\|_{\Lambda}, 1/2 ) \leq (3/(1/2))^{d(d+1)/2}$. Therefore, looking at the logarithm of the second term in \eqref{eq:proba_bound}, $\varepsilon$ should verify
\begin{equation}
\label{eq_with_the_min}
\frac{d (d+1)}{2} \log(6) + \log 2 - \frac{m}{8e^2} \min \left[ \frac{1}{\ssexpo} \left(\frac{\delta}{4 \varepsilon}-1\right), \frac{1}{\ssexpo^2} \left(\frac{\delta}{4 \varepsilon}-1\right)^2 \right] \leq \log(\rho/2)\,.
\end{equation}
Assuming that $\varepsilon \leq \frac{\delta}{4(\ssexpo+1)}$ to ensure that the minimum in the above expression is $\frac{1}{\ssexpo} \left(\frac{\delta}{4 \varepsilon}-1\right)$, \eqref{eq_with_the_min} is equivalent to
\begin{equation*}
  \varepsilon \leq \frac{\delta}{4} \left[ \frac{8e^2\ssexpo}{m} \left( \frac{d (d+1)}{2} \log(6) + \log 2 + \log(2/\rho) \right) + 1 \right]^{-1}.
\end{equation*}
In order to remove the dependency in $m$ and to simplify the expression, notice that it is sufficient to take
\begin{equation*}
  \varepsilon \leq \frac{\delta}{32 e^2 \ssexpo} \left[  d^2 \log(6) + 4\log(2/\rho) \right]^{-1} \stackrel{\D}{=} \varepsilon_0 \,.
\end{equation*}
In the following, we take $\varepsilon = \varepsilon_0$. Remark that it satisfies the assumption below \eqref{eq_with_the_min}. In particular, note that $\varepsilon \leq 1$. We now focus on the first term in \eqref{eq:proba_bound}. Notice that from Corollary~\ref{coro:cov_nb_appli} and Proposition~\ref{prop:lambda_norm_controls} giving $c_\Fro = 2/(9\sqrt{15}d)$ and $C_\Fro=1/\sqrt{d}$, the covering number is controlled as follows:
\begin{equation*}
  \begin{split}
         & \log \Ncal(S[\Sfrak_{k,a,b}],\|\cdot\|_{\Lambda},\varepsilon) \\
    \leq & 4k \log(\frac{ed^2}{2k}) + \log \left[\left(c_0\frac{\sqrt{d} C_\Fro^2}{\varepsilon^2 c_\Fro^2} \frac{b^5}{a^5} \right)^{2(d+k)} +\left(c_1 \frac{\sqrt{d} C_\Fro^2}{\varepsilon^2 c_\Fro}(\frac{2}{c_\Fro}+1)\frac{b^5}{a^5}\right)^{d+2k+1}\right] \\
    \leq & 4k \log(\frac{ed^2}{2k}) + \log \left[\left(\frac{c'_0 d^{3/2}}{\varepsilon^2}\frac{b^5}{a^5} \right)^{2(d+k)}+\left(\frac{c'_1 d^{3/2}}{\varepsilon^2}\frac{b^5}{a^5}\right)^{d+2k+1}  \right]\,.
  \end{split}
\end{equation*}
where $c_0,c_1,c'_0, c'_1$ are absolute constants greater than $1$. As $\frac{c'_1 d^{3/2}}{\varepsilon^2}\frac{b^5}{a^5} \geq 1$, $\left(\frac{c'_0 d^{3/2}}{\varepsilon^2}\frac{b^5}{a^5} \right)^{2(d+k)}+\left(\frac{c'_1 d^{3/2}}{\varepsilon^2}\frac{b^5}{a^5}\right)^{d+2k+1} \leq \left(\frac{c'_0 d^{3/2}}{\varepsilon^2}\frac{b^5}{a^5} \right)^{2(d+k)}+\left(\frac{c'_1 d^{3/2}}{\varepsilon^2}\frac{b^5}{a^5}\right)^{2(d+k)}  \leq \left(\frac{c'_0 d^{3/2}}{\varepsilon^2}\frac{b^5}{a^5}  + \frac{c'_1 d^{3/2}}{\varepsilon^2}\frac{b^5}{a^5}\right)^{2(d+k)}$. This gives
\begin{equation*}
  \begin{split}
         \log \Ncal(S[\Sfrak_{k,a,b}],\|\cdot\|_{\Lambda},\varepsilon) &\leq 4k \log(\frac{ed^2}{2k}) + 2(d+k)\log \left[(c'_0 +c'_1) \frac{d^{3/2}}{\varepsilon^2} \frac{b^5}{a^5}\right] \\
         &=4k \log(\frac{ed^2}{2k}) + 4(d+k)\log \left[c_{\frac{b}{a}} \frac{d}{\varepsilon}\right]\,,
  \end{split}
\end{equation*}
where
\begin{equation*}
  c_{\frac{b}{a}} \stackrel{\D}{=}  \sqrt{(c'_0+c'_1)}\frac{b^{5/2}}{a^{5/2}} \text{ only depends on } b/a \,.
\end{equation*}
Therefore, $m$ needs to verify 
\begin{equation*}
  4k \log(\frac{ed^2}{2k}) + 4(d+k)\log \left(c_{\frac{b}{a}} \frac{d}{\varepsilon}\right) + \log 2 - \frac{m \delta^2}{32 e^2 \ssexpo^2} \leq \log(\rho/2) \,,
\end{equation*}
which is equivalent to
\begin{equation*}
  \frac{m \delta^2}{32 e^2 \ssexpo^2}\geq 4k \log(\frac{ed^2}{2k}) + 4(d+k)\log \left(c_{\frac{b}{a}} \frac{d}{\varepsilon}\right) + \log 2 + \log(2/\rho) \,.
\end{equation*}
To simplify the expression, we derive a sufficient condition on $m$ given by
\begin{equation}
  m \geq \frac{32 e^2 \ssexpo^2}{\delta^2} \left[4k \log(\frac{ed^2}{2k}) + 4(d+k)  \log \left( \frac{32 e^2 c_{\frac{b}{a}} \ssexpo}{\delta} d \left[  d^2 \log(6) + 4\log(2/\rho) \right] \right) + 2\log(2/\rho) \right]\,.
\label{eq:m_for_RIP_ROP_quanti}
\end{equation}
This finishes the proof as we can find a constant $C=C(\delta, \rho, b/a)$ such that $m \geq C (d+2k) \log d$ implies \eqref{eq:m_for_RIP_ROP_quanti}. 
\end{proof}

\subsection{Connection with Bregman proximal gradient \label{sec:connections}}

The iterations of our algorithm \eqref{eq:_algorithm_iter} can be related to the iterations of Bregman Proximal Gradient (BPG). Originally introduced in \cite{bauschke2017descent}, BPG is a generalization of the classical proximal gradient method in which the proximal operator is replaced with a Bregman proximal operator. It aims at solving problems of the form $\min f+g$ where $f,g$ are proper, convex and lower semi-continuous. For $\lambda >0$, we consider the optimization problem
\begin{equation}
\label{probleme_optim_approx_2}
\underset{\Thetab \succ 0}{\min} \ F(\Thetab)+\lambda \|\Thetab\|_{1,\operatorname{off}} \text{ where } F(\Thetab) \stackrel{\D}{=} \frac{1}{2} \| \Acal(\Thetab^{-1})-\sbf\|_2^2 = f(\Thetab^{-1})\,.
\end{equation}
Interestingly the function $F$ is convex on the convex set $S := \{\Thetab \succ 0 : \lambda_{\max}(\Thetab) \leq 1/\lambda_{\max}(\widehat{\Sigmab})\} \subset S_{d}^{++}$. Indeed as shown in \cite[Example 3.4]{cboyd} the function $(\a, \Thetab) \rightarrow \a^\top \Thetab^{-1} \a$ is jointly convex on $\R^{d} \times S_d^{++}(\R)$. Consequently $\Thetab \to (\a_j^\top \Thetab^{-1} \a_j - s_j)^{2} = (\a_j^\top (\Thetab^{-1}-\widehat \Sigmab) \a_j)^2$ is convex on $S$ as the composition of a convex function and $t \to t^2$ on $\R_{+}$ since $\a_j^\top (\Thetab^{-1}-\widehat \Sigmab) \a_j \geq 0$ on $S$. Hence $F(\Thetab) = \frac{1}{2} \sum_{j=1}^{m} (\a_j^\top \Thetab^{-1} \a_j - s_j)^{2}$ is also convex on $S$ as the sum of convex functions. 
For a fixed step-size $\gamma > 0$ the BPG iterations with the Bregman divergence \eqref{eq:log_det_breg} are given by 
\begin{equation}
\Thetab_{t+1} = \underset{\Thetab \succ 0}{\operatorname{argmin}} \ \langle \gamma\nabla F(\Thetab_t), \Thetab \rangle + D_{h}(\Thetab|\Thetab_t) + \lambda \gamma \|\Thetab\|_{1,\operatorname{off}} \,.
\end{equation}
By expressing the divergence $D_h$ and using that $\nabla F(\Thetab_t) = -\Thetab_{t}^{-1} \nabla f(\Thetab_{t}^{-1}) \Thetab_{t}^{-1}$ these iterations are equivalent to 
\begin{equation}
\label{eq:almost_glasso}
\Thetab_{t+1} = \underset{\Thetab \succ 0}{\operatorname{argmin}} \ \langle \Thetab_{t}^{-1}-\gamma\Thetab_{t}^{-1} \nabla f(\Thetab_{t}^{-1}) \Thetab_{t}^{-1}, \Thetab \rangle -\log\det\Thetab + \lambda \gamma \|\Thetab\|_{1, \operatorname{off}} \,.
\end{equation}
These iterations also correspond to a graphical lasso since \eqref{eq:almost_glasso} rewrites as $\Thetab_{t+1}^{-1} =\operatorname{GLASSO}_{\lambda\gamma}[\Thetab_{t}^{-1}-\gamma\Thetab_{t}^{-1} \nabla f(\Thetab_{t}^{-1}) \Thetab_{t}^{-1}]$. With the change of variable $\Sigmab_t = \Thetab_{t}^{-1}$, the BPG iterations for solving \eqref{probleme_optim_approx_2} equivalently write
\begin{equation}
\label{eq:bgp_iter}
\Sigmab_{t+1} =\operatorname{GLASSO}_{\lambda\gamma}[\Sigmab_{t}-\gamma \Sigmab_t \nabla f(\Sigmab_{t}) \Sigmab_t] \,.
\end{equation}
We can notice that these iterations are very similar to the one in \eqref{eq:_algorithm_iter} but with $\Sigmab_t \nabla f(\Sigmab_{t}) \Sigmab_t$ instead of $\nabla f(\Sigmab_{t})$. In fact, the iterations \eqref{eq:_algorithm_iter} are equivalent to the BPG iterations \eqref{eq:bgp_iter} when using a Riemannian gradient instead of a Euclidean one. More precisely, when considering for $\X \succ 0$, the inner product $\langle \U, \V\rangle_{\X} \stackrel{\D}{=}\tr(\U \X^{-1} \V \X^{-1})$ and computing the gradient \text{w.r.t.} $\langle \cdot, \cdot \rangle_\X$ at $\X$ we get the formula $\operatorname{grad} f(\X) = \X \nabla f(\X) \X$ \cite{han2021riemannian}. This corresponds to endowing the space $S_{d}^{++}(\R)$ with the affine-invariant geometry \cite[Section 11.7]{boumal2023introduction}. In conclusion, if we consider our iterations \eqref{eq:_algorithm_iter} with the Riemannian gradient $\operatorname{grad} f$ instead of the Euclidean gradient we get the BPG iterations \eqref{eq:bgp_iter}. However, we observe in practice that the algorithm with the BPG has degraded performance compared to the one proposed in \eqref{eq:_algorithm_iter}.

\subsection{Safe step-size strategy \label{sec:safe_step}}

The goal of this section is to provide a step-size $\gamma >0$ ensuring that the matrix $\Sigmab_{t+\frac{1}{2}} := \Sigmab_{k}-\gamma \nabla f(\Sigmab_k)$ remains positive definite during the iterations. Recall that  $\nabla f(\Sigmab_t) = \Acal^\star(\Acal(\Sigmab_t)-\sbf)$ where $\sbf = \frac{1}{n} \sum_{i=1}^{n} \Phi(\xbf_i) = \frac{1}{m} \left(\a_1^\top \widehat{\Sigmab} \a_1, \cdots, \a_m^\top \widehat{\Sigmab} \a_m\right)$ is the sketch of the data and $\widehat{\Sigmab}$ is the empirical covariance matrix. The adjoint operator $\Acal^\star$ is given by $\ybf \rightarrow \Acal^\star(\ybf) = \frac{1}{m}\sum_{j=1}^{m} y_j \a_j \a_j^\top$.

The matrix $\Sigmab_{t+\frac{1}{2}}$ is positive definite when $\lambda_{\min}(\Sigmab_{t}-\gamma \nabla f(\Sigmab_t)) > 0$ that is when 
\begin{equation}
\label{cond_on_gamma}
\lambda_{\min}\left(\Sigmab_{t}-\gamma \frac{1}{m}\sum_{j=1}^{m} (\frac{1}{m} \a_j^\top \Sigmab_t \a_j-s_j) \a_j \a_j^\top\right) > 0\,.
\end{equation}

Moreover we have, 
\begin{equation}
\label{eq:first_bound}
\begin{split}
\lambda_{\min}\left(\Sigmab_{t}-\gamma \frac{1}{m}\sum_{j=1}^{m} (\frac{1}{m}\a_j^\top \Sigmab_t \a_j-s_j) \a_j \a_j^\top\right) &\geq \lambda_{\min}(\Sigmab_t) - \gamma \frac{1}{m}\lambda_{\max}\left(\sum_{j=1}^{m} (\frac{1}{m}\a_j^\top \Sigmab_t \a_j-s_j) \a_j \a_j^\top\right) \\
&=\lambda_{\min}(\Sigmab_t) - \gamma \frac{1}{m^2} \lambda_{\max}\left(\sum_{j=1}^{m} [\a_j^\top( \Sigmab_t-\widehat{\Sigmab})\a_j] \a_j \a_j^\top\right)\,.
\end{split}
\end{equation}

Using $\forall j \in \integ{m}, \a_j(\Sigmab_t-\widehat{\Sigmab}) \a_j \leq \lambda_{\max}(\Sigmab_t -\widehat{\Sigmab}) \|\a_j\|_2^2$ we have, for any $\zbf \in \R^{d}, \|\zbf\|_2 = 1$, 
\begin{equation}
\zbf^\top(\sum_{j=1}^{m} [\a_j^\top( \Sigmab_t-\widehat{\Sigmab})\a_j] \a_j \a_j^\top)\zbf = \sum_{j=1}^{m} [\a_j^\top( \Sigmab_t-\widehat{\Sigmab})\a_j] |\zbf^\top \a_j|^2 \leq \lambda_{\max}(\Sigmab_t -\widehat{\Sigmab}) \sum_{j=1}^{m} \|\a_j\|_2^2 |\zbf^\top \a_j|^2\,.
\end{equation}

By introducing the matrix $\Abf = (\a_1, \cdots, \a_m) \in \R^{d \times m}$ the previous inequality leads to
\begin{equation}
\begin{split}
\zbf^\top(\sum_{j=1}^{m} [\a_j^\top( \Sigmab_t-\widehat{\Sigmab})\a_j] \a_j \a_j^\top)\zbf &\leq \lambda_{\max}(\Sigmab_k -\widehat{\Sigmab}) \max_{j \in \integ{m}} \|\a_j\|_2^2 \sum_{j=1}^{m} |\zbf^\top\a_j|^2\\
& = \lambda_{\max}(\Sigmab_t -\widehat{\Sigmab}) (\max_{j \in \integ{m}} \|\a_j\|_2^2) \ \zbf^\top \Abf \Abf^\top \zbf\\
&\leq \lambda_{\max}(\Sigmab_t -\widehat{\Sigmab}) (\max_{j \in \integ{m}} \|\a_j\|_2^2) \lambda_{\max}(\Abf \Abf^\top) \\
&\leq [\lambda_{\max}(\Sigmab_t)-\lambda_{\min}(\widehat{\Sigmab})](\max_{j \in \integ{m}} \|\a_j\|_2^2) \lambda_{\max}(\Abf \Abf^\top)\,.
\end{split}
\end{equation}

Consequently, 
\begin{equation}
\label{eq:bound_max_eigen}
\lambda_{\max}\left(\sum_{j=1}^{m} [\a_j^\top( \Sigmab_t-\widehat{\Sigmab})\a_j] \a_j \a_j^\top\right) \leq [\lambda_{\max}(\Sigmab_t)-\lambda_{\min}(\widehat{\Sigmab})](\max_{j \in \integ{m}} \|\a_j\|_2^2) \lambda_{\max}(\Abf \Abf^\top)\,.
\end{equation}

This shows that if $\lambda_{\max}(\Sigmab_t) < \lambda_{\min}(\widehat{\Sigmab})$ then the condition \eqref{cond_on_gamma} is valid for any $\gamma > 0$ since in this case $\lambda_{\max}\left(\sum_{j=1}^{m} [\a_j^\top( \Sigmab_t-\widehat{\Sigmab})\a_j] \a_j \a_j^\top\right) < 0$. On the other hand if $\lambda_{\max}(\Sigmab_k) > \lambda_{\min}(\widehat{\Sigmab})$ then, by \eqref{eq:first_bound} and \eqref{eq:bound_max_eigen}, a step-size 
\begin{equation}
\gamma < \frac{m^2}{\max_{j \in \integ{m}} \|\a_j\|_2^2 \ \sigma^2_{\max}(\Abf)} \times \frac{\lambda_{\min}(\Sigmab_t)}{\lambda_{\max}(\Sigmab_t)-\lambda_{\min}(\widehat{\Sigmab})}\,,
\end{equation}
where $\sigma_{\max}(\Abf)$ is the largest singular value of $\Abf$, ensures that $\Sigmab_{k+\frac{1}{2}}$ is positive definite. Overall a safe step-size strategy is given by
\begin{equation}
\gamma \in \begin{cases} (0,+\infty[ & \text{ if } \lambda_{\max}(\Sigmab_t) < \lambda_{\min}(\widehat{\Sigmab}) \,,  \\ 
(0, \frac{m^2}{\max_{j \in \integ{m}} \|\a_j\|_2^2 \ \sigma^2_{\max}(\Abf)}  \frac{\lambda_{\min}(\Sigmab_t)}{\lambda_{\max}(\Sigmab_t)-\lambda_{\min}(\widehat{\Sigmab})}) & \text{ if } \lambda_{\max}(\Sigmab_t) > \lambda_{\min}(\widehat{\Sigmab}) \,.
\end{cases}
\end{equation}

We emphasize that this strategy is quite conservative, and requires the computation of both the maximum and minimum eigenvalues of $\Sigmab_t$ at every iteration. In practical scenarios, we find that searching for $\gamma$ in $\{1e-3, 1e-2, 1e-1\}$ is adequate for achieving convergence in our experimental setups.

\subsection{Covering number of a model set with a condition number constraint. \label{sec:covering_cond_nb}}

We would like to consider a model set of covariance matrices that does not restrict their spectra but rather their condition numbers. For a square matrix $\Mbf$, the condition number is defined as $\kappa(\Mbf) \stackrel{\D}{=} \|\Mbf\|_{2\to2} \|\Mbf^{-1}\|_{2\to2}$. In the special case of positive-definite matrices, it is the ratio between the largest and smallest eigenvalues. For some $\kappa_0 \geq 1$, we consider the following model set 
\begin{equation*}
\Sfrak_{k, \kappa_0} \stackrel{\D}{=} \left\{\Sigmab \in S_{d}^{++}(\R): \Thetab=\Sigmab^{-1} \succ 0, \|\Thetab\|_{0} \leq d+2k, \kappa(\Thetab) \leq \kappa_0 \right\}\,.
\end{equation*}
Remark that the above definition implies that $\Sfrak_{k, a, b} \subset \Sfrak_{k, \kappa_0}$, for all $a, b > 0$ such that $b/a \leq \kappa_0$. In particular, $\Sfrak_{k, 1/\kappa_0, 1} \subset \Sfrak_{k, \kappa_0}$.  However, $\Sfrak_{k, \kappa_0}$ is a much ``bigger'' set than sets like $\Sfrak_{k, a, b}$, as the latter are bounded sets while the former is not. Interestingly enough, we are able to upper-bound the covering number of the normalized secant of $\Sfrak_{k, \kappa_0}$ that is
\begin{equation*}
S[\Sfrak_{k, \kappa_0}] = \left\{ \frac{\Thetab_1^{-1} - \Thetab_2^{-1}}{\|\Thetab_1^{-1} - \Thetab_2^{-1}\|_{\Lambda}}: \ (\Thetab_1^{-1}, \Thetab_2^{-1}) \in \Sfrak_{k, \kappa_0}^2,\ \|\Thetab_1^{-1} - \Thetab_2^{-1}\|_{\Lambda}>0 \right\}.
\end{equation*}
The key element to obtain this bound is that we are able to rewrite $S[\Sfrak_{k, \kappa_0}]$ in term of matrices that are in 
\begin{equation*}
\Sfrak_0 \stackrel{\D}{=} \left\{\Sigmab \in S_{d}^{++}(\R): \Thetab=\Sigmab^{-1} \succ 0, \|\Thetab\|_{0} \leq d+2k,\ \spec(\Thetab) \subset [1/\kappa_0, 1],\ \|\Thetab\|_{2 \to 2} = 1 \right\}. 
\end{equation*}
Indeed, for an element $\frac{\tilde{\Thetab}_1^{-1} - \tilde{\Thetab}^{-1}}{\|\tilde{\Thetab}^{-1} - \tilde{\Thetab}^{-1}\|_{\Lambda}} \in S[\Sfrak_{k, \kappa_0}]$, notice that for any $\lambda > 0$ the matrix $\frac{\lambda \tilde{\Thetab}_1^{-1} - \lambda \tilde{\Thetab}^{-1}}{\|\lambda \tilde{\Thetab}^{-1} - \lambda \tilde{\Thetab}^{-1}\|_{\Lambda}}$ still is on $S[\Sfrak_{k, \kappa_0}]$. This implies that we can normalize the matrices involved in the secant. More precisely, by choosing $\lambda = \|\tilde{\Thetab}_1\|_{2\to2}$, setting $\Thetab_1 = \lambda^{-1} \tilde{\Thetab}_1$,  $\Thetab_2 = \tilde{\Thetab}_2 / \|\tilde{\Thetab}_2\|_{2 \to 2}$ and $\tau = \|\tilde{\Thetab}_1\|_{2 \to 2} / \|\tilde{\Thetab}_2\|_{2 \to 2}$, we have $\Thetab_1, \Thetab_2 \in \Sfrak_0$, $\tau >0$ and $\frac{\Thetab_1^{-1} - \tau \Thetab_2^{-1}}{\|\Thetab_1^{-1} - \tau \Thetab_2^{-1}\|_{\Lambda}} \in S[\Sfrak_{k, \kappa_0}]$. This shows that
\begin{align*}
S[\Sfrak_{k, \kappa_0}] &= \left\{ \frac{\Thetab_1^{-1} - \tau \Thetab_2^{-1}}{\|\Thetab_1^{-1} - \tau \Thetab_2^{-1}\|_{\Lambda}}:\ \tau > 0,\ (\Thetab_1^{-1}, \Thetab_2^{-1}) \in \Sfrak_0^2,\ \|\Thetab_1^{-1} - \tau \Thetab_2^{-1}\|_{\Lambda}>0 \right\}.  
\end{align*}
Now, up to exchanging the role of $\tilde{\Thetab}_1$ and $\tilde{\Thetab}_2$, which result in changing the sign of $\frac{\Thetab_1^{-1} - \tau \Thetab_2^{-1}}{\|\Thetab_1^{-1} - \tau \Thetab_2^{-1}\|_{\Lambda}}$, we can always assume that $\|\tilde{\Thetab}_1\|_{2 \to 2} \leq  \|\tilde{\Thetab}_2\|_{2 \to 2}$, meaning that $0 < \tau \leq 1$, which yields
\begin{align*}
S[\Sfrak_{k, \kappa_0}] &=  \left\{ \pm \frac{\Thetab_1^{-1} - \tau \Thetab_2^{-1}}{\|\Thetab_1^{-1} - \tau \Thetab_2^{-1}\|_{\Lambda}}:\ \tau \in (0,1],\ (\Thetab_1^{-1}, \Thetab_2^{-1}) \in \Sfrak_0^2:\ \|\Thetab_1^{-1} - \tau \Thetab_2^{-1}\|_{\Lambda}>0 \right\} \\
&= \overline{S}[\Sfrak_{k, \kappa_0}] \ \cup \ (-\overline{S}[\Sfrak_{k, \kappa_0}]) \,,
\end{align*}
where 
\begin{equation*}
\overline{S}[\Sfrak_{k, \kappa_0}] \stackrel{\D}{=} \left\{ \frac{\Thetab_1^{-1} - \tau \Thetab_2^{-1}}{\|\Thetab_1^{-1} - \tau \Thetab_2^{-1}\|_{\Lambda}}:\ \tau \in (0,1],\ (\Thetab_1^{-1}, \Thetab_2^{-1}) \in \Sfrak_0^2,\ \|\Thetab_1^{-1} - \tau \Thetab_2^{-1}\|_{\Lambda}>0 \right\}. 
\end{equation*}

Remark that for any $\varepsilon>0$, $\Ncal(S[\Sfrak_{k, \kappa_0}], \|\cdot\|_\Lambda, \varepsilon) \leq 2 \Ncal(\overline{S}[\Sfrak_{k, \kappa_0}], \|\cdot\|_\Lambda, \varepsilon)$, therefore we only have to control the covering number of $\overline{S}[\Sfrak_{k, \kappa_0}]$.
To do so, we follow the same line of proof as in the control of $S[\Sfrak_{k,a,b}]$ by splitting our set of interests into long and short chords. The analysis of these chords is similar, although more technical and more computation-heavy. A slight complication in this new setting is the need to ensure that $\tau$ is bounded away from zero in the case of short chords. In the following, we briefly detail the analysis of the long and short chords given for some $\eta>0$ by 
\begin{equation*}
\begin{split}
\overline{S}^+_\eta[\Sfrak_{k, \kappa_0}] &\stackrel{\D}{=} \left\{ \frac{\Thetab_1^{-1} - \tau \Thetab_2^{-1}}{\|\Thetab_1^{-1} - \tau \Thetab_2^{-1}\|_{\Lambda}}:\ \tau \in (0,1],\ (\Thetab_1^{-1}, \Thetab_2^{-1}) \in \Sfrak_0^2,\ \|\Thetab_1^{-1} - \tau \Thetab_2^{-1}\|_{\Lambda}>\eta \right\}, \\
\overline{S}^-_\eta[\Sfrak_{k, \kappa_0}] &\stackrel{\D}{=} \left\{ \frac{\Thetab_1^{-1} - \tau \Thetab_2^{-1}}{\|\Thetab_1^{-1} - \tau \Thetab_2^{-1}\|_{\Lambda}}:\ \tau \in (0,1],\ (\Thetab_1^{-1}, \Thetab_2^{-1}) \in \Sfrak_0^2,\ 0<\|\Thetab_1^{-1} - \tau \Thetab_2^{-1}\|_{\Lambda}\leq\eta \right\}. 
\end{split}
\end{equation*}

\subsubsection{Control of the long chords}

The strategy to control the covering number of long chords is to express $\overline{S}^+_\eta[\Sfrak_{k, \kappa_0}]$ as the image of a Lipschitz-continuous function and control the covering number of the original set. Let us consider the set $\overline{\Xfrak}_\eta \stackrel{\D}{=} \{ (\tau, \Thetab_1, \Thetab_2) \in (0,1] \times \Sfrak_0^{-1} \times \Sfrak_0^{-1}, \ \|\Thetab_1^{-1} - \tau \Thetab_2^{-1}\|_{\Lambda}>\eta  \}$ equipped with the norm $\|(\tau, \Mbf_1, \Mbf_2)\|_{\otimes} = |\tau| + \|\Mbf_1\|_\Fro + \|\Mbf_2\|_\Fro$. Then we have the following lemma. For the sake of conciseness, its proof is not provided, but it is based on the one of Proposition~\ref{prop:true_cov_number_long_chors} presented in Appendix \ref{proof:prop:true_cov_number_long_chors}.
\begin{lemma}
\label{lem:g_lipschitz_cond_nb}
Let $g : (\overline{\Xfrak}_\eta, \|\cdot\|_\otimes) \to (\overline{S}^+_\eta[\Sfrak_{k, \kappa_0}], \|\cdot\|_\Lambda)$ be the function defined by  
\begin{equation*}
g(\tau, \Thetab_1, \Thetab_2) \stackrel{\D}{=} \frac{\Thetab_1^{-1} - \tau \Thetab_2^{-1}}{\|\Thetab_1^{-1} - \tau \Thetab_2^{-1}\|_{\Lambda}} \,. 
\end{equation*}
Then, g is surjective and $L_0/\eta$-lipschitz continuous with $L_0 = 2 C_\Fro \kappa_0^2 \sqrt{d}$.
\end{lemma}

As a consequence, we can control the covering number of  $\overline{S}^+_\eta[\Sfrak_{k, \kappa_0}]$ using the one of $\overline{\Xfrak}_\eta$ which is easier to handle. This yields the following proposition which proof is also based on the one of Proposition~\ref{prop:true_cov_number_long_chors}.
\begin{proposition}
\label{prop:cov_long_chords_cond_nb}
For all $\varepsilon>0$ and $\eta>0$, we have 
\begin{equation*}
\Ncal( \overline{S}^+_\eta[\Sfrak_{k, \kappa_0}]  , \|\cdot\|_\Lambda, \varepsilon ) \leq \Ncal( (0,1], |\cdot|, \frac{\eta \varepsilon}{6L_0} ) \times \Ncal( \Sfrak_0^{-1}, \|\cdot\|_\Fro , \frac{\eta \varepsilon}{6L_0} )^2 \,. 
\end{equation*}
\end{proposition} 
Note that $\Ncal( (0,1], |\cdot|, \varepsilon )$ is bounded by $\varepsilon^{-1}$ and the control of the covering of $\Sfrak_0^{-1}$ will be provided later by Lemma~\ref{lemma:covering_secant_cond_nb}.

\subsubsection{Control of the short chords}

In order to control the covering number of the short chords we follow these two steps: 1) we show that any element of $\overline{S}^-_\eta[\Sfrak_{k, \kappa_0}]$ is close to an element of a certain ‘‘tangent space''  2) we will control the covering number of this space. 

\textbf{Assumption:} In all of this section we will assume that $0 < \eta \leq c_\Fro/2$. This requirement will be useful for various simplification and will be met when we calibrate $\eta$ for a good balance between the covering numbers of both long and short chords. 

Point 1) is done through the following lemma which proof can be found in Appendix~\ref{proof:short_chords_cond_nb}.
\begin{lemma}
\label{lem:exists_tanget_cond_nb}
For $\eta >0$, consider the set of short chords $\overline{S}^-_\eta[\Sfrak_{k, \kappa_0}]$ and the normalized secant set $S[\Sfrak_{k, \kappa_0}^{-1}]$. Define $C \stackrel{\D}{=} \{\lambda \Vbf: \ \lambda \in ]0,\lambda_0], \Vbf \in S[\Sfrak_{k, \kappa_0}^{-1}] \}$ and
\begin{equation*}
T_C \stackrel{\D}{=} \left\{\dr \inv_{\tilde{\Thetab}}(\Cbf): (\tilde{\Thetab},\Cbf) \in \Sfrak_{k, \frac{1}{\kappa_0}, 2}^{-1} \times C \right\},
\end{equation*}
with $\lambda_0 \stackrel{\D}{=} \frac{2}{c_\Fro}$. Defining $Z_0 \stackrel{\D}{=} \frac{C_\Fro \kappa_0^3}{c_\Fro^2}$, we have
\begin{equation*}
\forall \Ubf \in \overline{S}^-_\eta[\Sfrak_{k, \kappa_0}], \exists \Tbf \in \Tcal, \ \|\Ubf-\Tbf\|_\Lambda \leq Z_0 \eta \,. 
\end{equation*}
\end{lemma}

As $T_C$ is a good approximation of $\overline{S}^-_\eta[\Sfrak_{k, \kappa_0}]$, we can bound the covering number of the latter by the covering of the former (with a differnt scale), see Lemma~\ref{lemma:generalized_cover}. Hence, we need to control the covering number of $T_C$. 
\begin{lemma}
\label{lemma:subset_cond_nb}
For all $\varepsilon>0$, we have
\begin{equation}
\label{eq:covering_of_Tcal}
\Ncal(T_C, \|\cdot\|_\Lambda, \varepsilon)  \leq \Ncal(\Sfrak_{k,\frac{1}{\kappa_0}, 2}^{-1}, \|\cdot\|_\Fro, \frac{\varepsilon}{C_{0}}) \times \Ncal(C, \|\cdot\|_\Fro, \frac{\varepsilon}{C_{0}}) \,,
\end{equation}
with $ C_{0} = C_\Fro(2\kappa_0^3 \lambda_0 + \kappa_0^2)$.
\end{lemma}

Now, combining the above results, we are able to provide a control of the covering number of the short chords. See Appendix~\ref{proof:short_chords_cond_nb} for the proof.
\begin{proposition}[Similar to Proposition~\ref{prop:true_cov_number_short_chors}]
\label{prop:true_cov_number_short_chors_cond_nb}
For any $\varepsilon>0$ and $\eta >0$, we have 
\begin{equation*}
\Ncal(\overline{S}^-_\eta[\Sfrak_{k, \kappa_0}], \|\cdot\|_\Lambda, 2(\varepsilon+ Z_0 \eta)) \leq 2 \lambda_0 C_{0} \varepsilon^{-1} \Ncal(\Sfrak_{k,\frac{1}{\kappa_0}, 2}^{-1}, \|\cdot\|_\Fro, \frac{\varepsilon}{C_{0}}) \times \Ncal(S[\Sfrak_{k, \kappa_0}^{-1}], \|\cdot\|_\Fro, \frac{\varepsilon}{2 \lambda_0 C_{0}}) \,,
\end{equation*}
where $Z_0=\frac{C_\Fro \kappa_0^3}{c_\Fro^2}$, $\lambda_0= \frac{2}{c_\Fro}$, $ C_{0} = C_\Fro(2\kappa_0^3 \lambda_0 + \kappa_0^2)$.  
\end{proposition}

\subsubsection{Combining the results}

To finish this section and obtain the covering of $\overline{S}_\eta[\Sfrak_{k, \kappa_0}]$, we need the control of the covering of $\Sfrak_0^{-1}$, $\Sfrak_{k,\frac{1}{\kappa_0}, 2}^{-1}$ and $S[\Sfrak_{k, \kappa_0}^{-1}]$ as they appear in the control of the covering number of the long and short chords. This is done in the following lemma which proof can be found in Appendix~\ref{proof:easy_covering_nb}. 

\begin{lemma}
\label{lemma:covering_secant_cond_nb}
For any $\varepsilon>0$, and $0<a\leq b$, we have 
\begin{equation*}
\begin{split}
\Ncal(\Sfrak_0^{-1}, \|\cdot\|_\Fro, \varepsilon ) &\leq  \left( \frac{e d^2}{2k} \right)^{k} \left( \frac{18 \sqrt{d}}{\varepsilon}\right)^{d+k}, \\
\Ncal(\Sfrak_{k,\frac{1}{\kappa_0}, 2}^{-1}, \|\cdot\|_\Fro, \varepsilon ) &\leq  \left( \frac{e d^2}{2k} \right)^{k} \left( \frac{2\times 18 \sqrt{d}}{\varepsilon}\right)^{d+k}, \\
\Ncal(S[\Sfrak_{k, \kappa_0}^{-1}], \|\cdot\|_\Fro, \varepsilon ) &\leq  \left( \frac{e d^2}{4k} \right)^{2k} \left( \frac{18}{\varepsilon} \right)^{d+2k}.
\end{split}
\end{equation*}
\end{lemma}

Gathering up all the pieces, we obtain the following theorem.

\begin{restatable}{theorem}{big_theo_covering_cond_nb}
\label{theo:big_theo_covering_cond_nb}
There exist absolute constants $\tilde{c_1}$ and $\tilde{c_2}$ such that for any $\varepsilon$ such that $0 < \varepsilon \leq \frac{2 \kappa_0^3}{c_\Fro} \sqrt{d}$, we have
\begin{equation*}
\Ncal(S_\eta[\Sfrak_{k, \kappa_0}], \|\cdot\|_\Lambda, \varepsilon) \leq 2 \left( \frac{e d^2}{2k} \right)^{3k} \left[ \left( \frac{\tilde{c_1} \kappa_0^5 d^2}{\varepsilon^2} \right)^{2(d+k)+1} +   \left( \frac{\tilde{c_2} \kappa_0^3 d^2 }{\varepsilon} \right)^{2d+3k+1}\right].
\end{equation*}
\end{restatable}

See Appendix~\ref{proof:big_theo_covering_cond_nb} for the proof.

\subsection{Proof for the coverings with a condition number hypothesis \label{proof:covering_cond_nb}}

Before diving into the control of the covering numbers of the long and short chords, let us claim various inequalities related to the inverse function on matrices that will be useful in the following. 

\begin{lemma}[Inverse function properties]
Assume that there exist constants $c_\Fro$ and $C_\Fro$ such that $c_\Fro \|\Mbf\|_\Fro \leq \|\Mbf\|_\Lambda \leq C_\Fro \|\Mbf\|_\Fro$, for all $\Mbf \in S_d$. Let $\Mbf_1$ and $\Mbf_2$ be two matrices in $S_d^{++}$. Then we have the following inequalities:
\begin{align}
\| \Mbf_1^{-1} - \Mbf_2^{-1} \|_{\Lambda} &\leq C_\Fro \ \|\Mbf_1^{-1}\|_{2\to2} \ \|\Mbf_2^{-1}\|_{2\to2} \ \| \Mbf_1 - \Mbf_2 \|_{\Fro} \,, \label{eq:lipsch_cond_nb}\\
\| \Mbf_1 - \Mbf_2 \|_{\Fro} &\leq \frac{1}{c_\Fro} \ \|\Mbf_1\|_{2\to2} \ \|\Mbf_2\|_{2\to2} \ \|\Mbf_1^{-1} - \Mbf_2^{-1}\|_\Lambda\,, \label{eq:inv_lipsch_cond_nb}\\
\|\Mbf_1^{-1}-\Mbf_2^{-1}-\dr \inv_{\Mbf_2}(\Mbf_1-\Mbf_2)\|_\Lambda &\leq  C_\Fro \ \|\Mbf_1^{-1}\|_{2 \to 2} \ \|\Mbf_2^{-1}\|_{2 \to 2}^2 \ \|\Mbf_1-\Mbf_2\|_\Fro^{2}\,, \label{eq:approx_by_diff_cond_nb}\\
\|\dr \inv_{\Mbf_1}(\Mbf)-\dr \inv_{\Mbf_2}(\Mbf)\|_\op &\leq C_\Fro \left( \|\Mbf^{-1}_1\|_{2 \to 2} + \|\Mbf^{-1}_2\|_{2 \to 2} \right) \label{eq:lipsch_diff_cond_nb} \\
& \qquad \times \|\Mbf^{-1}_1\|_{2 \to 2} \|\Mbf^{-1}_2\|_{2 \to 2} \|\Mbf_1-\Mbf_2\|_\Fro\,. \nonumber
\end{align}
\end{lemma}

\begin{proof}[Idea of the proof]
It follows the ideas of the proof of Lemma~\ref{lemma:finvsatisfies}. 
\end{proof}

\subsubsection{Control of the short chords \label{proof:short_chords_cond_nb}}

 We begin with the following preliminary result that ensures that $\tau$ can not be too close to $0$. 

\begin{lemma}
\label{lemma:alpha_not_so_small}
Assume that $(\tau, \Thetab_1,\Thetab_2) \in (0,1] \times \Sfrak_0^{-1} \times \Sfrak_0^{-1}$ verifies $0 < \|\Thetab_1^{-1} - \tau \Thetab_2^{-1}\|_{\Lambda} \leq \eta$ and $\eta /leq c_\Fro/2$. Then $\tau$ is bounded away from $0$ \ie\
\begin{equation*}
\tau \geq  1 - \frac{\eta}{c_\Fro} \geq \frac{1}{2}\,.
\end{equation*}
\end{lemma}
\begin{proof}
From the inverse-lipschitz property of the inverse function in \eqref{eq:inv_lipsch_cond_nb}, we have :
\begin{equation*}
\| \Thetab_1 - \tau^{-1} \Thetab_2 \|_\Fro \leq \frac{1}{c_\Fro} \|\Thetab_1\|_{2\to2} \|\tau^{-1} \Thetab_2\|_{2\to2} \|\Thetab_1^{-1} - \tau \Thetab_2^{-1}\|_{\Lambda} \leq \frac{\eta}{c_\Fro \tau}.
\end{equation*}
Moreover, 
\begin{equation*}
\begin{split}
\| \Thetab_1 - \tau^{-1} \Thetab_2 \|_\Fro &\geq \|  \Thetab_1 - \tau^{-1} \Thetab_2 \|_{2 \to 2} \geq \tau^{-1} \| \Thetab_2 \|_{2 \to 2} - \| \Thetab_1 \|_{2 \to 2} = \frac{1}{\tau}-1 \,.
\end{split}
\end{equation*}
Combining the two inequalities yields $\tau \geq 1 - \frac{\eta}{c_\Fro}$. 
\end{proof}

Let us now prove Lemma~\ref{lem:exists_tanget_cond_nb}.
\begin{proof}[Proof of Lemma~\ref{lem:exists_tanget_cond_nb}]
Take $\Ubf = \frac{\Thetab_1^{-1} - \tau \Thetab_2^{-1}}{\|\Thetab_1^{-1} - \tau \Thetab_2^{-1}\|_{\Lambda}} \in \overline{S}^-_\eta[\Sfrak_{k, \kappa_0}]$ so we have $0< \|\Thetab_1^{-1} - \tau \Thetab_2^{-1}\|_{\Lambda} \leq \eta$. 
Note that by using \eqref{eq:lipsch_cond_nb} and \eqref{eq:inv_lipsch_cond_nb}, we have 
\begin{equation}
\label{eq:preimage_bound}
0 \stackrel{\eqref{eq:lipsch_cond_nb}}{<} \|\Thetab_1 - \tau^{-1} \Thetab_2\|_\Fro \stackrel{\eqref{eq:inv_lipsch_cond_nb}}{\leq} \frac{\eta}{c_\Fro \tau} \|\Thetab_1\|_{2\to2} \|\Thetab_2\|_{2\to2} \leq \frac{\eta}{c_\Fro \tau}. 
\end{equation}
Now, from \eqref{eq:approx_by_diff_cond_nb} with $\Mbf_1 = \Thetab_1$ and $\Mbf_2 = \tau^{-1} \Thetab_2$, we have
\begin{equation*}
\begin{split}
\|\Thetab_1^{-1} - \tau \Thetab_2^{-1}-\dr \inv_{\tau^{-1} \Thetab_2} (\Thetab_1-\tau^{-1} \Thetab_2)\|_\Lambda &\stackrel{\eqref{eq:approx_by_diff_cond_nb}}{\leq} C_\Fro \|\Thetab_1^{-1}\|_\Fro \|\tau \Thetab_2^{-1}\|_\Fro^2 \|\Thetab_1-\tau^{-1} \Thetab_2\|_\Fro^2  \\
&\leq C_\Fro \kappa_0^3 \tau^2 \|\Thetab_1-\tau^{-1} \Thetab_2\|_\Fro^{2} \,.
\end{split}
\end{equation*}
Dividing by $\|\Thetab_1^{-1} - \tau \Thetab_2^{-1}\|_\Lambda>0$ in the above inequality yields:
\begin{equation*}
\begin{split}
\|\frac{\Thetab_1^{-1} - \tau \Thetab_2^{-1}}{\|\Thetab_1^{-1} - \tau \Thetab_2^{-1}\|_{\Lambda}}-\dr \inv_{\tau^{-1} \Thetab_2} \frac{\Thetab_1-\tau^{-1} \Thetab_2}{\|\Thetab_1^{-1} - \tau \Thetab_2^{-1}\|_{\Lambda}} \|_\Lambda &\leq  C_\Fro \kappa_0^3 \tau^2 \|\Thetab_1-\tau^{-1} \Thetab_2\|_\Fro \frac{\|\Thetab_1-\tau^{-1} \Thetab_2\|_\Fro}{\|\Thetab_1^{-1} - \tau \Thetab_2^{-1}\|_{\Lambda}} \\
\text{(applying \eqref{eq:preimage_bound} and \eqref{eq:inv_lipsch_cond_nb})}\quad & \leq C_\Fro \kappa_0^3 \tau^2 \frac{\eta}{c_\Fro \tau} \frac{1}{c_\Fro} \|\Thetab_1\|_{2\to2} \|\tau^{-1} \Thetab_2\|_{2\to2} \\
& = \frac{C_\Fro \kappa_0^3}{c_\Fro^2} \eta = Z_0 \eta \,.
\end{split}
\end{equation*}
Moreover, as $\Thetab_2 \in \Sfrak_0^{-1} \subset \Sfrak_{k, \frac{1}{\kappa_0}, 1}^{-1}$, setting $\tilde{\Thetab} = \tau^{-1} \Thetab_2$, we have $\tilde{\Thetab} \in \Sfrak_{k, \frac{1}{\tau \kappa_0}, \frac{1}{\tau}}^{-1} \subset \Sfrak_{k, \frac{1}{\kappa_0}, 2}^{-1}$ (the inclusion comes from Lemma~\ref{lemma:alpha_not_so_small}). Remark that the element in the differential can be expressed as  
\begin{equation*}
\frac{\Thetab_1-\tau^{-1} \Thetab_2}{\|\Thetab_1^{-1} - \tau \Thetab_2^{-1}\|_{\Lambda}} = \frac{\|\Thetab_1-\tau^{-1} \Thetab_2\|_\Fro}{\|\Thetab_1^{-1} - \tau \Thetab_2^{-1}\|_{\Lambda}} \frac{\Thetab_1-\tau^{-1} \Thetab_2}{\|\Thetab_1-\tau^{-1} \Thetab_2\|_\Fro}.
\end{equation*} So, if we define $\lambda = \frac{\|\Thetab_1-\tau^{-1} \Thetab_2\|_\Fro}{\|\Thetab_1^{-1} - \tau \Thetab_2^{-1}\|_{\Lambda}}$, by \eqref{eq:inv_lipsch_cond_nb} and Lemma~\ref{lemma:alpha_not_so_small} it satisfies $0 < \lambda \leq \frac{1}{c_\Fro \tau} \leq \frac{2}{c_\Fro} \stackrel{\D}{=} \lambda_0$. Thus, there exists $\lambda \in ]0, \lambda_0 ]$ and $\tilde{\Thetab} \in \Sfrak_{k, \frac{1}{\kappa_0}, 2}^{-1}$ such that:
\begin{equation*}
\|\frac{\Thetab_1^{-1} - \tau \Thetab_2^{-1}}{\|\Thetab_1^{-1} - \tau \Thetab_2^{-1}\|_{\Lambda}}-\dr \inv_{\tilde{\Thetab}} \lambda \frac{\Thetab_1-\tau^{-1} \Thetab_2}{\|\Thetab_1-\tau^{-1} \Thetab_2\|_{\Fro}} \|_\Lambda \leq  Z_0 \eta \,.
\end{equation*}
Now we set $\Vbf = \frac{\Thetab_1-\tau^{-1} \Thetab_2}{\|\Thetab_1-\tau^{-1} \Thetab_2\|_{\Fro}}$ and we have $\Vbf \in S[\Sfrak_{k, \kappa_0}^{-1}]$, which finishes the proof.
\end{proof}

\begin{proof}[Proof of Lemma~\ref{lemma:subset_cond_nb}]
First observe that for all $\Cbf \in C$, we have $\|\Cbf\|_\Fro \leq \lambda_0$ since $\forall \Vbf \in S[\Sfrak_{k, \kappa_0}^{-1}], \|\Vbf\|_\Fro = 1$. 
Then, take $\overline{\Sfrak}_{k, \frac{1}{\kappa_0}, 2}^{-1}$ an $\varepsilon$-net of $\Sfrak_{k, \frac{1}{\kappa_0}, 2}^{-1}$ and $\overline{C}$ an $\varepsilon$-net of $C$. Take $\Tbf = \dr \inv_{\tilde{\Thetab}}(\Cbf) \in T_C$ and consider $(\overline{\Thetab}, \overline{\Cbf}) \in \overline{\Sfrak}_{k, \frac{1}{\kappa_0}, 2}^{-1} \times \overline{C}$ such that $\|\overline{\Cbf}-\Cbf\|_\Fro \leq \varepsilon$ and $\|\overline{\Thetab}-\tilde{\Thetab}\|_\Fro \leq \varepsilon$. Then with $\overline{\Tbf}= \dr \inv_{\overline{\Thetab}}(\overline{\Cbf}) \in T_C$, 
\begin{equation*}
\begin{split}
\|\Tbf-\overline{\Tbf}\|_\Lambda &= \|\dr \inv_{\tilde{\Thetab}}(\Cbf)-\dr \inv_{\overline{\Thetab}}(\overline{\Cbf})\|_\Lambda \\
&\leq \|\dr \inv_{\tilde{\Thetab}}(\Cbf)-\dr \inv_{\overline{\Thetab}}(\Cbf)\|_\Lambda+\|\dr \inv_{\overline{\Thetab}}(\Cbf)-\dr \inv_{\overline{\Thetab}}(\overline{\Cbf})\|_\Lambda \\
&\leq \|\dr \inv_{\tilde{\Thetab}}-\dr \inv_{\overline{\Thetab}}\|_{\op}\|\Cbf\|_\Fro + \|\dr \inv_{\overline{\Thetab}}\|_{\op} \|\Cbf-\overline{\Cbf}\|_\Fro \,.
\end{split}
\end{equation*}
According to \eqref{eq:lipsch_diff_cond_nb}, 
\begin{equation*}
\begin{split}
& \| \dr \inv_{\tilde{\Thetab}}-\dr \inv_{\overline{\Thetab}}\|_{\op} \\
\leq & C_\Fro \left( \|\tilde{\Thetab}^{-1}\|_{2 \to 2} + \|\overline{\Thetab}^{-1}\|_{2 \to 2} \right)\|\tilde{\Thetab}^{-1}\|_{2 \to 2} \|\overline{\Thetab}^{-1}\|_{2 \to 2} \|\tilde{\Thetab}-\overline{\Thetab}\|_\Fro\\
\leq & 2 C_\Fro \kappa_0^3 \|\tilde{\Thetab}-\overline{\Thetab}\|_\Fro.
\end{split}
\end{equation*} 
We can also show that $\|\dr \inv_{\overline{\Thetab}}\|_{\op} \leq C_\Fro \kappa_0^2$. Therefore, 
\begin{equation*}
\begin{split}
\|\Tbf-\overline{\Tbf}\|_\Lambda &\leq 2C_\Fro \kappa_0^3 \|\tilde{\Thetab}-\overline{\Thetab}\|_\Fro \delta + \|\dr \inv_{\overline{\Thetab}}\|_{\op} \varepsilon \\
&\leq 2C_\Fro \kappa_0^3 \varepsilon \delta + C_\Fro \kappa_0^2 \varepsilon \\
&= C_\Fro(2\kappa_0^3 \lambda_0 + \kappa_0^2) \varepsilon \,.
\end{split}
\end{equation*}
This gives $\Ncal(T_C, \|\cdot\|_\Lambda, C_\Fro(2\kappa_0^3 \lambda_\eta + \kappa_0^2) \varepsilon)  \leq \Ncal(\Sfrak_{k,\frac{1}{\kappa_0}, 2}^{-1}, \|\cdot\|_\Fro, \varepsilon) \times \Ncal(C, \|\cdot\|_E, \varepsilon)$. Therefore, by setting $C_{0} = C_\Fro(2\kappa_0^3 \lambda_0 + \kappa_0^2)$, we have
\begin{equation*}
\Ncal(T_C, \|\cdot\|_\Lambda, \varepsilon)  \leq \Ncal(\Sfrak_{k,\frac{1}{\kappa_0}, 2}^{-1}, \|\cdot\|_\Fro, \frac{\varepsilon}{C_{0}}) \times \Ncal(C, \|\cdot\|_\Fro, \frac{\varepsilon}{C_{0}}) \,.
\end{equation*}
\end{proof}

\begin{proof}[Proof of Proposition~\ref{prop:true_cov_number_short_chors_cond_nb}]
Recall that from Lemma~\ref{lemma:subset_cond_nb} we have
\begin{equation*}
\Ncal(T_C, \|\cdot\|_\Lambda, \varepsilon)  \leq \Ncal(\Sfrak_{k,\frac{1}{\kappa_0}, 2}^{-1}, \|\cdot\|_\Fro, \frac{\varepsilon}{C_{0}}) \times \Ncal(C, \|\cdot\|_\Fro, \frac{\varepsilon}{C_{0}}) \,,
\end{equation*}
with $ C_{0} = C_\Fro(2\kappa_0^3 \lambda_0 + \kappa_0^2)$. 
Using Lemma~\ref{lem:exists_tanget_cond_nb}, we also have the approximation of $\overline{S}^-_\eta[\Sfrak_{k, \kappa_0}]$ by the set $T_C$:
\begin{equation*}
\forall \Ubf \in \overline{S}^-_\eta[\Sfrak_{k, \kappa_0}], \exists \Tbf \in T_C, \ \|\Ubf-\Tbf\|_\Lambda \leq Z_0 \eta \,. 
\end{equation*}
Therefore, we can apply Lemma~\ref{lemma:generalized_cover} (with $\delta = Z_0 \eta$) to prove that for any $\varepsilon >0$:
\begin{equation}
\label{eq:covering_short_chords_with_Tcal}
\Ncal(\overline{S}^-_\eta[\Sfrak_{k, \kappa_0}], \|\cdot\|_\Lambda, 2(\varepsilon+Z_0 \eta)) \leq \Ncal(T_C, \|\cdot\|_\Lambda, \varepsilon) \,.
\end{equation}
Thus, combining \eqref{eq:covering_of_Tcal} and \eqref{eq:covering_short_chords_with_Tcal} yields
\begin{equation*}
\Ncal(\overline{S}^-_\eta[\Sfrak_{k, \kappa_0}], \|\cdot\|_\Lambda, 2(\varepsilon+ Z_0 \eta)) \leq \Ncal(\Sfrak_{k,\frac{1}{\kappa_0}, 2}^{-1}, \|\cdot\|_\Fro, \frac{\varepsilon}{C_{0}}) \times \Ncal(C, \|\cdot\|_\Fro, \frac{\varepsilon}{C_{0}}) \,.
\end{equation*}
All we need now is to control $\Ncal(C, \|\cdot\|_\Fro, \frac{\varepsilon}{C_{0}})$. 
Take $\overline{S[\Sfrak_{k, \kappa_0}^{-1}]}$ a $\varepsilon$-net of $S[\Sfrak_{k, \kappa_0}^{-1}]$ and $\overline{(0,\lambda_0]}$ an $(\lambda_0 \varepsilon)$-net of $(0,\lambda_0]$. Take $\Cbf = \lambda \Vbf \in C$ with $\lambda \in (0,\lambda_0]$ and $\Vbf \in S[\Sfrak_{k, \kappa_0}^{-1}]$. Then there exists $\overline{\Vbf} \in \overline{S[\Sfrak_{k, \kappa_0}^{-1}]}$ such that $\|\Vbf-\overline{\Vbf}\|_\Fro \leq \varepsilon$ and $\overline{\lambda} \in \overline{(0,\lambda_0]}$ such that $|\lambda - \overline{\lambda}|\leq \lambda_0 \varepsilon$. 
Define $\overline{\Cbf} = \overline{\lambda} \overline{\Vbf}$. 
We clearly have that $\overline{\Cbf} \in C$ and
\begin{equation*}
\begin{split}
\|\overline{\Cbf}-\Cbf\|_\Fro &= \| \overline{\lambda} \overline{\Vbf} - \lambda \Vbf \|_\Fro \\
&\leq \| \overline{\lambda} \overline{\Vbf} - \lambda \overline{\Vbf} \|_\Fro + \| \lambda \overline{\Vbf} - \lambda \Vbf \|_\Fro \\
&\leq  \| \Vbf \|_\Fro |\lambda - \overline{\lambda}| + \lambda \| \overline{\Vbf} - \Vbf \|_\Fro \leq 2 \lambda_0 \varepsilon \,.
\end{split}
\end{equation*}
Thus, for any $\varepsilon>0$ we have $\Ncal(C, \|\cdot\|_\Fro, 2 \lambda_0 \varepsilon) \leq  \Ncal((0,\lambda_0], |\cdot|, \lambda_0 \varepsilon) \Ncal(S[\Sfrak_{k, \kappa_0}^{-1}], \|\cdot\|_\Fro, \varepsilon) \leq \varepsilon^{-1} \Ncal(S[\Sfrak_{k, \kappa_0}^{-1}], \|\cdot\|_\Fro, \varepsilon)$  or equivalently for any $\varepsilon>0$ we have $\Ncal(C, \|\cdot\|_\Fro, \varepsilon) \leq 2 \lambda_0 \varepsilon^{-1} \Ncal(S[\Sfrak_{k, \kappa_0}^{-1}], \|\cdot\|_\Fro, \varepsilon / ( 2 \lambda_0))$. In conclusion we have
\begin{equation*}
\Ncal(\overline{S}^-_\eta[\Sfrak_{k, \kappa_0}], \|\cdot\|_\Lambda, 2(\varepsilon+ Z_0 \eta)) \leq 2 \lambda_\eta C_{0} \varepsilon^{-1} \Ncal(\Sfrak_{k,\frac{1}{\kappa_0}, 2}^{-1}, \|\cdot\|_\Fro, \frac{\varepsilon}{C_{0}}) \times \Ncal(S[\Sfrak_{k, \kappa_0}^{-1}], \|\cdot\|_\Fro, \frac{\varepsilon}{2 \lambda_0 C_{0}})\,,
\end{equation*}
which concludes the proof.
\end{proof}

\subsubsection{Covering number for bounded sparse symmetric matrices and their secant\label{proof:easy_covering_nb}}

We now prove Lemma~\ref{lemma:covering_secant_cond_nb}.
\begin{proof}[Proof of Lemma~\ref{lemma:covering_secant_cond_nb}]
Recall that Lemma~\ref{lemma:covering_simple} provides the following control of the covering number of the set of symmetric sparse matrices with bounded Frobenius norm $\mathfrak{W}_k = \left\{\Thetab \in S_{d}(\R) \ ;  \|\Thetab\|_{0} \leq d+2k, \|\Thetab\|_{\Fro} \leq 1\right\}$:
\begin{equation*}
\Ncal(\mathfrak{W}_k,\|\cdot\|_\Fro,\varepsilon) \leq (\frac{ed^2}{2k})^{k}(\frac{9}{\varepsilon})^{d+k}\,,
\end{equation*}

Noticing that $\Sfrak_0^{-1} \subset \mathfrak{W}_k$, $\Sfrak_{k, \frac{1}{\kappa_0},2}^{-1} \subset 2 \ \mathfrak{W}_k$ and that $S[\Sfrak_{k, \kappa_0}^{-1}] \subset \mathfrak{W}_{2k}$ yields the result.
\end{proof}

\subsubsection{Proof of Theorem~\ref{theo:big_theo_covering_cond_nb} \label{proof:big_theo_covering_cond_nb}}

\begin{proof}
First recall that for any $\varepsilon'>0$ and $\eta>0$, by Proposition~\ref{prop:true_cov_number_short_chors_cond_nb} we have
\begin{equation*}
\Ncal(\overline{S}^-_\eta[\Sfrak_{k, \kappa_0}], \|\cdot\|_\Lambda, 2(\varepsilon'+ Z_0 \eta)) \leq  2 \lambda_0 C_{0} \varepsilon^{-1} \Ncal(\Sfrak_{k,\frac{1}{\kappa_0}, 2}^{-1}, \|\cdot\|_\Fro, \frac{\varepsilon'}{C_{0}}) \times \Ncal(S[\Sfrak_{k, \kappa_0}^{-1}], \|\cdot\|_\Fro, \frac{\varepsilon'}{2 \lambda_0 C_{0}}) \,, \,,
\end{equation*}
Let us fix $\varepsilon$ such that $0 < \varepsilon \leq \frac{2 \kappa_0^3}{c_\Fro}\sqrt{d} = c_\Fro Z_0$, as in the hypothesis of Theorem~\ref{theo:big_theo_covering_cond_nb}. We now set $\varepsilon' = \varepsilon/4$ and $\eta = \varepsilon/(4Z_0)$ such that $2(\varepsilon'+ Z_0 \eta) = \varepsilon$. Remark that these choices satisfy $\eta \leq c_\Fro/2$ as desired in the previous section. Hence, 
\begin{equation*}
\begin{split}
&\Ncal(\overline{S}_\eta[\Sfrak_{k, \kappa_0}], \|\cdot\|_\Lambda, \varepsilon) \\
\leq &  \Ncal(\overline{S}^+_\eta[\Sfrak_{k, \kappa_0}], \|\cdot\|_\Lambda, \varepsilon) + \Ncal(\overline{S}^-_\eta[\Sfrak_{k, \kappa_0}], \|\cdot\|_\Lambda, \varepsilon) \\
\leq & \Ncal( (0,1], |\cdot|, \frac{\eta \varepsilon}{6L_0} ) \times \Ncal( \Sfrak_0^{-1}, \|\cdot\|_\Fro , \frac{\eta \varepsilon}{6L_0} )^2 \\
& + 2 \lambda_0 C_{0} \varepsilon'^{-1} \Ncal(\Sfrak_{k,\frac{1}{\kappa_0}, 2}^{-1}, \|\cdot\|_\Fro, \frac{\varepsilon'}{C_{0}}) \times \Ncal(S[\Sfrak_{k, \kappa_0}^{-1}], \|\cdot\|_\Fro, \frac{\varepsilon'}{2 \lambda_0 C_{0}}) \,. \\
\end{split}
\end{equation*}
Straightforward computations show that $\frac{\eta \varepsilon}{6L_0} = c_0 \varepsilon^2$ with $c_0 = \frac{1}{24Z_0 L_0}$, $\frac{\varepsilon'}{C_{0, \eta}} = c_1 \varepsilon$ where $c_1 = \frac{c_\Fro}{4C_\Fro(4\kappa_0^3  + c_\Fro \kappa_0^2 )}$ and $\frac{\varepsilon'}{2 \lambda_0 C_{0}} = c_2 \varepsilon$ with $c_2 = \frac{c_\Fro^2}{16 C_\Fro(4\kappa_0^3  + c_\Fro \kappa_0^2 )} \ (= c_\Fro c_1 / 4) $.  This allows us to continue the computation as follows:
\begin{equation*}
\begin{split}
&\Ncal(\overline{S}_\eta[\Sfrak_{k, \kappa_0}], \|\cdot\|_\Lambda, \varepsilon) \\
\leq & \Ncal( (0,1], |\cdot|, c_0 \varepsilon^2) \times N( \Sfrak_0^{-1}, \|\cdot\|_\Fro , c_0 \varepsilon^2)^2  \\
& + \frac{1}{c_2} \varepsilon^{-1} \Ncal(\Sfrak_{k,\frac{1}{\kappa_0}, 2}^{-1}, \|\cdot\|_\Fro, c_1 \varepsilon) \times \Ncal(S[\Sfrak_{k, \kappa_0}^{-1}], \|\cdot\|_\Fro, c_2 \varepsilon). \\
\leq & \frac{1}{c_0} \varepsilon^{-2} \times \left( \frac{e d^2}{2k} \right)^{2k} \left( \frac{18 \sqrt{d}}{c_0 \varepsilon^2}\right)^{2(d+k)} \\
& + \frac{1}{c_2} \varepsilon^{-1} \left( \frac{e d^2}{2k} \right)^{k} \left( \frac{2 \times 18 \sqrt{d}}{c_1 \varepsilon}\right)^{d+k} \times \left( \frac{e d (d-1)}{4k} \right)^{2k} \left( \frac{18 \sqrt{d}}{c_2 \varepsilon}\right)^{d+2k} \\
\leq & \frac{1}{c_0} \left( \frac{e d^2}{2k} \right)^{2k} \left( \frac{18 \sqrt{d}}{c_0 }\right)^{2(d+k)} \varepsilon^{-(4d+4k+2)} \\
& + \frac{1}{c_2}  \left( \frac{e d^2}{2k} \right)^{k} \left( \frac{2 \times 18 \sqrt{d}}{c_1}\right)^{d+k} \times \left( \frac{e d^2}{4k} \right)^{2k} \left( \frac{18 \sqrt{d}}{c_2 }\right)^{d+2k} \varepsilon^{-(2d+3k+1)}\\
\leq & \left( \frac{e d^2}{2k} \right)^{3k} \left[ \frac{(18 \sqrt{d})^{2(d+k)}}{c_0^{2(d+k)+1}} \varepsilon^{-(4d+4k+2)} +  \frac{2^{d+k} \ (18 \sqrt{d})^{2d+3k}}{c_1^{d+k} \ c_2^{d+2k+1}} \varepsilon^{-(2d+3k+1)} \right].
\end{split}
\end{equation*}

Recall that $c_0$, $c_1$ and $c_2$ depends on $d$ (directly or from $c_\Fro = 2/(9\sqrt{15}d)$ and $C_\Fro=1/\sqrt{d}$) and $\kappa_0$. Here are the explicit dependencies ($\gtrsim$ meaning here $\geq$ up to an absolute multiplicative constant): $c_0 \propto  \kappa_0^{-5} d^{-3/2}$, $c_1 \gtrsim \kappa_0^{-3} d^{-1/2}$ $c_2 \gtrsim \kappa_0^{-3} d^{-3/2}$. So we get, that there exists absolute constants $\tilde{c_1}$ and $\tilde{c_2}$ such that 
\begin{equation*}
\begin{split}
&\Ncal(\overline{S}_\eta[\Sfrak_{k, \kappa_0}], \|\cdot\|_\Lambda, \varepsilon) \\
\leq & \left( \frac{e d^2}{2k} \right)^{3k} \left[ \tilde{c_1}^{d+k} \frac{d^{d+k} \ (\kappa_0^{5} d^{3/2})^{2(d+k)+1}}{\varepsilon^{4d+4k+2}} +  \tilde{c_2}^{d+k}  \frac{\sqrt{d}^{2d+3k} (\kappa_0^3 d^{3/2})^{2d+3k+1}}{\varepsilon^{2d+3k+1}} \right] \\
\leq & \left( \frac{e d^2}{2k} \right)^{3k} \left[ \left( \frac{\tilde{c_1} \kappa_0^5 d^2}{\varepsilon^2} \right)^{2(d+k)+1} +   \left( \frac{\tilde{c_2} \kappa_0^3 d^2 }{\varepsilon} \right)^{2d+3k+1}\right],
\end{split}
\end{equation*}
where we change the value of the absolute constant $\tilde{c_1}$ and $\tilde{c_2}$ at the last inequality.
To finish the proof, recall that $\Ncal(S_\eta[\Sfrak_{k, \kappa_0}], \|\cdot\|_\Lambda, \varepsilon) \leq 2 \Ncal(\overline{S}_\eta[\Sfrak_{k, \kappa_0}], \|\cdot\|_\Lambda, \varepsilon)$. 
\end{proof}

\section*{Acknowledgments}
We gratefully acknowledge the support of the Centre Blaise Pascal's IT test platform at ENS de Lyon (Lyon, France) for Machine Learning facilities. The platform operates the SIDUS solution \cite{quemener2013sidus}. All the figures are generated using \texttt{Matplotlib} \cite{matplolib} and we also used \texttt{Scipy} and \texttt{Numpy} \cite{2020SciPyNMeth,harris2020array} for computing our estimator. The authors are grateful to Mathurin Massias for insightful discussions.

\section*{Funding}
This project was supported in part by the AllegroAssai ANR project ANR-19-CHIA-0009.

\section*{Data Availability Statements}
The data and code underlying this article are available at the following link \url{https://github.com/tvayer/CompressedPrecisionMatrix}.

\bibliographystyle{plain}

\begin{thebibliography}{}

\end{thebibliography}


\begin{thebibliography}{10}

\bibitem{7383810}
Sohail Bahmani and Justin Romberg.
\newblock Sketching for simultaneously sparse and low-rank covariance matrices.
\newblock In {\em 2015 IEEE 6th International Workshop on Computational
  Advances in Multi-Sensor Adaptive Processing (CAMSAP)}, pages 357--360, 2015.

\bibitem{Banerjee}
Onureena Banerjee, Laurent~El Ghaoui, and Alexandre d'Aspremont.
\newblock Model selection through sparse maximum likelihood estimation for
  multivariate gaussian or binary data.
\newblock {\em Journal of Machine Learning Research}, 9(15):485--516, 2008.

\bibitem{bauschke2017descent}
Heinz~H Bauschke, J{\'e}r{\^o}me Bolte, and Marc Teboulle.
\newblock A descent lemma beyond lipschitz gradient continuity: first-order
  methods revisited and applications.
\newblock {\em Mathematics of Operations Research}, 42(2):330--348, 2017.

\bibitem{bauschke2006joint}
Heinz~H Bauschke, Patrick~L Combettes, and Dominikus Noll.
\newblock Joint minimization with alternating bregman proximity operators.
\newblock {\em Pacific Journal of Optimization}, 2006.

\bibitem{bauschke2018regularizing}
Heinz~H Bauschke, Minh~N Dao, and Scott~B Lindstrom.
\newblock Regularizing with bregman--moreau envelopes.
\newblock {\em SIAM Journal on Optimization}, 28(4):3208--3228, 2018.

\bibitem{Beck}
Amir Beck.
\newblock {\em First-Order Methods in Optimization}.
\newblock SIAM-Society for Industrial and Applied Mathematics, Philadelphia,
  PA, USA, 2017.

\bibitem{squic}
Matthias Bollh\"{o}fer, Aryan Eftekhari, Simon Scheidegger, and Olaf Schenk.
\newblock Large-scale sparse inverse covariance matrix estimation.
\newblock {\em SIAM Journal on Scientific Computing}, 41(1), 2019.

\bibitem{boumal2023introduction}
Nicolas Boumal.
\newblock {\em An introduction to optimization on smooth manifolds}.
\newblock Cambridge University Press, 2023.

\bibitem{Bourrier}
Anthony Bourrier, Mike~E. Davies, Tomer Peleg, Patrick P{\'e}rez, and R{\'e}mi
  Gribonval.
\newblock {Fundamental performance limits for ideal decoders in
  high-dimensional linear inverse problems}.
\newblock {\em {IEEE Transactions on Information Theory}}, pages 7928--7946,
  December 2014.

\bibitem{cboyd}
Stephen Boyd and Lieven Vandenberghe.
\newblock {\em Convex Optimization}.
\newblock {Cambridge University Press}, March 2004.

\bibitem{bregman1967relaxation}
Lev~M Bregman.
\newblock The relaxation method of finding the common point of convex sets and
  its application to the solution of problems in convex programming.
\newblock {\em USSR computational mathematics and mathematical physics},
  7(3):200--217, 1967.

\bibitem{ROParticle}
T.~Tony Cai and Anru Zhang.
\newblock Rop: Matrix recovery via rank-one projections.
\newblock {\em The Annals of Statistics}, 43(1), Feb 2015.

\bibitem{clime}
Tony Cai, Weidong Liu, and Xi~Luo.
\newblock A constrained l1 minimization approach to sparse precision matrix
  estimation.
\newblock {\em Journal of the American Statistical Association},
  106(494):594--607, 2011.

\bibitem{candeslowrank}
Emmanuel Cand\`{e}s and Benjamin Recht.
\newblock Exact matrix completion via convex optimization.
\newblock {\em Commun. ACM}, 55(6):111–119, jun 2012.

\bibitem{candes2005decoding}
Emmanuel~J Candes and Terence Tao.
\newblock Decoding by linear programming.
\newblock {\em IEEE transactions on information theory}, 51(12):4203--4215,
  2005.

\bibitem{censor1992proximal}
Yair Censor and Stavros~Andrea Zenios.
\newblock Proximal minimization algorithm with d-functions.
\newblock {\em Journal of Optimization Theory and Applications},
  73(3):451--464, 1992.

\bibitem{chatalic:tel-03023287}
Antoine Chatalic.
\newblock {\em {Efficient and privacy-preserving compressive learning}}.
\newblock Theses, {Universit{\'e} Rennes 1}, November 2020.

\bibitem{chatalichal01701121}
Antoine Chatalic, R{\'e}mi Gribonval, and Nicolas Keriven.
\newblock {Large-Scale High-Dimensional Clustering with Fast Sketching}.
\newblock In {\em {ICASSP 2018 - IEEE International Conference on Acoustics,
  Speech and Signal Processing}}, pages 4714--4718, Calgary, Canada, April
  2018. {IEEE}.

\bibitem{Chen_yuxin}
Yuxin Chen, Yuejie Chi, and Andrea~J. Goldsmith.
\newblock Exact and stable covariance estimation from quadratic sampling via
  convex programming.
\newblock {\em IEEE Transactions on Information Theory}, 61(7):4034--4059,
  2015.

\bibitem{clakrson_chords}
Kenneth~L. Clarkson.
\newblock Tighter bounds for random projections of manifolds.
\newblock In {\em Proceedings of the Twenty-Fourth Annual Symposium on
  Computational Geometry}, SCG '08, page 39–48, New York, NY, USA, 2008.
  Association for Computing Machinery.

\bibitem{cohen2021has}
Regev Cohen, Yochai Blau, Daniel Freedman, and Ehud Rivlin.
\newblock It has potential: Gradient-driven denoisers for convergent solutions
  to inverse problems.
\newblock {\em Advances in Neural Information Processing Systems (NeurIPS)},
  34:18152--18164, 2021.

\bibitem{cohen2021regularization}
Regev Cohen, Michael Elad, and Peyman Milanfar.
\newblock Regularization by denoising via fixed-point projection (red-pro).
\newblock {\em SIAM Journal on Imaging Sciences}, 14(3):1374--1406, 2021.

\bibitem{coleman2012calculus}
R.~Coleman.
\newblock {\em Calculus on Normed Vector Spaces}.
\newblock Universitext. Springer New York, 2012.

\bibitem{delogne2022rop}
R{\'e}mi Delogne, Vincent Schellekens, and Laurent Jacques.
\newblock Rop inception: signal estimation with quadratic random sketching.
\newblock In {\em 30th European Symposium on Artificial Neural Networks,
  Computational Intelligence and Machine Learning}, 2022.

\bibitem{dempster1972covariance}
Arthur~P Dempster.
\newblock Covariance selection.
\newblock {\em Biometrics}, pages 157--175, 1972.

\bibitem{Ortega}
Hilmi~E. Egilmez, Eduardo Pavez, and Antonio Ortega.
\newblock Graph learning from data under laplacian and structural constraints.
\newblock {\em IEEE Journal of Selected Topics in Signal Processing},
  11(6):825--841, 2017.

\bibitem{erdos59a}
P.~Erd\"{o}s and A.~R\'{e}nyi.
\newblock On random graphs i.
\newblock {\em Publicationes Mathematicae Debrecen}, 6, 1959.

\bibitem{scadglasso}
Jianqing Fan, Yang Feng, and Yichao Wu.
\newblock {Network exploration via the adaptive LASSO and SCAD penalties}.
\newblock {\em The Annals of Applied Statistics}, 3(2):521 -- 541, 2009.

\bibitem{Fino}
Fino and Algazi.
\newblock Unified matrix treatment of the fast walsh-hadamard transform.
\newblock {\em IEEE Transactions on Computers}, C-25(11):1142--1146, 1976.

\bibitem{foucart13}
Simon Foucart and Holger Rauhut.
\newblock {\em A Mathematical Introduction to Compressive Sensing}.
\newblock 2013.

\bibitem{friedman_sparse_2008}
Jerome Friedman, Trevor Hastie, and Robert Tibshirani.
\newblock Sparse inverse covariance estimation with the graphical lasso.
\newblock {\em Biostatistics}, 9(3):432--441, July 2008.

\bibitem{frusquephd}
G.~Frusque.
\newblock {\em {Inf{\'e}rence et d{\'e}composition modale de r{\'e}seaux
  dynamiques en neurosciences}}.
\newblock PhD thesis, {Universit{\'e} de Lyon}, December 2020.

\bibitem{gribonval2020compressive}
R{\'e}mi Gribonval, Gilles Blanchard, Nicolas Keriven, and Yann Traonmilin.
\newblock {Compressive Statistical Learning with Random Feature Moments}.
\newblock {\em {Mathematical Statistics and Learning}}, 3, 2021.

\bibitem{gribonval2020statistical}
R{\'e}mi Gribonval, Gilles Blanchard, Nicolas Keriven, and Yann Traonmilin.
\newblock {Statistical Learning Guarantees for Compressive Clustering and
  Compressive Mixture Modeling}.
\newblock {\em {Mathematical Statistics and Learning}}, 3, 2021.

\bibitem{gribonval:hal-03350599}
R{\'e}mi Gribonval, Antoine Chatalic, Nicolas Keriven, Vincent Schellekens,
  Laurent Jacques, and Philip Schniter.
\newblock {Sketching Data Sets for Large-Scale Learning: Keeping only what you
  need}.
\newblock {\em {IEEE Signal Processing Magazine}}, 38(5):12--36, September
  2021.

\bibitem{gribonval2020characterization}
R{\'e}mi Gribonval and Mila Nikolova.
\newblock A characterization of proximity operators.
\newblock {\em Journal of Mathematical Imaging and Vision}, 62(6-7):773--789,
  2020.

\bibitem{hagberg2008exploring}
Aric Hagberg, Pieter Swart, and Daniel S~Chult.
\newblock Exploring network structure, dynamics, and function using networkx.
\newblock Technical report, Los Alamos National Lab.(LANL), Los Alamos, NM
  (United States), 2008.

\bibitem{time_glasso}
David Hallac, Youngsuk Park, Stephen Boyd, and Jure Leskovec.
\newblock Network inference via the time-varying graphical lasso.
\newblock In {\em Proceedings of the 23rd ACM SIGKDD International Conference
  on Knowledge Discovery and Data Mining}, KDD '17, page 205–213, New York,
  NY, USA, 2017. Association for Computing Machinery.

\bibitem{han2021riemannian}
Andi Han, Bamdev Mishra, Pratik~Kumar Jawanpuria, and Junbin Gao.
\newblock On riemannian optimization over positive definite matrices with the
  bures-wasserstein geometry.
\newblock {\em Advances in Neural Information Processing Systems (NeurIPS)},
  34:8940--8953, 2021.

\bibitem{harris2020array}
Charles~R. Harris, K.~Jarrod Millman, St{\'{e}}fan~J. van~der Walt, Ralf
  Gommers, Pauli Virtanen, David Cournapeau, Eric Wieser, Julian Taylor,
  Sebastian Berg, Nathaniel~J. Smith, Robert Kern, Matti Picus, Stephan Hoyer,
  Marten~H. van Kerkwijk, Matthew Brett, Allan Haldane, Jaime~Fern{\'{a}}ndez
  del R{\'{i}}o, Mark Wiebe, Pearu Peterson, Pierre G{\'{e}}rard-Marchant,
  Kevin Sheppard, Tyler Reddy, Warren Weckesser, Hameer Abbasi, Christoph
  Gohlke, and Travis~E. Oliphant.
\newblock Array programming with {NumPy}.
\newblock {\em Nature}, 585(7825):357--362, sep 2020.

\bibitem{ChoQUIC}
Cho-Jui Hsieh, M{{\'a}}ty{{\'a}}s~A. Sustik, Inderjit~S. Dhillon, and Pradeep
  Ravikumar.
\newblock Quic: Quadratic approximation for sparse inverse covariance
  estimation.
\newblock {\em Journal of Machine Learning Research}, 15(83):2911--2947, 2014.

\bibitem{bigQUIC}
Cho-Jui Hsieh, Matyas~A Sustik, Inderjit~S Dhillon, Pradeep~K Ravikumar, and
  Russell Poldrack.
\newblock Big \& quic: Sparse inverse covariance estimation for a million
  variables.
\newblock In C.J. Burges, L.~Bottou, M.~Welling, Z.~Ghahramani, and K.Q.
  Weinberger, editors, {\em Advances in Neural Information Processing Systems},
  volume~26. Curran Associates, Inc., 2013.

\bibitem{matplolib}
J.~D. Hunter.
\newblock Matplotlib: A 2d graphics environment.
\newblock {\em Computing in Science \& Engineering}, 9(3):90--95, 2007.

\bibitem{hurault2023convergent}
Samuel Hurault, Ulugbek Kamilov, Arthur Leclaire, and Nicolas Papadakis.
\newblock Convergent bregman plug-and-play image restoration for poisson
  inverse problems.
\newblock {\em arXiv preprint arXiv:2306.03466}, 2023.

\bibitem{hurault2022proximal}
Samuel Hurault, Arthur Leclaire, and Nicolas Papadakis.
\newblock Proximal denoiser for convergent plug-and-play optimization with
  nonconvex regularization.
\newblock In {\em International Conference on Machine Learning (ICML)}, pages
  9483--9505. PMLR, 2022.

\bibitem{jaganathan2016phase}
Kishore Jaganathan, Yonina~C Eldar, and Babak Hassibi.
\newblock Phase retrieval: An overview of recent developments.
\newblock {\em Optical Compressive Imaging}, pages 279--312, 2016.

\bibitem{low_rank_Kabanava}
Maryia Kabanava, Richard Kueng, Holger Rauhut, and Ulrich Terstiege.
\newblock Stable low-rank matrix recovery via null space properties.
\newblock {\em Information and Inference: A Journal of the IMA}, 5(4):405--441,
  2016.

\bibitem{keriven2018instance}
Nicolas Keriven and R{\'e}mi Gribonval.
\newblock Instance optimal decoding and the restricted isometry property.
\newblock In {\em Journal of Physics: Conference Series}, volume 1131, page
  012002. IOP Publishing, 2018.

\bibitem{KUENG201788}
Richard Kueng, Holger Rauhut, and Ulrich Terstiege.
\newblock Low rank matrix recovery from rank one measurements.
\newblock {\em Applied and Computational Harmonic Analysis}, 42(1):88--116,
  2017.

\bibitem{rope}
M.~O. Kuismin, J.~T. Kemppainen, and M.~J. Sillanpää.
\newblock Precision matrix estimation with rope.
\newblock {\em Journal of Computational and Graphical Statistics},
  26(3):682--694, 2017.

\bibitem{Sandeep}
Sandeep Kumar, Jiaxi Ying, Jos{\'e} Vin{\'\i}cius de~M Cardoso, and Daniel~P
  Palomar.
\newblock A unified framework for structured graph learning via spectral
  constraints.
\newblock {\em Journal of Machine Learning Research}, 21(22):1--60, 2020.

\bibitem{lau2022bregman}
Tim Tsz-Kit Lau and Han Liu.
\newblock Bregman proximal langevin monte carlo via bregman-moreau envelopes.
\newblock In {\em International Conference on Machine Learning}, pages
  12049--12077. PMLR, 2022.

\bibitem{lauritzen1996graphical}
Steffen~L Lauritzen.
\newblock {\em Graphical models}, volume~17.
\newblock Clarendon Press, 1996.

\bibitem{le2013fastfood}
Quoc Le, Tam{\'a}s Sarl{\'o}s, Alex Smola, et~al.
\newblock Fastfood-approximating kernel expansions in loglinear time.
\newblock In {\em Proceedings of the international conference on machine
  learning}, volume~85, page~8, 2013.

\bibitem{7605512}
Yuanxin Li, Yue Sun, and Yuejie Chi.
\newblock Low-rank positive semidefinite matrix recovery from corrupted
  rank-one measurements.
\newblock {\em IEEE Transactions on Signal Processing}, 65(2):397--408, 2017.

\bibitem{lingjaerde2021tailored}
Camilla Lingj{\ae}rde, Tonje~G Lien, {\O}rnulf Borgan, Helga Bergholtz, and
  Ingrid~K Glad.
\newblock Tailored graphical lasso for data integration in gene network
  reconstruction.
\newblock {\em BMC bioinformatics}, 22(1):1--22, 2021.

\bibitem{LIU2015153}
Weidong Liu and Xi~Luo.
\newblock Fast and adaptive sparse precision matrix estimation in high
  dimensions.
\newblock {\em Journal of Multivariate Analysis}, 135:153--162, 2015.

\bibitem{gabor}
Gabor Lugosi, Jakub Truszkowski, Vasiliki Velona, and Piotr Zwiernik.
\newblock Learning partial correlation graphs and graphical models by
  covariance queries.
\newblock {\em Journal of Machine Learning Research}, 22(203):1--41, 2021.

\bibitem{Rahul_new_perspective}
Rahul Mazumder and Trevor Hastie.
\newblock {The graphical lasso: New insights and alternatives}.
\newblock {\em Electronic Journal of Statistics}, 6(none):2125 -- 2149, 2012.

\bibitem{scikitlearn}
F.~Pedregosa, G.~Varoquaux, A.~Gramfort, V.~Michel, B.~Thirion, O.~Grisel,
  M.~Blondel, P.~Prettenhofer, R.~Weiss, V.~Dubourg, J.~Vanderplas, A.~Passos,
  D.~Cournapeau, M.~Brucher, M.~Perrot, and E.~Duchesnay.
\newblock Scikit-learn: Machine learning in {P}ython.
\newblock {\em Journal of Machine Learning Research}, 12:2825--2830, 2011.

\bibitem{benfenati2020proximal}
J-C Pesquet.
\newblock Proximal approaches for matrix optimization problems: Application to
  robust precision matrix estimation.
\newblock {\em Signal Processing}, 169:107417, 2020.

\bibitem{recipies_rip}
Gilles Puy, Mike~E. Davies, and R{\'e}mi Gribonval.
\newblock {Recipes for stable linear embeddings from Hilbert spaces to
  $\mathbb{R}^m$}.
\newblock {\em {IEEE Transactions on Information Theory}}, 2017.
\newblock Submitted in 2015.

\bibitem{quemener2013sidus}
Emmanuel Quemener and Marianne Corvellec.
\newblock Sidus—the solution for extreme deduplication of an operating
  system.
\newblock {\em Linux Journal}, 2013(235):3, 2013.

\bibitem{Ravikumar}
Pradeep Ravikumar, Martin~J. Wainwright, Garvesh Raskutti, and Bin Yu.
\newblock {High-dimensional covariance estimation by minimizing l1-penalized
  log-determinant divergence}.
\newblock {\em Electronic Journal of Statistics}, 5:935 -- 980, 2011.

\bibitem{robinson_2010}
James~C. Robinson.
\newblock {\em Dimensions, Embeddings, and Attractors}.
\newblock Cambridge Tracts in Mathematics. Cambridge University Press, 2010.

\bibitem{rolfs2012iterative}
Benjamin Rolfs, Bala Rajaratnam, Dominique Guillot, Ian Wong, and Arian Maleki.
\newblock Iterative thresholding algorithm for sparse inverse covariance
  estimation.
\newblock {\em Advances in Neural Information Processing Systems}, 25, 2012.

\bibitem{ryu2019plug}
Ernest Ryu, Jialin Liu, Sicheng Wang, Xiaohan Chen, Zhangyang Wang, and Wotao
  Yin.
\newblock Plug-and-play methods provably converge with properly trained
  denoisers.
\newblock In {\em International Conference on Machine Learning}, pages
  5546--5557. PMLR, 2019.

\bibitem{saade2016random}
Alaa Saade, Francesco Caltagirone, Igor Carron, Laurent Daudet, Ang{\'e}lique
  Dr{\'e}meau, Sylvain Gigan, and Florent Krzakala.
\newblock Random projections through multiple optical scattering: Approximating
  kernels at the speed of light.
\newblock In {\em 2016 IEEE International Conference on Acoustics, Speech and
  Signal Processing (ICASSP)}, pages 6215--6219. IEEE, 2016.

\bibitem{barnajeenonconvex}
K.~Sagar, S.~Banerjee, J.~Datta, and A.~Bhadra.
\newblock Precision matrix estimation under the horseshoe-like prior-penalty
  dual, 2021.

\bibitem{trace_reg}
Martin Slawski, Ping Li, and Matthias Hein.
\newblock Regularization-free estimation in trace regression with symmetric
  positive semidefinite matrices.
\newblock In C.~Cortes, N.~Lawrence, D.~Lee, M.~Sugiyama, and R.~Garnett,
  editors, {\em Advances in Neural Information Processing Systems}, volume~28.
  Curran Associates, Inc., 2015.

\bibitem{nonconvexglasso}
Q.~Sun, Kean~M. Tan, H.~Liu, and T.~Zhang.
\newblock Graphical nonconvex optimization via an adaptive convex relaxation.
\newblock In {\em International Conference on Machine Learning}, volume~80,
  pages 4810--4817, 2018.

\bibitem{teboulle2018simplified}
Marc Teboulle.
\newblock A simplified view of first order methods for optimization.
\newblock {\em Mathematical Programming}, 170(1):67--96, 2018.

\bibitem{vershynin2012}
Roman Vershynin.
\newblock {\em Introduction to the non-asymptotic analysis of random matrices},
  page 210–268.
\newblock Cambridge University Press, 2012.

\bibitem{vershynin2018high}
Roman Vershynin.
\newblock {\em High-dimensional probability: An introduction with applications
  in data science}, volume~47.
\newblock Cambridge university press, 2018.

\bibitem{2020SciPyNMeth}
Pauli Virtanen, Ralf Gommers, Travis~E. Oliphant, Matt Haberland, Tyler Reddy,
  David Cournapeau, Evgeni Burovski, Pearu Peterson, Warren Weckesser, Jonathan
  Bright, St{\'e}fan~J. {van der Walt}, Matthew Brett, Joshua Wilson, K.~Jarrod
  Millman, Nikolay Mayorov, Andrew R.~J. Nelson, Eric Jones, Robert Kern, Eric
  Larson, C~J Carey, {\.I}lhan Polat, Yu~Feng, Eric~W. Moore, Jake
  {VanderPlas}, Denis Laxalde, Josef Perktold, Robert Cimrman, Ian Henriksen,
  E.~A. Quintero, Charles~R. Harris, Anne~M. Archibald, Ant{\^o}nio~H. Ribeiro,
  Fabian Pedregosa, Paul {van Mulbregt}, and {SciPy 1.0 Contributors}.
\newblock {{SciPy} 1.0: Fundamental Algorithms for Scientific Computing in
  Python}.
\newblock {\em Nature Methods}, 17:261--272, 2020.

\bibitem{wainwright_2019}
Martin~J. Wainwright.
\newblock {\em High-Dimensional Statistics: A Non-Asymptotic Viewpoint}.
\newblock Cambridge Series in Statistical and Probabilistic Mathematics.
  Cambridge University Press, 2019.

\bibitem{robustCLIME}
Lingxiao Wang and Quanquan Gu.
\newblock Robust {G}aussian graphical model estimation with arbitrary
  corruption.
\newblock In Doina Precup and Yee~Whye Teh, editors, {\em Proceedings of the
  34th International Conference on Machine Learning}, volume~70 of {\em
  Proceedings of Machine Learning Research}, pages 3617--3626. PMLR, 06--11 Aug
  2017.

\bibitem{Wang_fast_and_sparse}
Lingxiao Wang, Xiang Ren, and Quanquan Gu.
\newblock Precision matrix estimation in high dimensional gaussian graphical
  models with faster rates.
\newblock In Arthur Gretton and Christian~C. Robert, editors, {\em Proceedings
  of the 19th International Conference on Artificial Intelligence and
  Statistics}, volume~51 of {\em Proceedings of Machine Learning Research},
  pages 177--185, Cadiz, Spain, 09--11 May 2016. PMLR.

\bibitem{woo2017characterization}
Hyenkyun Woo.
\newblock A characterization of the domain of beta-divergence and its
  connection to bregman variational model.
\newblock {\em Entropy}, 19(9):482, 2017.

\bibitem{ell0GLASSO}
P.~Xu, J.~Ma, and Q.~Gu.
\newblock Speeding up latent variable gaussian graphical model estimation via
  nonconvex optimization.
\newblock In {\em Advances in Neural Information Processing Systems},
  volume~30, 2017.

\bibitem{elementary_esti}
Eunho Yang, Aurelie~C Lozano, and Pradeep~K Ravikumar.
\newblock Elementary estimators for graphical models.
\newblock In Z.~Ghahramani, M.~Welling, C.~Cortes, N.~Lawrence, and K.Q.
  Weinberger, editors, {\em Advances in Neural Information Processing Systems},
  volume~27. Curran Associates, Inc., 2014.

\bibitem{Jieping}
Jieping Ye and Jun Liu.
\newblock Sparse methods for biomedical data.
\newblock {\em SIGKDD Explor. Newsl.}, 14(1):4–15, dec 2012.

\bibitem{Ying2020}
J~Ying, J.~de~Miranda~Cardoso, and D.~Palomar.
\newblock Nonconvex sparse graph learning under laplacian constrained graphical
  model.
\newblock In {\em Advances in Neural Information Processing Systems},
  volume~33, pages 7101--7113, 2020.

\bibitem{orthogonal_RF}
Felix Xinnan~X Yu, Ananda~Theertha Suresh, Krzysztof~M Choromanski, Daniel~N
  Holtmann-Rice, and Sanjiv Kumar.
\newblock Orthogonal random features.
\newblock In D.~Lee, M.~Sugiyama, U.~Luxburg, I.~Guyon, and R.~Garnett,
  editors, {\em Advances in Neural Information Processing Systems}, volume~29.
  Curran Associates, Inc., 2016.

\bibitem{lin_prog}
Ming Yuan.
\newblock High dimensional inverse covariance matrix estimation via linear
  programming.
\newblock {\em Journal of Machine Learning Research}, 11(79):2261--2286, 2010.

\bibitem{yuan2007model}
Ming Yuan and Yi~Lin.
\newblock Model selection and estimation in the gaussian graphical model.
\newblock {\em Biometrika}, 94(1):19--35, 2007.

\bibitem{dtrace}
Teng Zhang and Hui Zou.
\newblock {Sparse precision matrix estimation via lasso penalized D-trace
  loss}.
\newblock {\em Biometrika}, 101(1):103--120, 02 2014.

\bibitem{zuo2017incorporating}
Yiming Zuo, Yi~Cui, Guoqiang Yu, Ruijiang Li, and Habtom~W Ressom.
\newblock Incorporating prior biological knowledge for network-based
  differential gene expression analysis using differentially weighted graphical
  lasso.
\newblock {\em BMC bioinformatics}, 18(1):1--14, 2017.

\end{thebibliography}

\end{document}